\documentclass{article}
\usepackage{iclr2027_conference,times}
\usepackage{graphicx}
\usepackage{amsmath}

\usepackage[ruled,vlined,linesnumbered]{algorithm2e}
\usepackage{booktabs}
\usepackage{amssymb}
\usepackage{pifont}
\usepackage{xcolor}
\usepackage{colortbl}
\usepackage{booktabs}
\usepackage{makecell}
\usepackage[table]{xcolor}
\usepackage{adjustbox}
\usepackage{comment}
\usepackage{mathtools} 
\usepackage{multirow}
\usepackage{subcaption}
\usepackage{listings}
\usepackage{xcolor}
\usepackage{url}
\usepackage{enumitem}
\usepackage{parskip}
\usepackage{wrapfig}
\usepackage[bookmarks=false]{hyperref}
\usepackage[capitalise,nameinlink]{cleveref}
\usepackage{wrapfig}
\usepackage{placeins}

\usepackage{amsmath}
\usepackage{amssymb}
\usepackage{mathtools}
\usepackage{bm}
\usepackage{tikz}
\usepackage{dsfont}

\newcommand{\TODO}[1]{{\color{red}TODO: #1}}

\newcommand{\gf}[2]{{\color{gray}#1}{\color{purple}#2}}

\newcommand{\AD}[2]{{\color{gray}#1}{\color{blue}#2}}

\newcommand{\OP}[2]{#2}

\definecolor{darkgreen}{RGB}{0,120,0}
\definecolor{darkred}{RGB}{160,0,0}

\definecolor{SDG}{HTML}{1F77B4}
\definecolor{VisionSDG}{HTML}{2CA02C}
\DeclareRobustCommand{\FSDG}[1]{{%
  \setlength{\fboxsep}{1pt}\colorbox{SDG!30}{#1}}}
\DeclareRobustCommand{\VSDG}[1]{{%
  \setlength{\fboxsep}{1pt}\colorbox{VisionSDG!30}{#1}}}

\DeclareRobustCommand{\VSDGmath}[1]{\VSDG{$\displaystyle #1$}}

\newcommand{\NLstr}[1]{\texttt{"}\!\hookleftarrow\!\texttt{#1"}}
\newcommand{\lstar}{\ensuremath{\ell^*}\xspace}
\newcommand{\lstarprev}{\ensuremath{\ell^*\!-\!1}\xspace}
\newcommand{\dimred}{Dimensionality reduction\xspace}

\title{
Are In-Context Images Worth 10 Dimensions?
}

\author{
\textbf{Adhemar de Senneville$^{1}$ \quad Xavier Bou$^{2}$ \quad J\'er\'emy Anger$^{1}$}\\
\textbf{Rafael Grompone$^{1}$ \quad Gabriele Facciolo$^{1,3}$}\\[0.7em]
\textnormal{$^{1}$Universit\'e Paris-Saclay, CNRS, ENS Paris-Saclay, Centre Borelli, Paris, France}\\
\textnormal{$^{2}$P\^ole recherche de l'AMIAD, Palaiseau, France}\\
\textnormal{$^{3}$Institut Universitaire de France, Paris, France}
}

\iclrfinalcopy 
\begin{document}

\maketitle
\lhead{} 

\begin{abstract}
There has been significant work on understanding the In-Context Learning capabilities of Large Language Models, especially on the induction circuit.
For a few-shot classification task, the induction circuit leverages
linear representations of each labeled example in-context in order to classify an unlabeled query.
However, few works focus on how those linear representations are built in the first place.
Leveraging the expressivity of the vision modality compared to text, we uncover a Shared Discriminative Geometry (SDG) inside Large Vision Language Models (LVLMs).
It is a low-dimensional space, shared across all image classification tasks, in which in-context images are compressed into linearly separable representations later used to perform classification.
We observe that this is the result of the model performing a dimensionality reduction of vision representations in early layers.
In order to explain this phenomenon:
(1) We show analytically 
that linear self-attention can perform a dimensionality reduction by projecting in-context data onto its principal components, with each layer implementing one gradient descent step toward this objective.
(2) We provide evidence that trained LVLMs reduce the dimensionality of vision representations in early layers via a similar mechanism.

\end{abstract}


\section{Introduction}

%
A fundamental problem in computer vision is learning to recognize novel concepts from only a handful of image examples. 
To address this problem, the standard approach is fine-tuning, where a pretrained feature extractor is adapted to the \OP{target}{}task through in-weight learning.
A promising alternative is \textbf{I}n-\textbf{C}ontext \textbf{L}earning (ICL),
which frames the problem as a conditional prediction problem, 
where the model receives a small labeled support set together with an unlabeled query image. 
%

ICL was first observed as an emerging capability in \textbf{L}arge \textbf{L}anguage \textbf{M}odels (LLMs)~\citep{brown2020language} in the text modality and was later observed in \textbf{L}arge \textbf{V}ision \textbf{L}anguage \textbf{M}odels (LVLMs)~\citep{alayrac2022flamingo} on multimodal vision-text tasks.
The extension of ICL to the vision modality motivates us to investigate how LVLMs perform it.
While the induction circuit~\citep{olsson2022incontextlearninginductionheads} 
by which the model compares in-context examples using linear representations 
to perform classification is well studied, here we focus on the formation of those representations.
In this study, we investigate this process through few-shot in-context image classification. 
Whereas prior work has focused on text classification, the continuous nature of visual inputs and the diversity of image classes allow us to better trace feature formation across a wide range of classification tasks.
\AD{
}{}

Surprisingly, preliminary experiments show that similarity-based classification performs much worse on text tokens predicting the class labels than on vision tokens.
Motivated by this observation, we use a meta-training protocol {to find the subspace that best classifies the query image given the support images.}
This protocol reveals a subspace inside of text tokens in which in-context images are compressed 
into representations later used by the model to predict the query label.
We name this subspace the \textbf{Shared Discriminative Geometry (SDG)} as it has 3 key properties:
(1) It is shared across all image classification tasks. 
(2) Classes are linearly separable in that space.
(3) It has a small effective dimensionality ($\sim$10 dimensions for Ministral 3 (3B)~\citep{liu2026ministral}). 

\begin{figure}[t]
    \centering
    \includegraphics[width=0.62\linewidth]{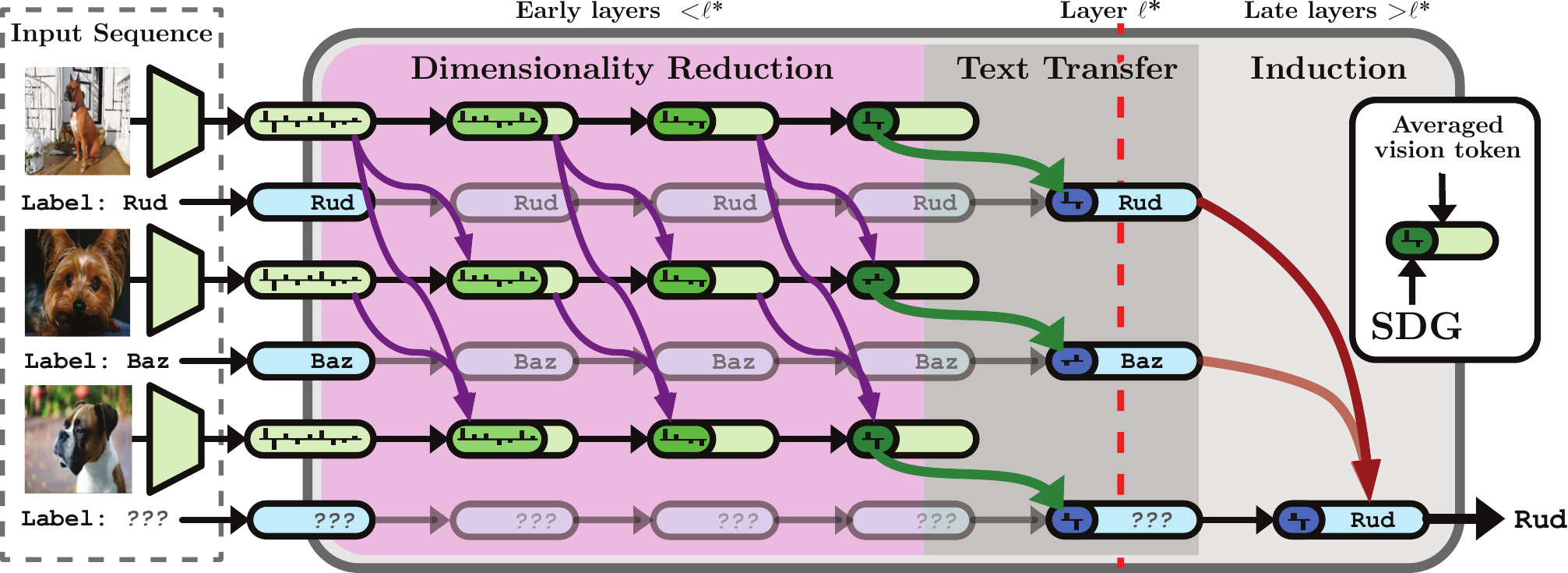}
    \includegraphics[width=0.37\linewidth]{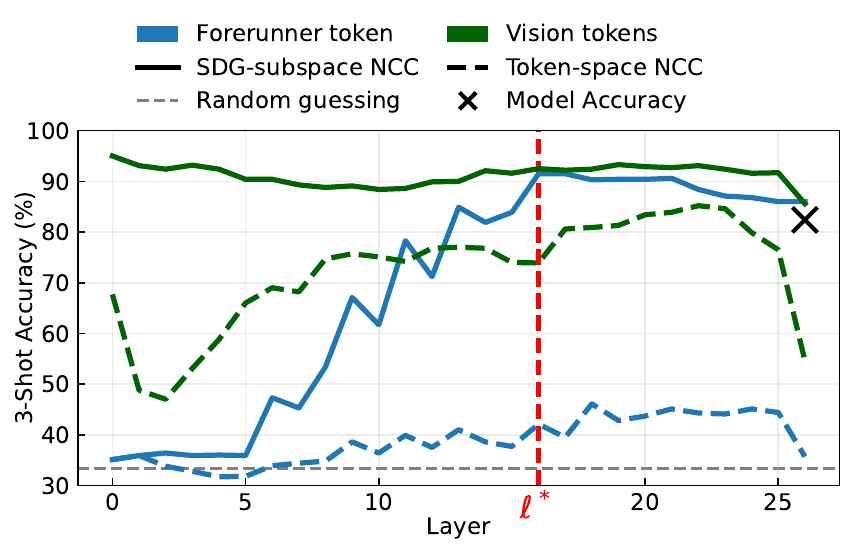}
    \caption{
\textbf{Left (our contribution):} The first layers of the LVLM reduce the dimension of image tokens in-context using sample-to-sample comparisons.
These representations are transferred to text and later used for in-context learning.
\textbf{Right:} Nearest Centroid Classifier (NCC) accuracy across layers for \FSDG{forerunner tokens} and \VSDG{vision tokens} after and before projecting onto the \FSDG{Text SDG} and \VSDG{Vision SDG}, respectively.
    }
    \vspace{-15pt}
    \label{fig:main}
\end{figure}

We posit  
that this surprisingly low dimensionality 
results from the model performing a 
dimensionality reduction in-context,
effectively projecting discriminative high-dimensional vision features onto the principal components of the in-context examples, preserving the linear separability with fewer dimensions. 
As illustrated in \cref{fig:main}, this dimensionality reduction occurs in early layers in vision tokens. The resulting representations are then transferred to text tokens. 
The model finally uses this subspace to attend to similar in-context examples.

\textbf{Our contributions can be summarized as: }
(I)
We introduce a meta-training protocol that uncovers a Shared Discriminative Geometry in text token representations. It is a low-dimensional space shared across image classification tasks in which classes are linearly separable.
The model uses this geometry to make predictions.
(II)
We show theoretically that non-causal linear transformers 
can perform a dimensionality reduction through in-context sample-to-sample attention.
(III) We observe a dimensionality reduction of vision token representations in LVLMs and provide empirical evidence that it relies on a circuit similar to our theoretical construction.



\section{Related Work} 
\paragraph{LLMs ICL Mechanistic Interpretability.}
Mechanistic Interpretability aims to find an algorithmic description and circuits that describe the complex capabilities of models.
In-context Classification has been a common test bed for studying ICL mechanisms.
For text classification, prior work has identified two independent circuits~\citep{cho2025revisiting,wang2023label,yang2025unifying} distributed in the early and the later layers of the transformer.
In early layers, the last token of each in-context example {aggregates} information into a linearly separable representation.
In later layers, the induction circuit uses induction heads~\citep{olsson2022incontextlearninginductionheads} that exploit these representations to infer the query class label from similar support examples.
Little work has focused on the mechanism by which those high-quality representations are formed. 

\paragraph{LLMs Subspaces.}
Previous work has identified specific subspaces within LLMs.
\cite{hu2025understanding} found a 4D subspace tracking the unit digit through
trigonometric functions. 
\cite{ICLR2025_d3221cdb} found a 2D subspace characterizing weekday, month, and year.
\cite{pmlr-v267-wollschlager25a} found up to 5D subspaces characterizing refusal behavior across several LLM families.
Recently, \cite{huang2026decomposing} proposed a method to learn subspaces in an unsupervised manner. 
We uniquely study in LVLMs a subspace tied to the vision modality.
Unlike prior fixed semantic subspaces, SDG occupies a fixed subspace, yet we show that its inner geometry is context dependent.

\paragraph{Transformers ICL Theory.}
Prior work has shown that linear transformers can implement algorithms in-context such as 
linear regression~\citep{von2023transformers,ahn2023transformers}, 
expectation maximization
~\citep{chen2026transformers} and 
reinforcement-learning algorithms
~\citep{ICLR2024_84b5e56a}.
To our knowledge, we show 
for the first time that linear transformers can also perform dimensionality reduction by projecting examples onto the principal components of the current context.
Nonetheless, the convergence of this algorithm has been studied previously~\citep{8811622}.
\section{Preliminaries}

%
%
\subsection{Experiment Setting}

\paragraph{Datasets.}
We use 10 standard publicly available image classification datasets across a diverse range of domains: 
EuroSAT~\citep{CD_eurosat}, 
UCF101~\citep{CD_ucf101}, 
DTD~\citep{CD_DTD}, 
Caltech101~\citep{CD_calthech_101},
SUN397~\citep{CD_sun},
OxfordPets~\citep{CD_oxford_pets}, 
StanfordCars~\citep{CD_stanford_cars}, 
Flowers102~\citep{CD_flowers_102}, 
Food101~\citep{CD_food_101}, and 
FGVCAircraft~\citep{CD_aircraft}. 
The last 5 are fine-grained image classification datasets.
The expressivity of the vision modality helps us probe LVLM mechanisms shared across classification tasks for two reasons. 
First, working with images allows us to sample diverse classification tasks from $\sim$ 2000 fine-grained classes.
This is significantly harder to achieve with the less diverse set of classification tasks in text.
Second, text offers a less diverse set of inputs due to its discrete nature {compared to vision}.

\paragraph{Task.}
We consider in-context $K$-way $N$-shot few-shot classification 
where the LVLM is given $M:=NK+1$ images $\{\mathrm{img}_i\}_{i=1}^{M}$.
Only the first $NK$ images are labeled $\mathbf{c}=\{c_i\}_{i=1}^{NK}$, with $N$ examples for each of the $K$ classes $c_i \in \{1,\ldots,K\}$, while $\mathrm{img}_M$ is the unlabeled query image.
LVLMs are composed of a vision encoder, and an LLM decoder transformer.
The LLM decoder takes as input a sequence of interleaved vision and text tokens. 
Text tokens are deduced from the embedding vocabulary $f_t$, whereas 
vision tokens are outputs of the vision encoder given an image
\(
 f_v(\text{img})
\).
We build the ICL classification task as the following token sequence: 
\begin{equation}
\begin{aligned}
S_i &= \Big[
f_t(\NLstr{image:}),
\VSDGmath{f_v(\text{img}_i)},
f_t(\NLstr{label\FSDG{\strut:}}),
f_t(y_{c_i})
\Big], \quad i=1,\ldots,NK \\[2mm]
Q &= \Big[
f_t(\NLstr{image:}),
\VSDGmath{f_v(\text{img}_M)},
f_t(\NLstr{label\FSDG{\strut:}})
\Big], 
\mathcal{T} = \Big[ S_1,\ldots,S_{NK},Q \Big],
\end{aligned}
\label{eq:prompt}
\end{equation}
where \([\cdot]\) denotes token concatenation, and \(y_{c_i}\) denotes the text associated with class label \(c_i\).
\(S_i\) denotes the \(i\)-th support example, \(Q\) the query example without its label, and \(\mathcal{T}\) the full in-context sequence.
For each task, we randomly select one of the ten datasets and sample a 3-way, 3-shot classification task from it. Thus, 10 images are given to the LVLM for each task.
Label tokens are abstract and semantically unaligned with the visual classes.
Following \cite{confavreuxcomparing} we use \texttt{"\_Rud"}, \texttt{"\_Baz"} and \texttt{"\_Gip"}.
Given that multimodal ICL has been shown to rely strongly on textual inputs over visual inputs~\citep{chen2025can_wacv,bias_world_vision}, using abstract labels removes the influence of label semantics from the learning dynamics.
{Indeed, this forces the model to classify the query image using the support set without “zero-shotting” the class from memory. }
See supplementary material~\ref{sec:label_space} for more experiments on the impact of the choice of label space.

We refer to the forerunner token as the token \texttt{"\FSDG{\strut:}"} placed after \texttt{"label"} and responsible for predicting the {class} label token; it is therefore responsible for the image classification task.
Throughout this work we will probe vision and forerunner representations across layers. Thus, we denote the forerunner token hidden state for image $i$ at layer $\ell$ output by \FSDG{$h_{i,\ell}\in\mathbb{R}^D$} and the average-pooled vision token hidden state by \VSDG{$v_{i,\ell}\in\mathbb{R}^D$}, where $D$ is the LVLM hidden-state dimension.
We also include the input to the first layer, denoted by \(\ell=0\) by convention.
In the absence of a dedicated aggregation mechanism, such as a CLS token~\citep{dosovitskiy2020image} or a trained pooling~\citep{lee2019set,he2022revisiting}, we average-pool vision tokens to obtain image representations~\citep{tian2020rethinking}. Note that this is just a proxy of vision token representations, for example the token from the center of the object classifies better than tokens on the background.


\paragraph{Models.}
All results in the paper are shown using Ministral~3 (3B)
as its architecture does not use linear attention or local attention, simplifying the analysis.
Additional results for Qwen~3.5 (4B and 9B)~\citep{qwen3.5}, Qwen~3 (4B)~\citep{yang2025qwen3} and Ministral~3 (8B) are provided in the supplementary material~\ref{sec:more_models}.
%
%

Following the formulation of \cite{elhage2021mathematical}, an attention head is parameterized by query, key, value, and output projections $W_Q,W_K,W_V,W_O$. Let $R=[r_1,\ldots,r_n]$ be its input states, the head contribution to the update of the token $r_n$ is
\begin{equation}
\Delta r_n
=
W_{\mathrm{OV}}R\,
\operatorname{softmax}\!\left(
\frac{R^\top W_{\mathrm{KQ}}r_n}{\sqrt{d_h}}
\right),
\qquad
W_{\mathrm{KQ}}:=W_K^\top W_Q,
\quad
W_{\mathrm{OV}}:=W_OW_V .
\end{equation}
The KQ circuit 
$
r_i^\top W_{\mathrm{KQ}}r_n
=
(W_Kr_i)^\top(W_Qr_n)
$
determines the attention score of each input to the query token.
This score is a \textit{bilinear similarity}, meaning that it is linear in either token representation when the other is fixed.
The OV circuit maps via $W_{\mathrm{OV}}$ each input token to its contribution to $r_n$.


\subsection{Motivation}

In this section we motivate our work by the following observation: 
\textbf{Raw representations of the forerunner token, which predicts the class label, are not class linear separable, whereas vision-token representations are.}
Throughout this study, we will use the Nearest Centroid Classifier (NCC) accuracy as a proxy to assess class linear separability, or more broadly speaking “feature quality”.
This is similar to the standard practice of using linear probing classification accuracy as a proxy for feature quality of foundation models~\citep{chen2020simple,oquab2023dinov2,pmlr-v119-wang20k}.
NCC  measures class separability {of a representation space}{} by testing whether examples of the same class cluster around a common direction in the representation space.
{
}
It is a classical few-shot classifier~\citep{NIPS2017_cb8da676,hu2022pushing,li2021universal}
that can be seen as a simple form of in-context learning.
Given an in-context feature matrix
\(X=[x_1,\ldots,x_{NK},x_{M}]\in\mathbb{R}^{D\times(NK+1)}\)
and the labels \(\mathbf{c}\) of the first $NK$ examples, 
the NCC predicts the label of the query $x_{M}$ as
\begin{equation}
\operatorname{NCC}_{\mathrm{pred}}(X,\mathbf{c})=\arg\max_{k\in\{1,\ldots,K\}}
x_{M}^{\top}\mu_k,
\qquad \mu_k=\frac{1}{N}\sum_{i:c_i=k}x_i.
\label{eq:ncc_pred}
\end{equation}
As commonly done with NCC, we use the dot product as the similarity function. Note this corresponds to a bilinear similarity in the full representation space 
$x_M^\top I_D\mu_k=x_M^\top\mu_k.$
Unless specified otherwise, reported accuracies are averages over 1000 tasks for the 3-shot in-context settings.

\Cref{fig:main} (Right) shows the LVLM accuracy
(cross \tikz[baseline=(x.base)]\node[inner sep=0pt] (x) {$\times$};), as well as the NCC accuracy 
(dotted lines) 
of \FSDG{forerunner tokens $h_{i,\ell}$} and \VSDG{vision tokens $v_{i,\ell}$} across LLM decoder layers.
Vision token NCC accuracy is comparable to model accuracy across most layers.
However, we observe that the forerunner token is significantly less class discriminative, with an accuracy close to random guessing.
This is surprising, since the forerunner token is expected to aggregate visual information to correctly classify the query image~\citep{cho2025revisiting}.
\textbf{This indicates that either information is not transferred to the forerunner token or that it is only present in a small subspace of its high-dimensional embedding.}

\section{Shared Discriminative Geometry }
In this section, 
we show the existence of a Shared Discriminative Geometry in both the text forerunner token (\FSDG{Text SDG}) and vision tokens (\VSDG{Vision SDG}), in which image classes are linearly separable.
Then, we show that the Text SDG mechanistically contributes to the model's in-context performance.
Finally, we identify a two-step circuit describing the formation of the Text SDG.

\subsection{Learning the Shared Discriminative Geometry}

\begin{wrapfigure}{r}{0.58\textwidth}
    \vspace{-10pt}
    \centering
    \includegraphics[width=\linewidth,clip,trim={0 0 0.3cm 0}]{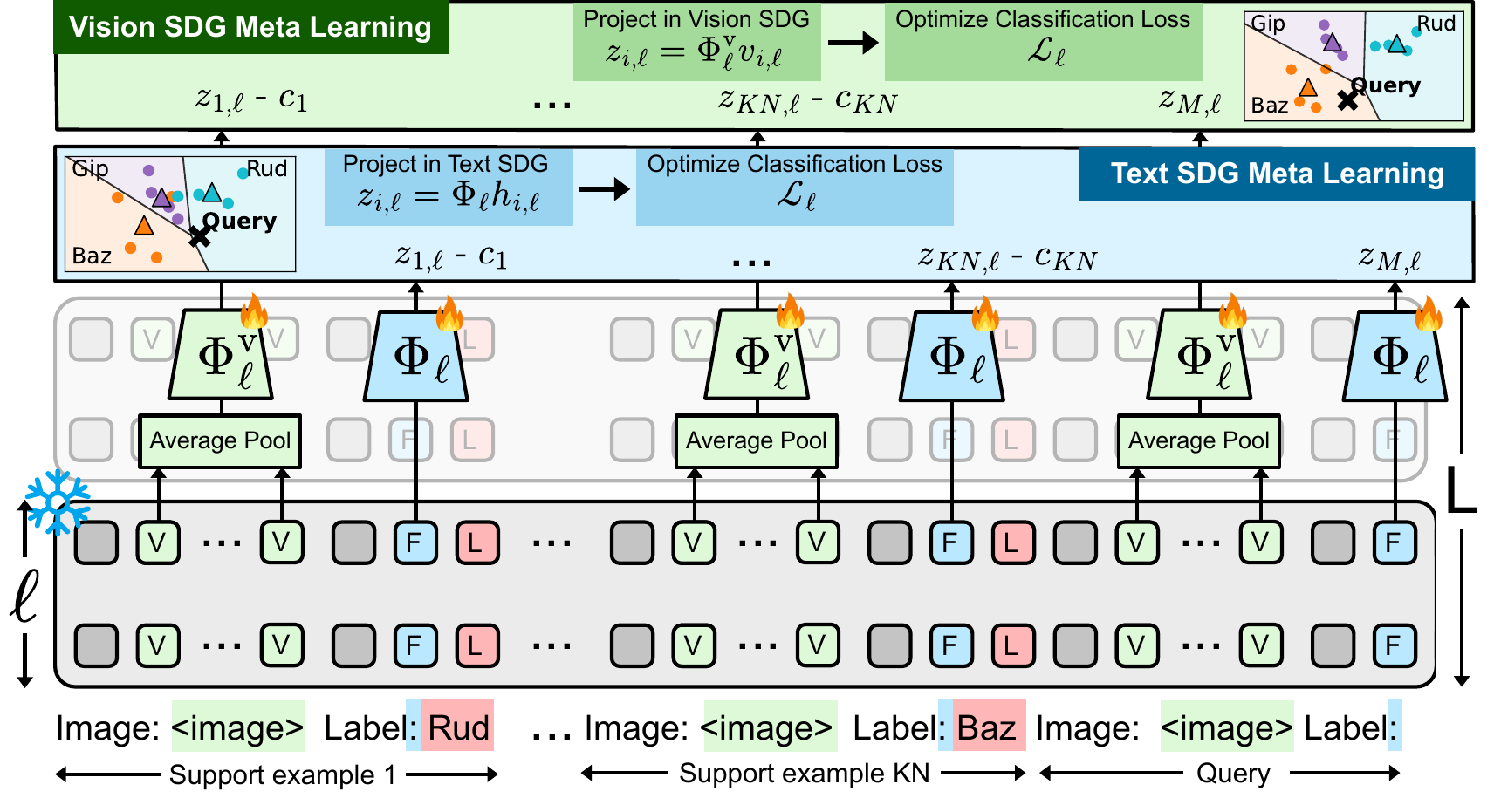}
    \vspace{-15pt}
    \caption{
    Training setup. At each layer, we meta-learn the \FSDG{Text SDG} projection \(\Phi_\ell\) and \VSDG{Vision SDG} projection \(\Phi_\ell^{\mathrm v}\) so that projected embeddings from the same class cluster around a common direction.
    }
    \vspace{-10pt}
    \label{fig:training}
    \label{fig:meta_train_combined}
\end{wrapfigure}
In order to probe discriminative representations within the forerunner token, in this section we introduce a protocol to learn the \FSDG{Text SDG} (see \cref{fig:meta_train_combined}), 
which is a set of linear down-projections \(\{\Phi_\ell\}_{0,\dots,L}\) for each output layer of the LVLM. 
Each \(\Phi_\ell\) projects the forerunner token into a subspace \(\Phi_\ell:\mathbb{R}^D \to \mathbb{R}^{D_\downarrow}\) where the dot product is {class} discriminative between the support examples.
For Ministral 3B $D=3072$ and $D_\downarrow=128$.
We freeze the LVLM as only the projections \(\Phi_\ell\) are learned. 
We refer to SDG as a geometry because each $\Phi_\ell$ defines a bilinear similarity
$
{h_{i,\ell}}^\top\Phi_\ell^\top\Phi_\ell h_{j,\ell},
$
which measures similarity between forerunner token representations at layer $\ell$.
%
During training, we first project forerunner tokens \(h_{i,\ell}\) onto the SDG \(z_{i,\ell}=\Phi_\ell h_{i,\ell}\).
Then, for each in-context classification task, we compute one centroid per class and define the per-layer meta-training loss as
\begin{equation}
\mu_{k,\ell}=\frac{1}{N}\sum_{i:c_i=k}z_{i,\ell},
\qquad
\mathcal L_{\ell}=\sum_{k=1}^{K}\left(
z_{M,\ell}^\top\mu_{k,\ell}-\mathds{1}[c_M=k]
\right)^2,
\end{equation}
where $\mathds{1}$ denotes the indicator function. 
The loss trains \(\Phi_\ell\) to align the projected query with the centroid of its class (dot product \(\sim 1\)) while making it orthogonal to the centroids of other classes (dot product \(\sim 0\)). 
Importantly, the forerunner token cannot carry information about its current support label thanks to the causal nature of LLMs. Indeed, the label token is placed after the forerunner token, ensuring no leakage of the correct label into the representations.
We train for 32k tasks with a batch size of 32 using AdamW, a learning rate of \(0.001\), and a weight decay of \(0.001\).
Weight decay is necessary to stabilize training by suppressing directions that carry no discriminative signal or are never activated by the training data.
In order to better trace the formation of the Text SDG, using the same protocol we also learn the \VSDG{Vision SDG} \(\{\Phi_\ell^{\mathrm v}\}_{0,\dots,L}\). Similarly, each \(\Phi_\ell^{\mathrm v}\) projects the average-pooled vision token $v_{i,\ell}$ into a subspace 
where the dot product is {class} discriminative between the support images 
(see \cref{fig:meta_train_combined}).

\paragraph{Results.}
The NCC accuracy of the forerunner token after projecting onto the \FSDG{Text SDG} across layers is shown in \cref{fig:main} (Right, solid line).
The accuracy starts rising at layer 6 and peaks at layer $\lstar=16$. Here, we denote the layer of peak accuracy by \lstar. 
At \lstar the accuracy even surpasses the model performance. 
This finding is consistent with prior work identifying the induction circuit in late layers as a bottleneck to model performance~\citep{cho2025token} and showing that internal representations can surpass the model performance~\citep{de2026unlocking}.
We train \(\Phi_\ell\) to down-project onto \(\mathbb{R}^{128}\)
which is the dimensionality of the representation processed by attention heads.
However, we find that its effective dimension is $\sim10$ at \lstar~(estimated with the participation ratio; \cite{gao2017theory}).
{This measure quantifies the effective number of directions contributing to the pairwise similarities induced by $\Phi_\lstar$.}
For comparison, the average effective dimension of the model’s attention heads is $\sim90$ out of a maximum of 128.
While the effective dimensionality of \(\Phi_\ell\) is a proxy and may vary with the training setup (see supplementary material~\ref{sec:meta_training_hparams} and~\ref{sec:meta_training_single_dataset}), \textbf{it is striking that we can identify a fixed 10-dimensional subspace in which representations still achieve 90\% accuracy across 10 domain-specific datasets.}
\begin{figure}[t]
\centering
\begin{minipage}[c]{0.34\linewidth}
    \centering
    \includegraphics[width=\linewidth]{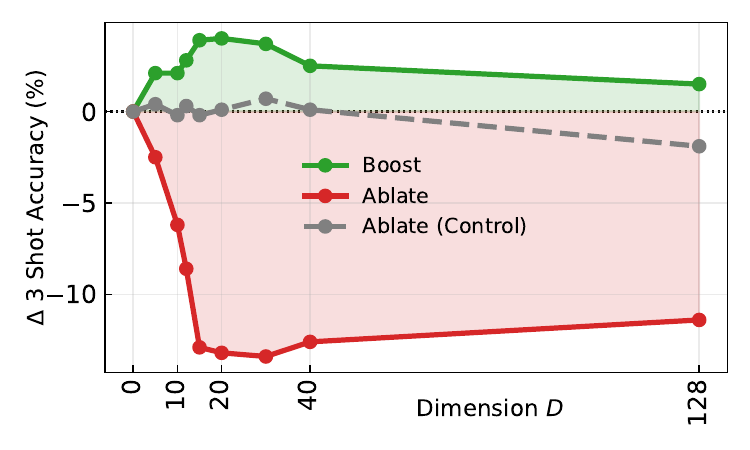}
\end{minipage}\hfill
\begin{minipage}[c]{0.62\linewidth}
\centering
\small
\setlength{\tabcolsep}{3pt}
\resizebox{\linewidth}{!}{%
\begin{tabular}{lccccc}
\toprule
Intervention & 1-shot & 2-shot & 3-shot & 4-shot & 5-shot \\
\midrule
LVLM Accuracy & 34.4\% & 77.2\% & 82.4\% & 84.2\% & 85.4\% \\
\midrule
128D-Control Ablation & -0.4 & -1.2 & -1.9 & -2.5 & -2.2 \\
15D-SDG Ablation & \textbf{\textcolor{darkred}{-7.1}} & \textbf{\textcolor{darkred}{-10.8}} & \textbf{\textcolor{darkred}{-12.9}} & \textbf{\textcolor{darkred}{-12.0}} & \textbf{\textcolor{darkred}{-10.6}} \\
15D-SDG Boost & \textbf{\textcolor{darkgreen}{+7.5}} & \textbf{\textcolor{darkgreen}{+4.7}} & \textbf{\textcolor{darkgreen}{+3.9}} & \textbf{\textcolor{darkgreen}{+3.0}} & \textbf{\textcolor{darkgreen}{+2.9}} \\
\bottomrule
\end{tabular}
}
\end{minipage}
\vspace{-6pt}
\caption{\textbf{Left:} 
Effect of SDG intervention as a function of the number of top dimensions $D$ ablated.
\textbf{Right:} 
Effect of interventions on the top 15 SDG dimensions across shot counts.
}
\vspace{-15pt}
\label{fig:sdg_interventions}
\end{figure}
\paragraph{Ablation.}
Next, we ablate the \FSDG{Text SDG} at layer $\lstar$ to test if it mechanistically contributes to the LVLM's ability to classify images.
Let $V_k$ contain the top-$k$ right singular vectors of $\Phi_{\lstar}$.
For all hidden states at $\lstar$ output, we remove the top-$k$ Text SDG dimensions by projecting onto their orthogonal complement giving
\(
\tilde{\mathbf{h}}_{\lstar}
=(I -
V_kV_k^\top
)\mathbf{h}_{\lstar}
\). 
As a control, we do the same for a 128-dimensional random subspace. 
We also try boosting the top-$k$ SDG dimensions at \lstar:\linebreak
\(
\tilde{\mathbf{h}}_{\lstar}
=(I + V_kV_k^\top)\mathbf{h}_{\lstar}.
\)
This operation essentially doubles the magnitude of the SDG and should directly increase the image-to-image comparison signal used for classification. 
As shown in \cref{fig:sdg_interventions} (left), only 15 dimensions capture most of the ablation effect.
Ablating the top 15 dimensions of the SDG decreases model accuracy by 12.9\%, boosting them increases accuracy by 3.9\%, whereas ablating a random subspace has a negligible effect. 
This is consistent with our observation that the geometry lies in a few dimensions,
although completely eliminating its contribution to classification requires ablating 5 additional dimensions.
{In this paper, we retain effective dimensionality as our measure of the SDG dimensionality because it can be computed at every layer and measures a property of its geometry.} 
Additional results from 1 to 5 shots for the 15-dimensional SDG ablation and the 128-dimensional control random-subspace  are shown in the right panel of \cref{fig:sdg_interventions}.
\textbf{This shows that SDG is low-dimensional and is necessary to the model’s ability to classify images in-context.} 
More ablations are provided in supplementary material~\ref{sec:more_ablation_exp}.

\subsection{Text SDG formation as a two-step process}
%
%
%
%
%
In this section, we provide empirical evidence that this low-dimensionality is the result of the LVLM performing a two-step circuit: first the model reduces the dimensionality of vision tokens in the \VSDG{Vision SDG} from layer \(1\) to \lstarprev, then the model transfers the resulting geometry to the \FSDG{Text SDG} in the forerunner token at layer \lstar.

\begin{figure*}[t]
    \centering

    \begin{subfigure}[t]{0.32\textwidth}
        \centering
        \includegraphics[width=\linewidth]{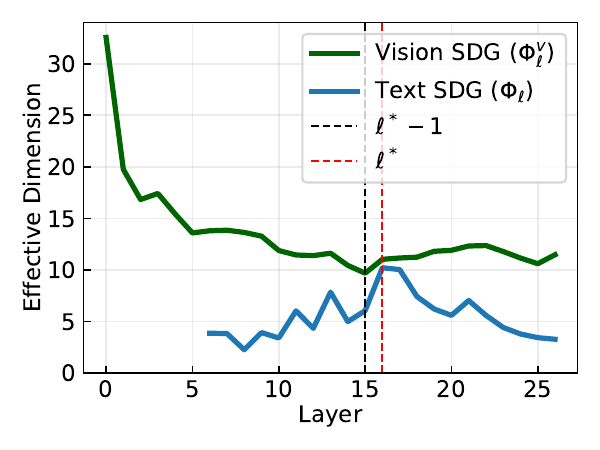}
        \label{fig:effective_dim}
    \end{subfigure}
    \hfill
    \begin{subfigure}[t]{0.32\textwidth}
        \centering
        \includegraphics[width=\linewidth]{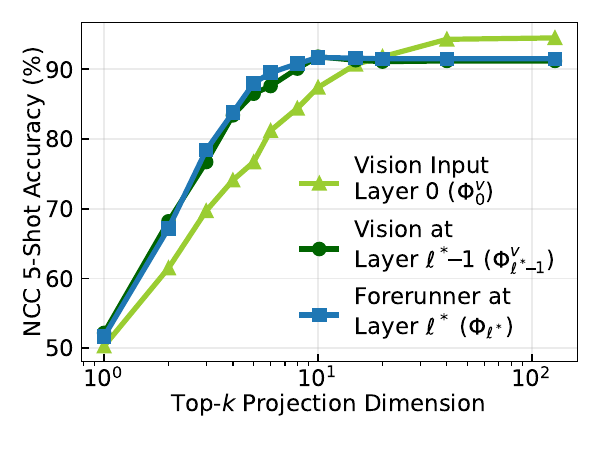}
        \label{fig:dim_compression}
    \end{subfigure}
    \hfill
    \begin{subfigure}[t]{0.32\textwidth}
        \centering
        \includegraphics[width=\linewidth]{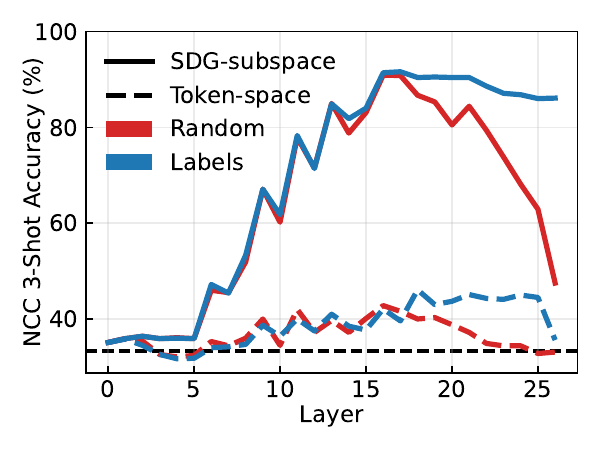}
        \label{fig:rand_norand}
    \end{subfigure}
    \vspace{-20pt}
    \caption{
    \textbf{Left}: Effective Dimension across layers of the $\Phi_\ell$ (Text SDG) and $\Phi^{\mathrm v}_\ell$ (Vision SDG).
    \textbf{Middle}: NCC Accuracy in the top-$k$ SDG dimensions of the input vision tokens, the vision tokens at layer \lstarprev and forerunner token at \lstar.
    \textbf{Right}: Text SDG with and without random label assignment.
    }
    \label{fig:empirical}
    \vspace{-10pt}
\end{figure*}

\paragraph{Step 1: Reducing dimensionality in vision tokens from layer $1$ to \lstarprev.}
\label{sec:step_1}
\Cref{fig:empirical} (Left)
reports the effective dimension of both the Vision and Text SDG across layers. 
The Vision SDG decreases from 32 dimensions for the input vision tokens to a minimum of 10 at layer \lstarprev while preserving similar accuracy.
To investigate this further, we test whether vision tokens at layer \lstarprev concentrate class-discriminative information into fewer dimensions than the vision tokens output by the vision encoder.
In \cref{fig:empirical} (Middle),
we report NCC accuracy for the input vision tokens and the vision tokens at layer \lstarprev after projecting them onto the top \(k\) dimensions of \(\Phi_0^{\mathrm v}\) and \(\Phi_{\lstarprev}^{\mathrm v}\), respectively.
We observe that the low dimensional projections are  more class discriminative at layer \lstarprev in vision tokens than in input vision tokens. At the same time the full space accuracy is not higher at layer \lstarprev.
Interestingly the top 10 dimensions at layer \lstarprev have 29\% error reduction compared to input vision representations ($+4$ accuracy points). Put differently at layer \lstarprev vision tokens have a 90\% accuracy using $\sim$ 2 times less dimensions. 
%
%

\begin{wraptable}{l}{0.38\textwidth}
\centering
\small
\setlength{\tabcolsep}{3pt}
\vspace{-10pt}
\begin{tabular}{lc}
\toprule
Ablation effect (3-shot) & $\Delta\%$ \\ %
\midrule
(1) 128D-\FSDG{Text SDG at \lstarprev}                         & $-3.2$  \\
(2) 128D-\VSDG{Vision SDG at \lstarprev} & $-6.4$  \\
(3) Attention from vision at & $-5.2$  \\
\lstarprev to text at \lstar  &    \\
\cmidrule(lr){1-2}
\quad\enspace128D-\FSDG{Text SDG at \lstar}                   & $-11.4$ \\
\bottomrule
\end{tabular}
\caption{
Ablations from \lstarprev to \lstar 
}
\label{tab:vision_attention_ablation}
\vspace{-10pt}
\end{wraptable}
\paragraph{Step 2: Transfer to text from \lstarprev to \lstar.}
\Cref{fig:empirical} (Middle) shows the NCC accuracy for the forerunner token at layer \lstar after projecting it onto the top \(k\) dimensions of \(\Phi_{\lstar}\). It shows that the forerunner token at layer \lstar closely matches the feature quality of the average-pooled vision token at the previous layer \lstarprev. 
In \cref{fig:empirical} (Left),
similarly the effective dimension of $\Phi_{\lstar}$ matches the one of $\Phi^{\mathrm v}_{\lstarprev}$.
This suggests that most of the vision representations are transferred to the forerunner token between layer \lstarprev and \lstar.
To test this, we separately ablate (1) Text SDG at \lstarprev, (2) Vision SDG at \lstarprev and (3) attention from vision tokens at \lstarprev to text tokens at \lstar, see Table~\ref{tab:vision_attention_ablation}. 
Ablating the Vision SDG at \lstarprev reduces accuracy twice as much as ablating the Text SDG at the same layer ($6.4$ vs.\ $3.2$ points). Ablating attention from vision tokens at \lstarprev to the forerunner token at \lstar causes a similar drop ($5.2$ points), suggesting that this attention transfers class-discriminative information from the Vision SDG at \lstarprev to text tokens at \lstar.
Additional experiments in the supplementary material~\ref{sec:vision_forerunner_flow} examine this vision-to-text transfer in detail. Specifically, we observe that heads performing this operation copy class-discriminative representations from specific vision tokens called sink tokens~\citep{kang2025see}.

\section{Why the SDG is so low-dimensional}

\subsection{Explaining What Drives In-Context Dimensionality Reduction}

\paragraph{In-Context Labels Do Not Drive Dimensionality Reduction.}
A possible explanation is that the Text SDG formation is driven by induction or supervision: the model may use {preceding} support labels to progressively compare in-context examples and optimally compress them in a class discriminative geometry. To test this hypothesis we learn Text SDG with and without assigning random label tokens to images.
\Cref{fig:empirical} (Right) shows that random label assignment has close to no effect during the  formation phase of Text SDG on its accuracy and on its peak accuracy. 
The formation circuit therefore extracts the same class-discriminative visual features regardless of label assignment of precedent support examples.
It is important to note that this is not the case after Text SDG formation, where accuracy is higher with correct label assignments than with random label assignments, as shown in \cref{fig:empirical} (Right).
The well studied induction circuit~\citep{olsson2022incontextlearninginductionheads} uses image-label assignment to compress input examples in a class-discriminative lower dimensional space. 
{
Indeed, after induction, at the final layer, the model accuracy is the accuracy of the forerunner token in a 3-dimensional space.
This space is spanned by the unembedding directions of the 3 possible labels: \texttt{"Rud"}, \texttt{"Baz"} and  \texttt{"Gip"}.
}

\paragraph{Preserving the In-Context Gram Matrix Can Drive Dimensionality Reduction.}
Since Text SDG formation is independent of support-label assignments, it is impressive that the model performs well classifying planes, satellite images or even textures by building an intermediary image representation in such a small shared dimensional space.
However, we make two observations.
(1) For a given classification task, the model receives a fixed context of \(M\) images, whose high-dimensional representations we collect as \(X\in \mathbb{R}^{D\times M}\). 
We assume that \(\operatorname{rank}(X)=M\).
(2)~Although the examples are represented in a \(D\)-dimensional space, NCC classifies the query using only similarities to the support examples. The classification score for class \(k\) in Eq.~\eqref{eq:ncc_pred} is the average similarity to support of class $k$ 
\begin{equation}
x_M^\top \mu_k
= \frac{1}{N}\sum_{i:c_i=k}x_M^\top x_i
= \frac{1}{N}\sum_{i:c_i=k}(X^\top X)_{M,i},
\end{equation}
and therefore depends only on the Gram matrix \(X^\top X\) or equivalently on sample-to-sample similarity.
%
Thus, for a fixed context, there exists a low-dimensional \(Z\in\mathbb{R}^{D_\downarrow\times M}\) with \(M\leq D_\downarrow \ll D\) such that \(Z^\top Z=X^\top X\) in which NCC accuracy is unchanged.
All solutions are orthonormal arrangements of the principal components of \(X\):
\(Z=QU^\top X\) (\(Q\in\mathbb{R}^{D_\downarrow\times M}\), \(Q^\top Q=I_M\)),
where \(U\) contains the \(M\) principal directions of \(X\). 
These principal directions are determined using (uncentred) principal component analysis (PCA), with $XX^\top=U\Lambda U^\top$.
This observation raises the  following questions: 
Can transformers reduce the dimensionality of in-context high dimensional inputs while preserving their pairwise similarities?
If yes, do we observe a similar circuit in LVLMs?

\subsection{\dimred through Self-Attention}
In this section, we show that in a simplified setting, transformer layers can perform dimensionality reduction via sample-to-sample attention. 

\paragraph{In-context setup.}
Given $M$ in-context examples
$x_i \in \mathbb{R}^{D_\uparrow}$ in a high dimensional space, we associate each token with a compressed state
$z_{i,\ell} \in \mathbb{R}^{D_\downarrow}$, where $D_\downarrow \ll D_\uparrow$. We define the
context matrix at layer $\ell$ as
\begin{equation}
C_\ell
=
\begin{pmatrix}
x_1 & \cdots & x_M \\
z_{1,\ell} & \cdots & z_{M,\ell}
\end{pmatrix}
\in \mathbb{R}^{(D_\uparrow+D_\downarrow)\times M}.
\end{equation}
The upper block $X\in \mathbb{R}^{D_\uparrow\times M}$ stores the fixed high-dimensional in-context input tokens, while the lower block $Z_\ell\in \mathbb{R}^{D_\downarrow\times M}$ stores the compressed variables updated by the attention layers.
Note that $Z_\ell$ is in this simplified setup tied to a canonical basis $(e_{D_\uparrow+1},\ldots,e_{D_\uparrow+D_\downarrow})$ meaning $\Phi_\ell=(0\;\;I_{D_\downarrow})$. 
However the model can represent the compressed in-context examples in any \(D_\downarrow\)-dimensional orthonormal basis.
In the real model, this basis may even partially overlap with the subspace carrying $X$. Importantly, $Z_\ell$ is expressed in a fixed basis making it shared across tasks while the main direction of $X$ may vary from a task to another.
The geometries of the original and compressed representations are encoded by the Gram matrices $X^\top X$ and $Z^{\top} Z$, 
and the objective is to make the compressed Gram matrix match the original one.
This is equivalent to minimizing 
\begin{equation}
\mathcal{E}(Z)
=
\frac{1}{4}
\left\|
Z^\top Z- X^\top X
\right\|_F^2,
\qquad 
\text{where} 
\qquad 
\nabla_Z \mathcal{E} = Z\left(Z^\top Z - X^\top X\right) .
\end{equation}

\paragraph{Assumptions.} \textbf{1. Non-causal attention.}
Here, we assume that each token can attend to every other token.
This is the strongest assumption, as most LVLMs are causal. This means that the first example cannot modify its state given the context as it does not have access to the other tokens, making the process less efficient.
\textbf{2. Linear self-attention.}
Following the standard linear  self-attention \citep{ahn2024linear, von2023transformers}, we remove the softmax, as well as the normalization and write 
\begin{equation}
\operatorname{LSA}(C_{\ell})
=
C_\ell
+
W_{OV} C_\ell C_\ell^\top W_{KQ} C_\ell
=C_{\ell+1}
.
\end{equation}
\paragraph{Optimization.}
Under these assumptions, choosing
\begin{equation}
W_{KQ}
=
\begin{pmatrix}
I_{D_\uparrow} & 0 \\
0 & -I_{D_\downarrow}
\end{pmatrix}
\qquad\text{and }\qquad
W_{OV}
=
\eta
\begin{pmatrix}
0 & 0 \\
0 & I_{D_\downarrow}
\end{pmatrix},
\label{eq:OV_and_QK}
\end{equation}
the lower block of $\operatorname{LSA}(C_\ell)$ is equivalent to a gradient step toward minimizing $\mathcal{E}$ 
\begin{equation}
Z_{\ell+1}
=
Z_\ell
+\eta Z_\ell(X^\top X-Z_\ell^\top Z_\ell)
=
Z_\ell
-
\eta \nabla_Z \mathcal{E}(Z_\ell)
\end{equation}
with $\eta$ the learning rate. 
This is similar to prior work showing that linear attention can implement a single gradient-descent step toward solving linear regression~\citep{ahn2023transformers}. Therefore, converging to the solution requires multiple layers.

\paragraph{Convergence.}

Let \(\lambda_1 \geq \cdots \geq \lambda_M > 0\) denote the eigenvalues of \(X^\top X\).
To obtain an explicit convergence bound, we assume
\begin{equation}
\operatorname{rank}(Z_0)=M
\qquad\text{and}\qquad
0<\eta<
\frac{1}{4\left(\lambda_1+\left\|Z_0^\top Z_0-X^\top X\right\|_F\right)}.
\end{equation}
A standard transformer can realize $Z_0$ initialization through positional embeddings assigning linearly independent compressed states to the \(M\) in-context examples, or simply by starting with a random subspace of X. Additionally $\eta$ must be small enough.
Under those assumptions, the linear self-attention dynamics converge locally at an exponential rate:
\begin{equation}
\mathcal{E}(Z_\ell)
=O\!\left(\exp\!\left(-4\eta\lambda_M\ell\right)\right).
\label{eq:main_convergence}
\end{equation}
We provide the convergence proof, empirically validate the result and include the effect of each assumption, in the supplementary material~\ref{sec:convergence} and~\ref{sec:numerical}.
\textbf{
Overall, under our idealized assumptions, linear self-attention can compress \(M\) input representations from \(D_\uparrow\) to \(D_\downarrow\) dimensions, where the sample-to-sample similarity error  converges locally to zero at an exponential rate.
}



%

\subsection{Evidence of a similar mechanism in LVLMs}
\begin{figure}[t]
    \centering
    \includegraphics[width=0.33\linewidth]{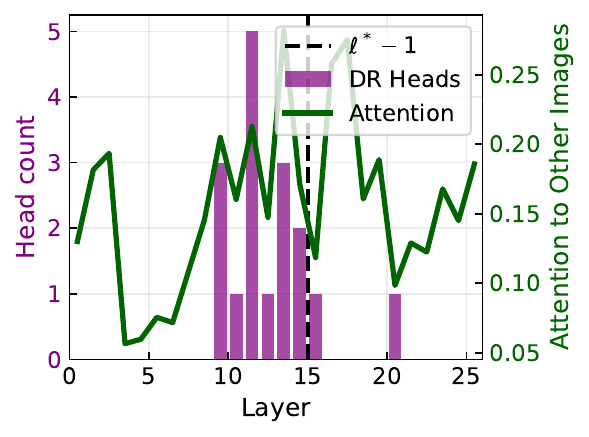}%
    \hfill
    \includegraphics[width=0.33\linewidth]{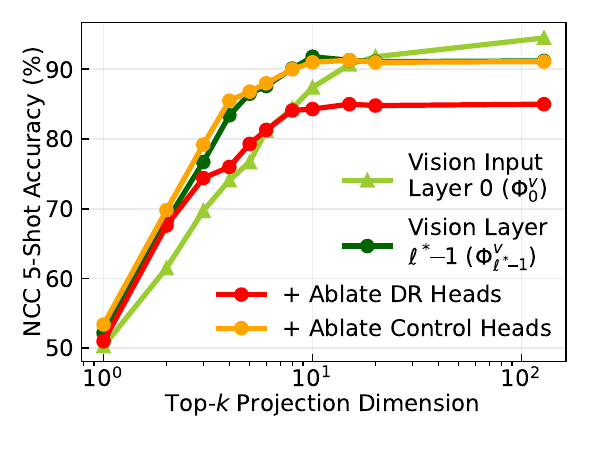}%
    \hfill
    \includegraphics[width=0.33\linewidth]{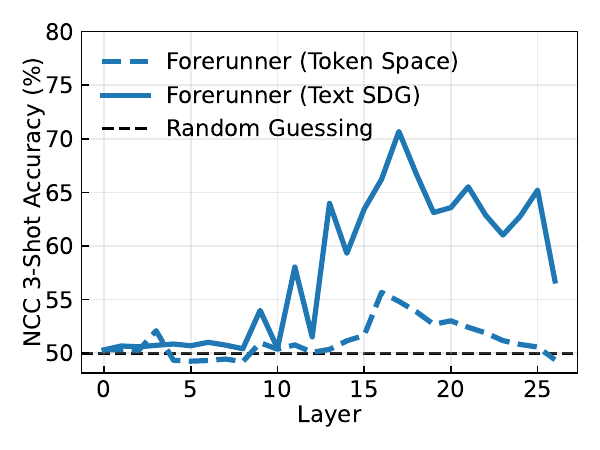}
    \caption{
        \textbf{Left:} Number of Dimensionality Reduction Heads (DR Heads) across layers, as well as the average attention to other images in-context.
        \textbf{Middle:} NCC Accuracy in the top-$k$ SDG dimensions after ablating DR Heads attention.
        \textbf{Right:} Forerunner NCC accuracy on SST-2.
    }
    \label{fig:dim_compression_ablation}
    \label{fig:sst_2vision_vs_forerunner}
    \vspace{-10pt}
\end{figure}

We first aim to select heads that might implement the dimensionality reduction, focusing on the OV circuit in~\cref{eq:OV_and_QK}.
For that, we first introduce the Bilinear Distortion Error (BDE) as 
\begin{equation}
\operatorname{BDE}(S,S')
:= \frac{1}{2}\mathbb{E}_{x,y\overset{\mathrm{i.i.d.}}{\sim}\mathcal{N}(0,I_D)}
\left[\left(x^\top S y-x^\top S' y\right)^2\right]
= \frac{1}{2}\|S-S'\|_F^2.
\label{eq:BDE}
\end{equation}
The BDE tests whether two geometries $S$ and $S'$ are the same by measuring whether they induce the same bilinear similarity for pairs of independently sampled inputs.
Thanks to the assumption that $x$ and $y$ follow standard Gaussian distributions, we can compute the error offline using only the two similarity matrices. Also we normalize each matrix $S$ to get a score bounded between $0$ and $1$.
A score of zero means that the two normalized geometries induce identical dot products.
\textbf{Now, if we hypothesize that the learned \VSDG{Vision SDG} describes the $Z$ subspace, a dimensionality reduction head should map $\Phi_{\ell}^{\mathrm v}$ via $W_{OV}$ to $\Phi_{\ell+1}^{\mathrm v}$.}
Thus, we define a score $s_{OV}$ that measures the expected error between the bilinear similarity of the output Vision SDG $\Phi_{\ell+1}^{\mathrm v}$ and the bilinear similarity of the input Vision SDG $\Phi_\ell^{\mathrm v}$ after being transported through $W_{OV}$ giving $
s_{OV}
=
\operatorname{BDE}\!\left(
\Phi_{\ell+1}^{\mathrm v\top}\Phi_{\ell+1}^{\mathrm v},
\bigl(\Phi_\ell^{\mathrm v}W_{OV}^{\top}\bigr)^\top
\bigl(\Phi_\ell^{\mathrm v}W_{OV}^{\top}\bigr)
\right).$
For the theoretical circuit in Eq.~(10), $s_{OV}=0$, as $\Phi_\ell=(0\;\;I_{D_\downarrow})$.
We classify the 2\% of heads with the lowest scores as dimensionality reduction heads.
See~\Cref{sec:vision_compression_heads} for additional analysis of dimensionality reduction heads and the choice of the score $s_{OV}$.

\cref{fig:dim_compression_ablation} (Left) shows that dimensionality reduction heads concentrate before \lstarprev. Notably, a high number of dimensionality reduction heads correlate with a high amount of attention to the other in-context images, although attention patterns are not part of the selection criterion. This is consistent with our theoretical circuit which reduces dimensionality through comparisons between in-context examples.
%
Next, we ablate attention from vision tokens to vision tokens in other in-context images for the dimensionality reduction heads.
For a fair comparison, we re-learn the Vision SDG with this attention ablated.
In addition, as a control, we do the same to an equal number of randomly selected heads before $\lstarprev$.
In \cref{fig:dim_compression_ablation} (Middle), we no longer observe the NCC accuracy boost in 10 dimensions when this attention is ablated.
Thus, ablating only attention from vision tokens to vision tokens in other in-context images, for just 2\% of the heads which match our theoretical OV circuits, impairs the dimensionality reduction. \textbf{This hints at the fact that LVLMs implement a circuit similar to our theoretical circuit.} 
Also, prior work observed empirically that sample-to-sample attention was beneficial to model performance which they call \emph{contextualization}~\citep{bakalova2025contextualizethenaggregate}.
Our results show dimensionality reduction as one mechanism explaining contextualization.

\section{SDG in other ICL tasks}

\paragraph{Text classification.}
A natural question raised by preceding observations is whether the \FSDG{Text SDG} is shared across modalities or specific to vision.
To test this, we evaluate the forerunner token NCC accuracy on a binary text classification task (SST-2 \citep{SST2}), using the same protocol.
We replace images with sequences of text tokens.
In \cref{fig:sst_2vision_vs_forerunner} (Right), at \lstar, the forerunner NCC accuracy is close to random before projection onto the Text SDG, but increases after projection. 
This suggests that the SDG is not tied only to the vision modality, even if it was trained only using visual inputs.


\begin{table}[t]
\centering
\small
\setlength{\tabcolsep}{2pt}
\begin{tabular}{@{}l|lll|llll@{}}
\toprule
& \multicolumn{3}{c}{Classification} & \multicolumn{4}{c}{Matching} \\
\cmidrule(lr){2-4} \cmidrule(lr){5-8}
 & Open MI & VLGuard & VizWiz & Matching MI & SugarCrepe & MHaluBench & NaturalBench \\
\midrule
Baseline & 84.0 & 63.0 & 72.0 & 81.0 & 57.5 & 75.0 & 61.0 \\
128D-SDG Ablation & \cellcolor{darkred!10!white}\textcolor{darkred}{-11.0} & \cellcolor{darkred!10!white}\textcolor{darkred}{-17.0} & \cellcolor{darkred!10!white}\textcolor{darkred}{-4.0} & \cellcolor{darkgreen!10!white}\textcolor{darkgreen}{+6.5} & \cellcolor{darkgreen!10!white}\textcolor{darkgreen}{+5.5} & \cellcolor{darkgreen!10!white}\textcolor{darkgreen}{+3.0} & \cellcolor{darkred!10!white}\textcolor{darkred}{-2.0} \\
128D-SDG Boost & \cellcolor{darkgreen!10!white}\textcolor{darkgreen}{+11.5} & \cellcolor{darkgreen!10!white}\textcolor{darkgreen}{+6.0} & \cellcolor{darkgreen!10!white}\textcolor{darkgreen}{+1.5} & \cellcolor{darkred!10!white}\textcolor{darkred}{-4.0} & \cellcolor{darkred!10!white}\textcolor{darkred}{-3.0} & \cellcolor{darkred!10!white}\textcolor{darkred}{-5.5} & \cellcolor{darkred!10!white}\textcolor{darkred}{-1.0} \\
\bottomrule
\end{tabular}
\caption{Impact of the Text SDG ablation and boost on multimodal benchmarks in the 2-shot setting.}
\vspace{-15pt}
\label{tab:other_tasks}
\end{table}

\paragraph{Other Multimodal Tasks.}
We also ask if our learned \FSDG{Text SDG} is involved in multimodal tasks other than image classification.
We ablate and boost the Text SDG at $\lstar$ as the model performs these tasks in-context.
Each task is categorized either as Classification or Matching.
Classification Tasks involve learning to classify ImageNet images (Open MI~\citep{zong2025VLICL}), if an image and text user query pair is harmful or not (VLGuard~\citep{zong2024VLGuard}) and if an image and question pair is answerable (VizWiz~\citep{gurari2018VizWiz}).
Matching Tasks involve learning if two images match (Matching MI~\citep{zong2025VLICL}) and if a text description matches an image (SugarCrepe~\citep{hsieh2023SugarCrepe}, MHaluBench~\citep{chen2024MHaluBench}, NaturalBench~\citep{li2024NaturalBench}).

The results in \cref{tab:other_tasks} show that for classification tasks, modifying the SDG has the same effect as in the image classification setup, meaning the SDG is involved in multimodal classification tasks.
For matching tasks, we observe the opposite effect: ablating the SDG improves performance.
Here, support examples indicate which task to perform, while the answer depends on matching inputs within the query rather than on their visual similarity to support examples. Ablating the SDG therefore helps the model by avoiding basing its answer on visual similarity to support examples.

\section{Conclusion}
In summary we have shown that the formation of linear image representations cannot be reduced to a simple transfer from input vision tokens to text tokens.
Remarkably, we find that across LVLM layers, the number of dimensions required to preserve linear class separability decreases. Therefore, we offer an explanation by combining a theoretical framework with empirical analysis.



\paragraph{Future Work.}
While vision makes this geometry easier to study, future work should test whether the same mechanism generalizes to LLMs and to other tasks beyond classification, and study how to learn additional geometries, and how they interact.
We think that \cref{tab:other_tasks} is an interesting direction for test-time modification and monitoring of what the model learns in-context.
Future work could also study our theoretical circuit in more detail.


\paragraph{Limitations.}
We show that one theoretical circuit can perform dimensionality reduction, but we do not directly identify the theoretical circuits within trained LVLMs. We only identify a similar circuit that might be more sophisticated.
Because our theoretical circuit assumes linear non-causal attention, it does not explain how LVLMs approximate it.
Moreover, using average pooling as an aggregation of all visual information for an image is a simplification that erases the more complex token-level interaction mechanisms.

\subsection*{AI use statement}

In this work, we used generative AI tools to aid and polish writing, support retrieval and discovery of related work, assistance in interpreting results, and assist in proving mathematical claims. We have not used generative AI tools for research ideation, drafting sections of the paper, or generating synthetic datasets.
Additionally, we used generative AI tools to assist with code generation, particularly for plotting and formatting, and carefully reviewed all generated code.
The convergence proof in Section~\ref{sec:convergence} was developed with some AI assistance and formally verified in Lean. An author carefully checked every step and revised its presentation to improve its usefulness and clarity for future readers.
We have reviewed all AI-assisted work.
We take responsibility for the final content of this work, including text, claims or artifacts produced with the aid of generative AI.

\subsection*{Reproducibility statement}

The code is included in the supplementary material for reproducibility and will be publicly released. Please follow the instructions in this repository to reproduce the experiments.
Training details and hyperparameters are provided in Sec.~\ref{sec:meta_training_details}, numerical experiment settings in Sec.~\ref{sec:numerical}, and evaluation protocols for other multimodal tasks and text classification in Sec.~\ref{sec:multimodal_task}.
All datasets used in this work are cited and publicly available for download.


%
%
\bibliographystyle{iclr2027_conference}
\bibliography{main}

\clearpage
\appendix

\section*{Supplementary Material Overview}

\begin{itemize}
    \item Sec.~\ref{sec:exp_setup}: Experimental Setup
    \begin{itemize}
        \item Sec.~\ref{sec:label_space}: Label Space
    \end{itemize}
    \item Sec.~\ref{sec:meta_training}: Meta-training
    \begin{itemize}
        \item Sec.~\ref{sec:meta_training_details}: Training Details
        \item Sec.~\ref{sec:meta_training_hparams}: Dimensionality
        \item Sec.~\ref{sec:meta_training_single_dataset}: Training on One Dataset Only
    \end{itemize}
    \item Sec.~\ref{sec:theory}: Theoretical Analysis
    \begin{itemize}
        \item Sec.~\ref{sec:convergence}: Convergence Proof
        \item Sec.~\ref{sec:numerical}: Numerical Validation
    \end{itemize}
    \item Sec.~\ref{sec:circuit_analysis}: Circuit Analysis
    \begin{itemize}
        \item Sec.~\ref{sec:more_ablation_exp}: More Ablation Experiments
        \item Sec.~\ref{sec:vision_forerunner_flow}: Analysis of Flow from Vision to Forerunner
        \item Sec.~\ref{sec:sdg_alignment}: SDG Alignment across Layers
        \item Sec.~\ref{sec:vision_compression_heads}: Analysis of Dimensionality Reduction Heads
    \end{itemize}
    \item Sec.~\ref{sec:more_models}: Other Models Experiments
    \item Sec.~\ref{sec:multimodal_task}: Other Tasks Details
\end{itemize}

\clearpage
\section{Experimental Setup}
\label{sec:exp_setup}

\subsection{Label Space}
\label{sec:label_space}

\begin{figure}[t]
    \centering
    \includegraphics[width=0.55\linewidth]{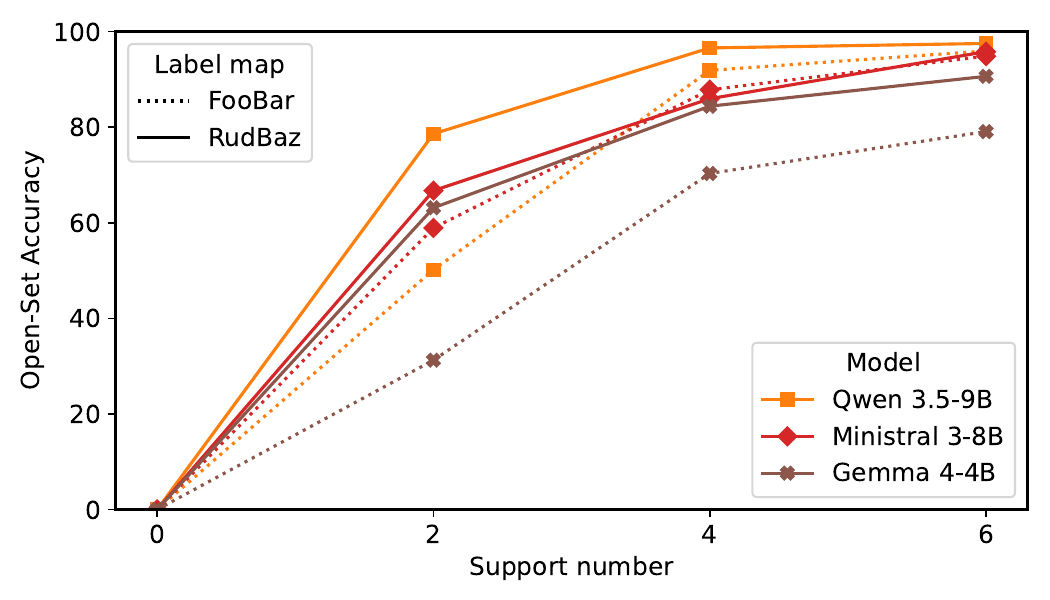}
    \caption{Model performance versus the total number of support examples (0, 2, 4, and 6, corresponding to 0, 1, 2, and 3 shots per class) for two label spaces in the 2-way setting.}
    \label{fig_supp:foobar_shots}
\end{figure}

\begin{wraptable}{l}{0.5\textwidth}
    \centering
    \vspace{-10pt}
    \setlength{\tabcolsep}{5pt}
    \begin{tabular}{lrr}
        \toprule
         & $\{\texttt{Foo}, \texttt{Bar}\}$ & $\{\texttt{Rud}, \texttt{Baz}\}$ \\
        \midrule
        Ministral 3 (8B) & 48.9 & 0.0 \\
        Qwen 3.5 (9B) & 23.3 & 0.0 \\
        \bottomrule
    \end{tabular}
    \caption{Probability (\%) assigned to the other class label at the second in-context example, given the first example's label (2-way, 1-shot setting).}
    \label{fig_supp:foobar_table}
    \vspace{-10pt}
\end{wraptable}

As stated in the main paper, choosing an appropriate semantically unaligned label space is important for removing language biases.
Standard practice uses the $\{\texttt{Foo}, \texttt{Bar}\}$ label space. However, $\{\texttt{Foo}, \texttt{Bar}\}$ has been widely used and may therefore appear in the pretraining data.
As shown in \cref{fig_supp:foobar_table}, the model assigns a non-zero probability to \texttt{Foo} after seeing \texttt{Bar} as a label, and vice versa. This behavior does not occur with $\{\texttt{Rud}, \texttt{Baz}\}$.
Furthermore, \cref{fig_supp:foobar_shots} shows that using $\{\texttt{Rud}, \texttt{Baz}\}$ improves model performance by removing this bias.
For these reasons, we choose $\{\texttt{Rud}, \texttt{Baz}\}$ as our label space.
For 3-way classification, we extend this label space with the additional label \texttt{Gip}.

\section{Meta-training}
\label{sec:meta_training}

\subsection{Training Details}
\label{sec:meta_training_details}

We train on 32,768 episodes sampled from the 10 datasets.
Each episode selects a dataset uniformly, then samples three
distinct classes uniformly from that dataset. 
Each episode contains three support images per class and one additional query image
whose class is selected uniformly among the three classes.
Support and query images are distinct within an episode.
The nine support examples are arranged into three consecutive
blocks, each containing one example per class in independently
shuffled class order. 
This arrangement lets us compute one-, two-, and three-shot classification losses from prefixes of three, six, and nine support examples, respectively, in a single forward pass.
Images are resized to $256 \times 256$
pixels using bicubic interpolation before applying the model's
pretrained image processor. No additional image augmentation
is applied.

Each hidden state is $\ell_2$-normalized before applying
$\Phi_\ell \in \mathbb{R}^{128 \times 3072}$.
This is mainly to stabilize the training, as certain LLM decoders exhibit high $\ell_2$-norm in the residual stream in last layers, this keeps training dynamic similar across layers.
The projections are initialized independently using the default
PyTorch linear-layer initialization, with weights sampled
uniformly from $[-1/\sqrt{3072},\,1/\sqrt{3072}]$.
The LVLM remains frozen throughout training.

We apply the loss after 3, 6, and 9 support images in-context to obtain more training signal from each task.
We use a batch size of 32 episodes, which stabilizes convergence given the high variance across tasks.
We optimize all projections jointly with AdamW using learning
rate $0.001$, weight decay $0.001$,
$(\beta_1,\beta_2)=(0.9,0.999)$, and $\epsilon=10^{-8}$.
We initially keep the learning rate fixed at 0.001, then gradually reduce it to zero using cosine decay over the final 324 training steps.

We use a single NVIDIA A100-SXM4 GPU with 80\,GB
memory and FlashAttention-2. The frozen LVLM forward pass used
bfloat16, while projection parameters and loss computations
used float32. The complete run took approximately
3 hours and 20 minutes.

\subsection{Dimensionality}
\label{sec:meta_training_hparams}

\begin{figure}[t]
    \centering
    \includegraphics[width=\textwidth]{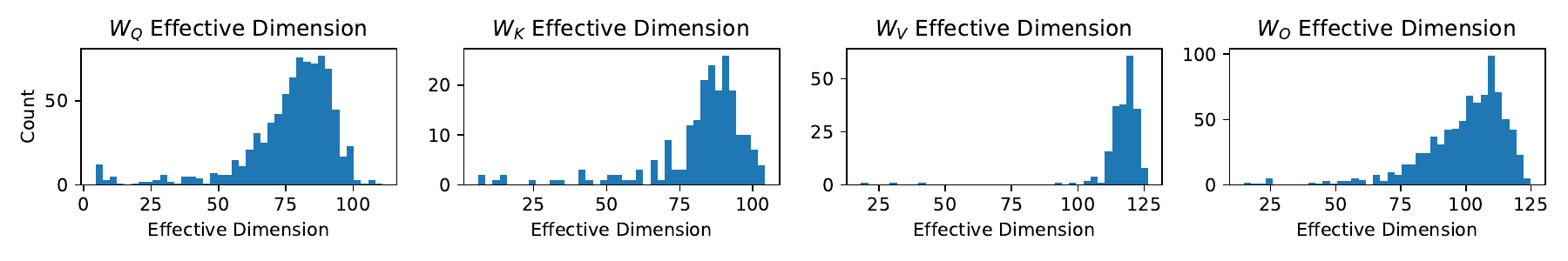}
    \vspace{-20pt}
    \caption{Histograms of effective dimension for the query ($W_Q$), key ($W_K$), value ($W_V$), and output ($W_O$) projection matrices of the model.}
    \label{fig:model_effective_dim}
\end{figure}

\begin{figure}[t]
    \centering
    \includegraphics[width=0.4\textwidth]{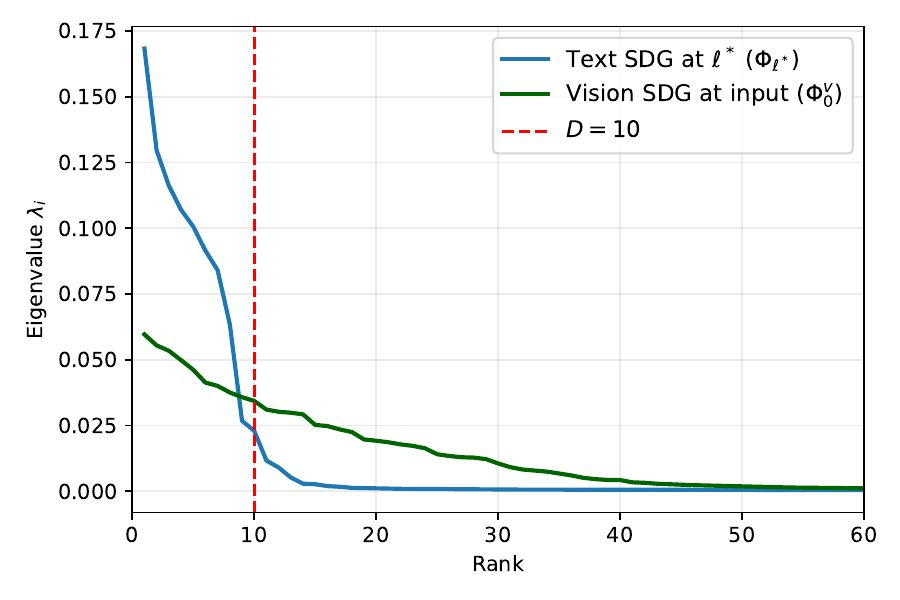}
    \caption{Ordered eigenvalues of $\Phi_{\lstar}^{\top}\Phi_{\lstar}$ (Text SDG, blue) and $(\Phi_0^{\mathrm v})^\top\Phi_0^{\mathrm v}$ (input Vision SDG, green). The Text SDG spectrum drops sharply around rank 10, with approximately 90\% of its eigenvalue mass in the top 10 directions; the input Vision SDG decays more gradually.}
    \label{fig:sdg_spectrum}
\end{figure}

We measure the effective dimension of each SDG projection $\Phi$ using the participation ratio~\citep{gao2017theory}, where $\lambda_i$ are the eigenvalues of the bilinear similarity matrix $\Phi^\top\Phi$:
\begin{equation}
    d_{\mathrm{eff}}(\Phi)
    =
    \frac{\left(\sum_i \lambda_i\right)^2}
         {\sum_i \lambda_i^2}.
\end{equation}
\Cref{fig:model_effective_dim} shows for comparison the distributions of effective dimension across the query, key, value, and output projections of the model.
Averaged over all heads and the four projection types, the effective dimension is $\sim 90$.
\Cref{fig:sdg_spectrum} compares the ordered eigenvalues of
$\Phi_{\lstar}^{\top}\Phi_{\lstar}$ (Text SDG at $\lstar$, blue)
and $(\Phi_0^{\mathrm v})^\top\Phi_0^{\mathrm v}$ (input Vision SDG, green).
The Text SDG spectrum drops sharply around rank 10, whereas the input Vision SDG
decays more gradually. The top 10 directions account for 91.06\% of the projected variance of the Text SDG at $\ell^*$, compared with 45.36\% for the input Vision SDG.

We test whether the low dimensionality of the SDG depends on the training setup by varying the number of classes and shots, and whether the loss is applied at intermediate support counts or only after the full support set.
As shown in \cref{tab:sdg_hparam_sensitivity}, removing the intermediate losses has little effect on the effective dimension of $\Phi_{\lstarprev}^{\mathrm v}$.
Moving from 3-way 3-shot (10 in-context images) to 4-way 4-shot (17 in-context images) training slightly increases it, but all four configurations yield an effective dimension close to 10.
Thus, the precise dimensionality varies with the training setup, but that fluctuation remains negligible.

\begin{table}[t]
    \centering
    \small
    \begin{tabular}{lcc}
        \toprule
        Training setup & Support counts used for the loss & Effective dimension of $\Phi_{\lstarprev}^{\mathrm v}$ \\
        \midrule
        3-way 3-shot & 3, 6, 9       & 9.69 \\
        3-way 3-shot & 9             & 9.68 \\
        4-way 4-shot & 4, 8, 12, 16  & 11.08 \\
        4-way 4-shot & 16            & 10.79 \\
        \bottomrule
    \end{tabular}
    \caption{Effective dimension of the Vision SDG at $\lstarprev$ under different training setups. Support counts indicate when the query classification loss is computed and exclude the query image.}
    \label{tab:sdg_hparam_sensitivity}
\end{table}

\clearpage
\subsection{Training on One Dataset Only}
\label{sec:meta_training_single_dataset}

\begin{figure*}[t]
    \centering
    \includegraphics[width=\linewidth]{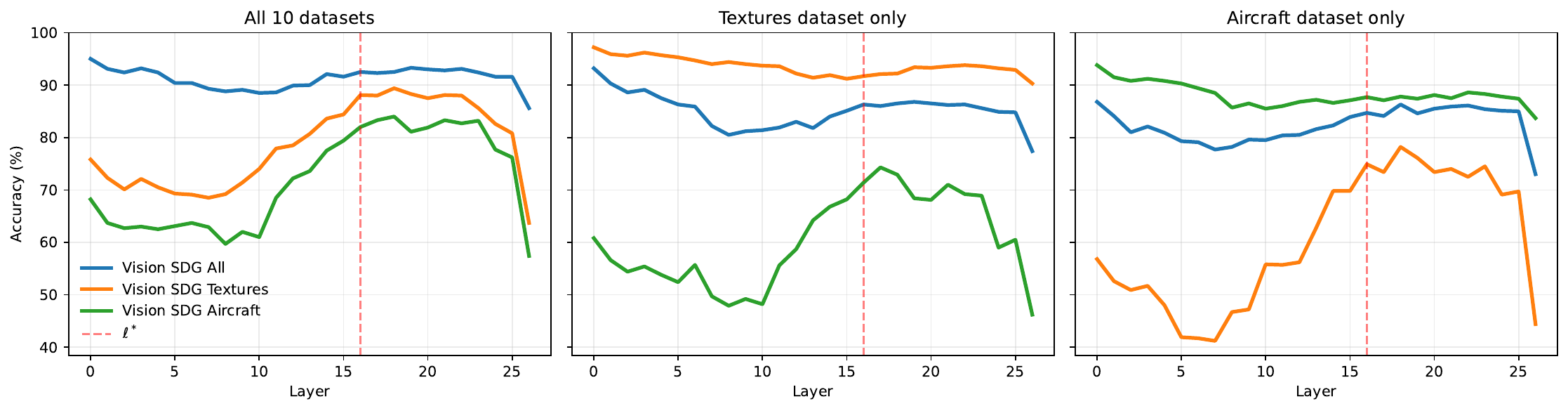}
    \caption{NCC accuracy of average-pooled vision tokens across layers. NCC Accuracy is computed in 3 different Vision SDG on 3 different datasets.
    \textbf{Left}: On all 10 datasets combined.
    \textbf{Middle}: On the Textures dataset.
    \textbf{Right}: On the Aircraft dataset.
    }
    \label{fig:texture_sdg}
\end{figure*}

\begin{wraptable}{l}{0.42\textwidth}
\vspace{-5pt}
\centering
\small
\setlength{\tabcolsep}{4pt}
\begin{tabular}{@{}lc@{}}
\toprule
Vision SDG & Effective dimension \\
training data & at $\lstarprev$ ($\Phi^{{\mathrm v}}_{\lstarprev}$) \\
\midrule
All 10 datasets & 9.69 \\
Textures only  & 9.71 \\
Aircraft only  & 6.57 \\
\bottomrule
\end{tabular}
\caption{Effective dimension of the Vision SDG at layer $\lstarprev$ when trained on all 10 datasets or on a single dataset.}
\vspace{-5pt}
\label{tab:single_dataset_sdg_dim}
\end{wraptable}
The shared property of the SDG of having a high accuracy on 10 different domains could result from training on those 10 domains. In order to challenge that property, we experiment with training the shared discriminative geometry only on 1 dataset instead of 10.
We follow the same training protocol in two separate experiments, training once on Aircraft alone and once on Textures alone.
We focus here only on the Vision SDG.
As shown in Table~\ref{tab:single_dataset_sdg_dim}, training on Textures alone yields nearly the same effective dimension as training on all 10 datasets (9.71 versus 9.69), showing that the approximately 10-dimensional Vision SDG can emerge from a single domain. Training on Aircraft alone yields a lower effective dimension (6.57), indicating that the learned geometry’s effective dimensionality depends on the training domain.

\begin{wrapfigure}{l}{0.4\textwidth}
    \centering
    \includegraphics[width=\linewidth]{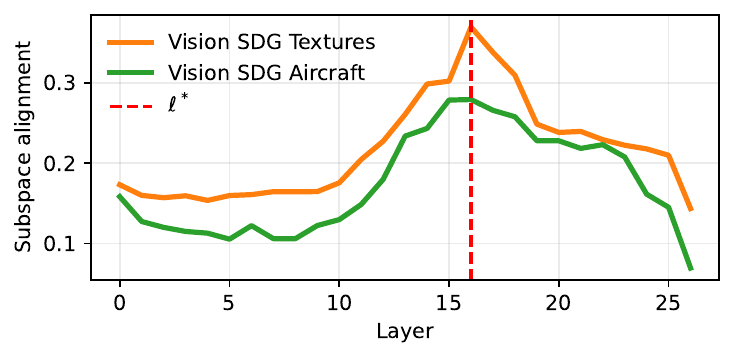}
    \vspace{-14pt}
    \caption{Layer-wise subspace alignment of the Textures and Aircraft
    Vision SDGs with the Vision SDG trained on all datasets.}
    \vspace{-10pt}
    \label{fig:sdg-subspace-alignment}
\end{wrapfigure}

\paragraph{Evidence of shared geometry.}
Next, we test whether a Vision SDG trained on a single dataset generalizes to other domains by comparing NCC accuracy across layers on Textures, Aircraft, and the full set of 10 datasets (Figure~\ref{fig:texture_sdg}).
Between layers 7 and 15, where dimensionality reduction occurs, we see that the accuracy of the Textures Vision SDG increases for Aircraft and vice versa. Similarly, the accuracy of the Textures Vision SDG and Aircraft Vision SDG increases on the full set of all 10 datasets in the similar range. This indicates that in those layers, discriminative directions get more common between tasks and domains. This presents an additional independent observation of the dimensionality reduction and of the shared property of the SDG between domains.
Additionally, we examine whether the Textures and Aircraft Vision SDGs converge toward the All datasets Vision SDG by measuring their layer-wise subspace alignment 
(using \(
1-\operatorname{BDE}\!\left(
(\Phi_{\ell,\mathrm{One}}^{\mathrm v})^\top\Phi_{\ell,\mathrm{One}}^{\mathrm v},
(\Phi_{\ell,\mathrm{All}}^{\mathrm v})^\top\Phi_{\ell,\mathrm{All}}^{\mathrm v}
\right)
\) from~\cref{eq:BDE})
(Figure~\ref{fig:sdg-subspace-alignment}).
Over the same layer range (7--15), both the Aircraft and Textures Vision SDGs become increasingly aligned with the Vision SDG trained on all 10 datasets.
This provides complementary geometric evidence that discriminative directions become increasingly shared across domains in these layers.

\paragraph{Evidence of domain-specific geometry.}
However, on Textures, the SDG trained on Textures achieves higher accuracy than the SDG trained on Aircraft, and vice versa on Aircraft. This means that part of the training is learning to select features that are domain-specific. Training on 10 different and diverse domains should in practice diminish that effect. \textbf{In conclusion, our training protocol produces an SDG that combines features shared across domains with features specific to its training domain.} Future work could investigate how to better disentangle both.

\section{Theoretical Analysis and Validation}
\label{sec:theory}

\subsection{Proof of the convergence rate}
\label{sec:convergence}

Let $n=M$ and $\lambda_1\geq\cdots\geq\lambda_n>0$ be the eigenvalues of $X^\top X$ and
\begin{equation}
\mathcal E(Z_\ell)
=
\frac14
\left\|
(Z_\ell)^\top Z_\ell-X^\top X
\right\|_F^2.
\end{equation}
Assume
\begin{equation}
D_\downarrow\geq n,
\qquad
\operatorname{rank}(Z_0)=n,
\qquad
0<\eta<
\frac{1}{
4\left[
\lambda_1+
\left\|(Z_0)^\top Z_0-X^\top X\right\|_F
\right]}.
\end{equation}
We prove that the dynamics
\begin{equation}
Z_{\ell+1}
=
Z_\ell
+
\eta Z_\ell
\left(
X^\top X-(Z_\ell)^\top Z_\ell
\right)
\end{equation}
satisfy
\begin{equation}
\mathcal E(Z_\ell)
=
O\!\left(
(1-2\eta\lambda_n)^{2\ell}
\right).
\end{equation}

Define the Gram-matrix error
\begin{equation}
E_\ell
=
Z_\ell^\top Z_\ell-X^\top X.
\end{equation}

\paragraph{Proof outline.}
The proof proceeds in three steps:
\begin{itemize}
    \item \textbf{Descent and boundedness.}
    We establish that $\mathcal E(Z_\ell)$ decreases, from which
    $Z_\ell E_\ell\to0$ and the boundedness of $(Z_\ell)_\ell$ follow.

    \item \textbf{Convergence to zero error.}
    Using the full-rank initialization and the determinant dynamics, we exclude
    convergence to a rank-deficient stationary point and conclude that $E_\ell\to0$.

    \item \textbf{Exponential convergence rate.}
    Once $E_\ell$ is sufficiently small, its dynamics consist of a linear
    contraction with rate $1-2\eta\lambda_n$ and a quadratic remainder, yielding
    $\mathcal E(Z_\ell)=O(\exp(-4\eta\lambda_n\ell))$.
\end{itemize}

\subsubsection{Descent and boundedness. $\|E_\ell\|_F\leq\|E_0\|_F$ and $Z_\ell E_\ell\longrightarrow0$}
The update is
\begin{equation}
Z_{\ell+1}
=
Z_\ell(I_n-\eta E_\ell).
\end{equation}
Consequently,
\begin{align}
E_{\ell+1}
&=
(I_n-\eta E_\ell)Z_\ell^\top Z_\ell(I_n-\eta E_\ell)-X^\top X
\\
&=
E_\ell
-\eta\left(E_\ell Z_\ell^\top Z_\ell+Z_\ell^\top Z_\ell E_\ell\right)
+\eta^2E_\ell Z_\ell^\top Z_\ell E_\ell,
\\
E_{\ell+1}-E_\ell
&=
-\eta\left(E_\ell Z_\ell^\top Z_\ell+Z_\ell^\top Z_\ell E_\ell\right)
+\eta^2E_\ell Z_\ell^\top Z_\ell E_\ell.
\label{eq:error_increment}
\end{align}

We first prove recursively that
\begin{equation}
\|E_\ell\|_F\leq\|E_0\|_F
\qquad
\text{for every }\ell.
\label{eq:error_bounded}
\end{equation}
Assume that this inequality holds at layer $\ell$. Then
\begin{align}
\|Z_\ell\|_2^2
&=
\|Z_\ell^\top Z_\ell\|_2
\label{eq:z_bound_spectral}\\
&=
\|X^\top X+E_\ell\|_2
\label{eq:z_bound_error}\\
&\leq
\lambda_1+\|E_\ell\|_2
\label{eq:z_bound_triangle}\\
&\leq
\lambda_1+\|E_0\|_F.
\label{eq:z_bound}
\end{align}
Equality \eqref{eq:z_bound_spectral} follows from
$\|Z_\ell\|_2^2=\lambda_{\max}(Z_\ell^\top Z_\ell)$.
Equality \eqref{eq:z_bound_error} follows from the definition of $E_\ell$.
Inequality \eqref{eq:z_bound_triangle} follows from the triangle inequality and
$\|X^\top X\|_2=\lambda_1$.
Finally, \eqref{eq:z_bound} follows from
$\|E_\ell\|_2\leq\|E_\ell\|_F\leq\|E_0\|_F$.

Taking the Frobenius inner product of
Eq.~\eqref{eq:error_increment} with $E_\ell$ gives
\begin{align}
\left\langle E_\ell,E_{\ell+1}-E_\ell\right\rangle
&=
\operatorname{tr}\left[E_\ell^\top(E_{\ell+1}-E_\ell)\right]
\\
&=
-\eta\operatorname{tr}\left(E_\ell^2Z_\ell^\top Z_\ell\right)
-\eta\operatorname{tr}\left(E_\ell Z_\ell^\top Z_\ell E_\ell\right)
+\eta^2\operatorname{tr}\left(E_\ell^2Z_\ell^\top Z_\ell E_\ell\right)
\\
&=
-2\eta\operatorname{tr}\left(Z_\ell^\top Z_\ell E_\ell^2\right)
+\eta^2\operatorname{tr}\left(Z_\ell^\top Z_\ell E_\ell^3\right),
\end{align}
where the last equality follows from the symmetry of $E_\ell$ and the cyclicity of the trace.
Since $E_\ell$ is symmetric,
\begin{equation}
-\|E_\ell\|_2E_\ell^2
\preceq
E_\ell^3
\preceq
\|E_\ell\|_2E_\ell^2.
\end{equation}
Because $Z_\ell^\top Z_\ell$ is positive semidefinite,
\begin{equation}
\left|
\operatorname{tr}
\left(
(Z_\ell)^\top Z_\ell E_\ell^3
\right)
\right|
\leq
\|E_\ell\|_2
\operatorname{tr}
\left(
(Z_\ell)^\top Z_\ell E_\ell^2
\right).
\end{equation}
Moreover \(
\operatorname{tr}
\left(
(Z_\ell)^\top Z_\ell E_\ell^2
\right)
=
\|Z_\ell E_\ell\|_F^2
\)
, therefore,
\begin{align}
\left\langle E_\ell,E_{\ell+1}-E_\ell\right\rangle
&\leq
-2\eta\|Z_\ell E_\ell\|_F^2
+\eta^2\|E_\ell\|_2\|Z_\ell E_\ell\|_F^2\\
&=
\left(-2\eta+\eta^2\|E_\ell\|_2\right)
\|Z_\ell E_\ell\|_F^2\\
&\leq
\left(-2\eta+\eta^2\|E_0\|_F\right)
\|Z_\ell E_\ell\|_F^2.
\label{eq:inner_product_bound}
\end{align}
The first inequality bounds the cubic trace term using
$\left|\operatorname{tr}((Z_\ell)^\top Z_\ell E_\ell^3)\right|
\leq \|E_\ell\|_2\|Z_\ell E_\ell\|_F^2$.
The second line factors out the common term $\|Z_\ell E_\ell\|_F^2$, while the final inequality uses
$\|E_\ell\|_2\leq\|E_\ell\|_F\leq\|E_0\|_F$.

The three terms in Eq.~\eqref{eq:error_increment} satisfy
\begin{align}
\text{\rm(i)}\quad
\left\|
E_\ell(Z_\ell)^\top Z_\ell
\right\|_F
&=
\left\|
(Z_\ell E_\ell)^\top Z_\ell
\right\|_F \notag\\
&\leq
\|Z_\ell E_\ell\|_F\|Z_\ell\|_2,\\
\text{\rm(ii)}\quad
\left\|
(Z_\ell)^\top Z_\ell E_\ell
\right\|_F
&\leq
\|Z_\ell\|_2\|Z_\ell E_\ell\|_F,\\
\text{\rm(iii)}\quad
\left\|
E_\ell(Z_\ell)^\top Z_\ell E_\ell
\right\|_F
&=
\left\|
(Z_\ell E_\ell)^\top(Z_\ell E_\ell)
\right\|_F \notag\\
&\leq
\|Z_\ell E_\ell\|_F
\|Z_\ell E_\ell\|_2 \notag\\
&\leq
\|E_0\|_F
\|Z_\ell\|_2
\|Z_\ell E_\ell\|_F.
\end{align}
We obtain
\begin{align}
\|E_{\ell+1}-E_\ell\|_F
&\leq
\eta\|Z_\ell\|_2
\left(
2+\eta\|E_0\|_F
\right)
\|Z_\ell E_\ell\|_F\\
&\leq
\eta
\sqrt{
\lambda_1+\|E_0\|_F
}
\left(
2+\eta\|E_0\|_F
\right)
\|Z_\ell E_\ell\|_F.
\label{eq:increment_bound}
\end{align}
The first inequality combines the bounds on the three terms of the error increment, while the second uses Eq.~\eqref{eq:z_bound} to eliminate the dependence on $\|Z_\ell\|_2$.

Since $\mathcal E(Z_\ell)=\frac14\|E_\ell\|_F^2$, using Eqs.~\eqref{eq:inner_product_bound} and~\eqref{eq:increment_bound},
\begin{align}
\mathcal E(Z_{\ell+1})-\mathcal E(Z_\ell)
&=
\frac12
\left\langle
E_\ell,E_{\ell+1}-E_\ell
\right\rangle
+
\frac14
\|E_{\ell+1}-E_\ell\|_F^2\\
&\leq
\left[
-\eta
+\frac{\eta^2\|E_0\|_F}{2}
+\frac{\eta^2
\left(\lambda_1+\|E_0\|_F\right)
\left(2+\eta\|E_0\|_F\right)^2}{4}
\right]
\|Z_\ell E_\ell\|_F^2\\
&=
-\eta
\left[
1
-\frac{\eta\|E_0\|_F}{2}
-\frac{
\eta
\left(
\lambda_1+\|E_0\|_F
\right)
\left(
2+\eta\|E_0\|_F
\right)^2
}{4}
\right]
\|Z_\ell E_\ell\|_F^2.
\end{align}
By the step-size assumption,
\begin{equation}
\eta
<
\frac{1}
{4\left(\lambda_1+\|E_0\|_F\right)}.
\end{equation}
Multiplying by $\lambda_1+\|E_0\|_F>0$ gives
\begin{equation}
\eta\left(\lambda_1+\|E_0\|_F\right)<\frac14.
\end{equation}
Moreover, since $\|E_0\|_F<\lambda_1+\|E_0\|_F$, we also have
\begin{equation}
\eta\|E_0\|_F
<
\eta\left(\lambda_1+\|E_0\|_F\right)
<
\frac14.
\end{equation}
Hence
\begin{align}
1
-\frac{\eta\|E_0\|_F}{2}
-\frac{
\eta
\left(
\lambda_1+\|E_0\|_F
\right)
\left(
2+\eta\|E_0\|_F
\right)^2
}{4}
&>
1-\frac18-\frac{81}{256}\\
&=
\frac{143}{256}
>
\frac12.
\end{align}
It follows that
\begin{equation}
\mathcal E(Z_{\ell+1})
\leq
\mathcal E(Z_\ell)
-
\frac{\eta}{2}
\|Z_\ell E_\ell\|_F^2.
\label{eq:loss_decrease}
\end{equation}
Thus $\mathcal E(Z_{\ell+1})\leq\mathcal E(Z_\ell)$, which proves
Eq.~\eqref{eq:error_bounded} recursively. Summing
Eq.~\eqref{eq:loss_decrease} from $\ell=0$ to $L$ gives
\begin{equation}
\begin{aligned}
\frac{\eta}{2}
\sum_{\ell=0}^{L}
\|Z_\ell E_\ell\|_F^2
&\leq
\sum_{\ell=0}^{L}
\left(
\mathcal E(Z_\ell)-\mathcal E(Z_{\ell+1})
\right)\\
&=
\mathcal E(Z_0)-\mathcal E(Z_{L+1})\\
&\leq
\mathcal E(Z_0),
\end{aligned}
\label{eq:summed_loss_decrease}
\end{equation}
where the second line follows by telescoping and the last inequality uses
$\mathcal E(Z_{L+1})\geq0$. Letting $L\to\infty$ in
Eq.~\eqref{eq:summed_loss_decrease} yields
\begin{equation}
\sum_{\ell=0}^{\infty}
\|Z_\ell E_\ell\|_F^2
<\infty.
\label{eq:gradient_summable}
\end{equation}
Since the summands in Eq.~\eqref{eq:gradient_summable} are nonnegative, they
must converge to zero. Therefore,
\begin{equation}
Z_\ell E_\ell\longrightarrow0.
\label{eq:gradient_zero}
\end{equation}

\subsubsection{Convergence to zero error. $\|E_\ell\|_F\longrightarrow0.$}
We next prove that the error must enter the local contraction region.
From Eq.~\eqref{eq:error_bounded},
\begin{align}
I_n-\eta E_\ell
&\succeq
\left(
1-\eta\|E_\ell\|_2
\right)I_n\\
&\succeq
\left(
1-\eta\|E_0\|_F
\right)I_n\\
&\succ
\frac34I_n.
\end{align}
Thus $I_n-\eta E_\ell$ is invertible. Since
\begin{equation}
Z_{\ell+1}=Z_\ell(I_n-\eta E_\ell),
\end{equation}
the full-rank assumption gives
\begin{equation}
\operatorname{rank}(Z_\ell)=n
\qquad
\text{for every finite }\ell.
\label{eq:rank_preserved}
\end{equation}

The sequence is also bounded because
\begin{align}
\|Z_\ell\|_F^2
&=
\operatorname{tr}
\left(
(Z_\ell)^\top Z_\ell
\right)\\
&\leq
n
\|(Z_\ell)^\top Z_\ell\|_2\\
&\leq
n
\left(
\lambda_1+\|E_0\|_F
\right).
\end{align}
Consider any convergent subsequence
\begin{equation}
Z_{\ell_k}\longrightarrow\overline Z.
\end{equation}
Equation~\eqref{eq:gradient_zero} and continuity give
\begin{equation}
\overline Z
\left(
\overline Z^\top\overline Z-X^\top X
\right)
=
0.
\end{equation}
Multiplication by $\overline Z^\top$ gives
\begin{equation}
\overline Z^\top\overline Z
\left(
\overline Z^\top\overline Z-X^\top X
\right)
=
0.
\label{eq:limit_equation}
\end{equation}
Taking the transpose gives
\begin{equation}
\left(
\overline Z^\top\overline Z-X^\top X
\right)
\overline Z^\top\overline Z
=
0.
\end{equation}
Consequently,
\begin{equation}
\overline Z^\top\overline Z\,X^\top X
=
\left(
\overline Z^\top\overline Z
\right)^2
=
X^\top X\,\overline Z^\top\overline Z.
\end{equation}
Thus the two symmetric matrices
$\overline Z^\top\overline Z$ and $X^\top X$ commute and have a common
orthonormal eigenbasis. In that basis,
\begin{equation}
X^\top X
=
\operatorname{diag}(\lambda_1,\ldots,\lambda_n),
\qquad
\overline Z^\top\overline Z
=
\operatorname{diag}(\gamma_1,\ldots,\gamma_n).
\end{equation}
Equation~\eqref{eq:limit_equation} then becomes
\begin{equation}
\gamma_i(\gamma_i-\lambda_i)=0
\qquad
\text{for every }i.
\end{equation}
Therefore,
\begin{equation}
\gamma_i=0
\qquad\text{or}\qquad
\gamma_i=\lambda_i.
\label{eq:limit_eigenvalues}
\end{equation}

Suppose, for contradiction, that
$\mathcal E(Z_\ell)$ converges to a positive value.
Every convergent subsequence then has a nonzero limiting error. By
Eq.~\eqref{eq:limit_eigenvalues}, at least one corresponding
$\gamma_i$ must be zero. At such a limit,
\begin{align}
\det
\left[
I_n-\eta
\left(
\overline Z^\top\overline Z-X^\top X
\right)
\right]
&=
\prod_{i=1}^n
\left[
1+\eta(\lambda_i-\gamma_i)
\right]\\
&=
\prod_{i:\gamma_i=0}
(1+\eta\lambda_i)\\
&\geq
1+\eta\lambda_n.
\label{eq:limit_determinant}
\end{align}

We now show explicitly that this bound must hold near every sufficiently
late iterate. Otherwise, infinitely many layers would satisfy
\begin{equation}
\det(I_n-\eta E_\ell)
\leq
1+\frac{\eta\lambda_n}{2}.
\end{equation}
The corresponding $Z_\ell$ form a bounded sequence, so they contain a
convergent subsequence. Taking the limit along that subsequence would give
\begin{equation}
\det
\left[
I_n-\eta
\left(
\overline Z^\top\overline Z-X^\top X
\right)
\right]
\leq
1+\frac{\eta\lambda_n}{2},
\end{equation}
which contradicts Eq.~\eqref{eq:limit_determinant}. Hence there exists a
finite $L$ such that
\begin{equation}
\det(I_n-\eta E_\ell)
>
1+\frac{\eta\lambda_n}{2}
\qquad
\text{for every }\ell\geq L.
\label{eq:determinant_growth}
\end{equation}

On the other hand,
\begin{align}
\det
\left(
(Z_{\ell+1})^\top Z_{\ell+1}
\right)
&=
\det
\left(
(Z_\ell)^\top Z_\ell
\right)
\det(I_n-\eta E_\ell)^2.
\end{align}
Equations~\eqref{eq:rank_preserved} and
\eqref{eq:determinant_growth} would therefore imply
\begin{equation}
\det
\left(
(Z_\ell)^\top Z_\ell
\right)
>
\det
\left(
(Z_L)^\top Z_L
\right)
\left(
1+\frac{\eta\lambda_n}{2}
\right)^{2(\ell-L)}.
\end{equation}
The right-hand side diverges, whereas Eq.~\eqref{eq:z_bound} gives
\begin{equation}
\det
\left(
(Z_\ell)^\top Z_\ell
\right)
\leq
\left(
\lambda_1+\|E_0\|_F
\right)^n.
\end{equation}
This contradiction proves
\begin{equation}
\|E_\ell\|_F\longrightarrow0.
\label{eq:error_zero}
\end{equation}

\subsubsection{Exponential convergence rate. $\mathcal E(Z_\ell)=O\!\left(
\exp(-4\eta\lambda_n\ell)
\right)$}

It remains to derive the exponential rate. We use the following consequence
of Ostrowski's fixed-point theorem~\cite{ARGYROS199977}. Let
$E_{\ell+1}=\mathcal F(E_\ell)$ converge to a fixed point $E^\star$.
If $\mathcal F$ is continuously differentiable near $E^\star$ and
\begin{equation}
\rho\!\left(D\mathcal F(E^\star)\right)<1,
\end{equation}
then, for every
\begin{equation}
\rho\!\left(D\mathcal F(E^\star)\right)<q<1,
\end{equation}
the iterates satisfy
\begin{equation}
\|E_\ell-E^\star\|_F=O(q^\ell).
\label{eq:ostrowski_rate}
\end{equation}

Consider the polynomial map
\begin{equation}
\mathcal F(E)
=
(I_n-\eta E)
\left(
X^\top X+E
\right)
(I_n-\eta E)
-
X^\top X.
\end{equation}
The error update satisfies $E_{\ell+1}=\mathcal F(E_\ell)$, and
$\mathcal F(0)=0$. Moreover,
\begin{equation}
D\mathcal F(0)[\Delta]
=
\Delta
-
\eta
\left(
\Delta X^\top X+X^\top X\Delta
\right).
\label{eq:error_jacobian}
\end{equation}

Let $v_i$ be an eigenvector of $X^\top X$ associated with $\lambda_i$.
Equation~\eqref{eq:error_jacobian} gives
\begin{equation}
D\mathcal F(0)
\left[
v_iv_j^\top
\right]
=
\left[
1-\eta(\lambda_i+\lambda_j)
\right]
v_iv_j^\top.
\end{equation}
Hence, the eigenvalues of $D\mathcal F(0)$ are
$1-\eta(\lambda_i+\lambda_j)$. The step-size assumption implies
\begin{equation}
\eta<\frac{1}{\lambda_1+\lambda_n},
\end{equation}
and therefore
\begin{equation}
-\left(1-2\eta\lambda_n\right)
<
1-2\eta\lambda_1
\leq
1-\eta(\lambda_i+\lambda_j)
\leq
1-2\eta\lambda_n.
\end{equation}
Since the upper bound is attained for $i=j=n$,
\begin{equation}
\rho\!\left(D\mathcal F(0)\right)
=
1-2\eta\lambda_n
<1.
\label{eq:error_jacobian_radius}
\end{equation}

By Eq.~\eqref{eq:error_zero}, $E_\ell\to0$. Let
$a_\ell=\|E_\ell\|_F$ and $\rho=1-2\eta\lambda_n$.
Since $\mathcal F$ is polynomial and
$\|D\mathcal F(0)\|_{F\to F}=\rho$, there is a constant $C>0$
such that, for all sufficiently large $\ell$,
\begin{equation}
a_{\ell+1}\leq \rho a_\ell+Ca_\ell^2.
\end{equation}
Equation~\eqref{eq:ostrowski_rate} gives $a_\ell=O(q^\ell)$
for any $\rho<q<1$, so $\sum_{\ell=0}^\infty a_\ell<\infty$.
For sufficiently large $L$, iterating the preceding inequality yields
\begin{align}
a_\ell
&\leq a_L\rho^{\ell-L}
\prod_{j=L}^{\ell-1}\left(1+\frac{C}{\rho}a_j\right)\\
&\leq a_L\rho^{\ell-L}
\exp\!\left(\frac{C}{\rho}\sum_{j=L}^{\infty}a_j\right)
=O(\rho^\ell).
\end{align}
Consequently,
\begin{equation}
\mathcal E(Z_\ell)
=\frac14a_\ell^2
=O\!\left((1-2\eta\lambda_n)^{2\ell}\right)
=O\!\left(\exp(-4\eta\lambda_n\ell)\right).
\end{equation}

\subsection{Numerical Experiments}
\label{sec:numerical}

We first verify the predicted convergence of the Gram error of the transformer in the idealized setting considered, then test its robustness to initialization, causal masking, normalization, and softmax attention.

\paragraph{Experimental setup.}
We initialize both $x_i \in \mathbb{R}^{D_\uparrow}$ and $z_{i,0} \in \mathbb{R}^{D_\downarrow}$ from standard Gaussian distributions then normalize, with $D_\uparrow=1024$, $D_\downarrow=32$ and $M=9$.
We run $16$ updates (layers) with $\eta=0.3$.
Since $D_\downarrow \geq M$, the compressed states can represent the input Gram matrix exactly.
We report the relative Gram error
\begin{equation}
    \varepsilon_\ell
    =
    \frac{\left\|Z_\ell^\top Z_\ell-X^\top X\right\|_{\mathrm{F}}^2}
    {\left\|X^\top X\right\|_{\mathrm{F}}^2}.
\end{equation}
Curves report the mean error over 30 seeds, and shaded regions indicate the 25th to 75th percentiles.

\begin{figure*}[t]
    \centering
    \begin{minipage}[t]{0.49\textwidth}
        \vspace{0pt}
        \begin{subfigure}[t]{\linewidth}
            \centering
            \includegraphics[width=\linewidth]{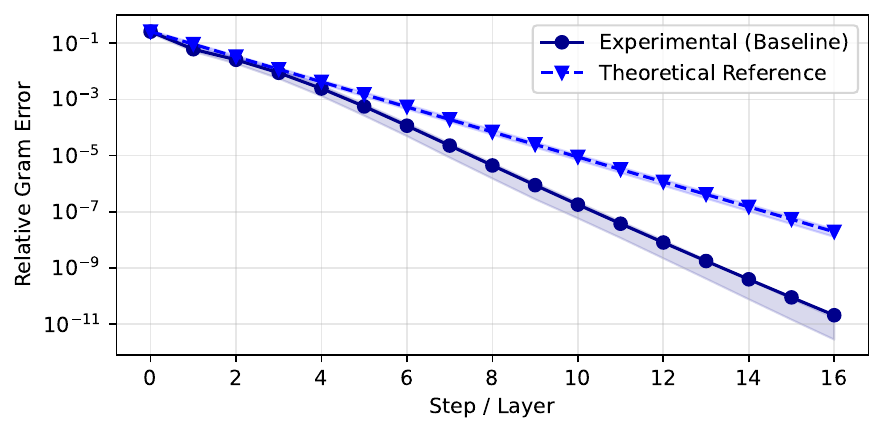}
            \caption{Baseline convergence and theoretical rate}
            \label{fig:gram_numerics_baseline}
        \end{subfigure}

        \vspace{0.6em}
        \begin{subfigure}[t]{\linewidth}
            \centering
            \includegraphics[width=\linewidth]{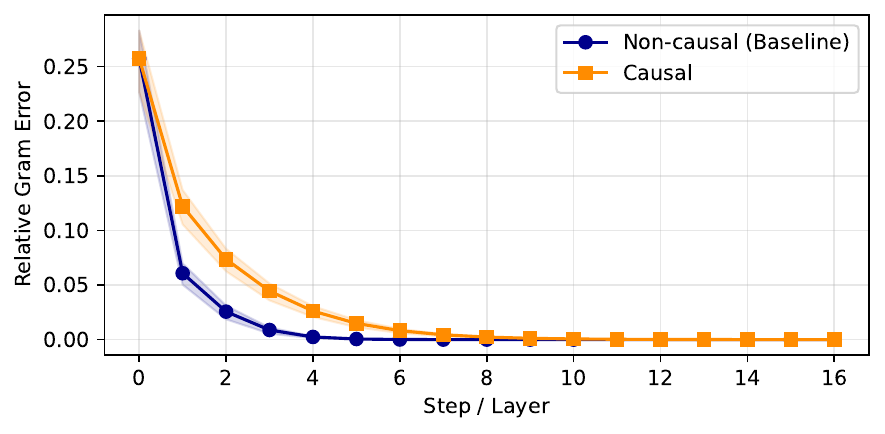}
            \caption{Causal masking}
            \label{fig:gram_numerics_causal}
        \end{subfigure}

        \vspace{0.6em}
        \begin{subfigure}[t]{\linewidth}
            \centering
            \includegraphics[width=\linewidth]{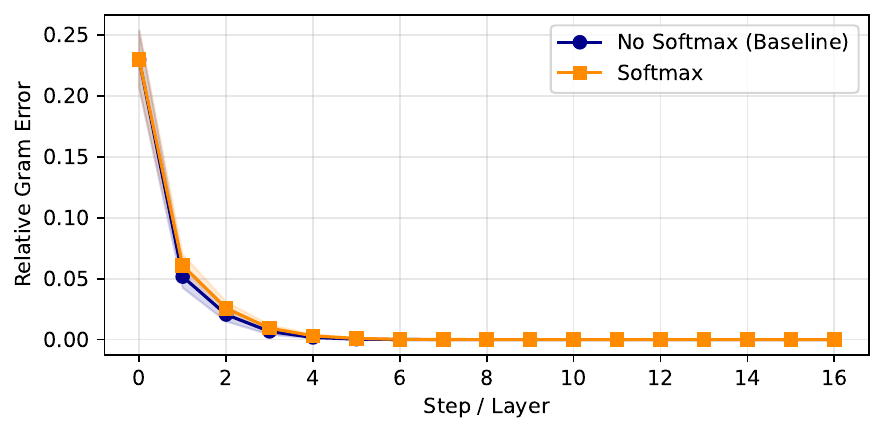}
            \caption{Softmax attention}
            \label{fig:gram_numerics_softmax}
        \end{subfigure}
        
        \vspace{0.6em}
        \begin{subfigure}[t]{\linewidth}
            \centering
            \includegraphics[width=\linewidth]{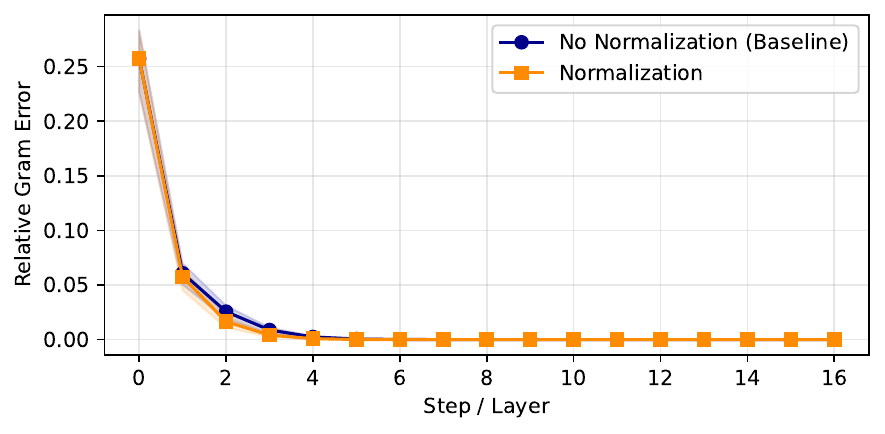}
            \caption{Normalization}
            \label{fig:gram_numerics_normalization}
        \end{subfigure}
    \end{minipage}
    \hfill
    \begin{minipage}[t]{0.49\textwidth}
        \vspace{0pt}
        \begin{subfigure}[t]{\linewidth}
            \centering
            \includegraphics[width=\linewidth]{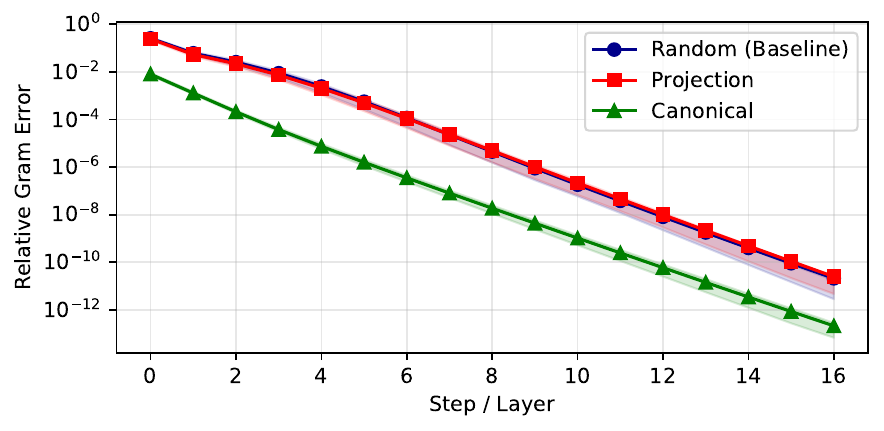}
            \caption{Initialization}
            \label{fig:gram_numerics_initialization}
        \end{subfigure}

        \vspace{0.6em}
        \begin{subfigure}[t]{\linewidth}
            \centering
            \includegraphics[width=\linewidth]{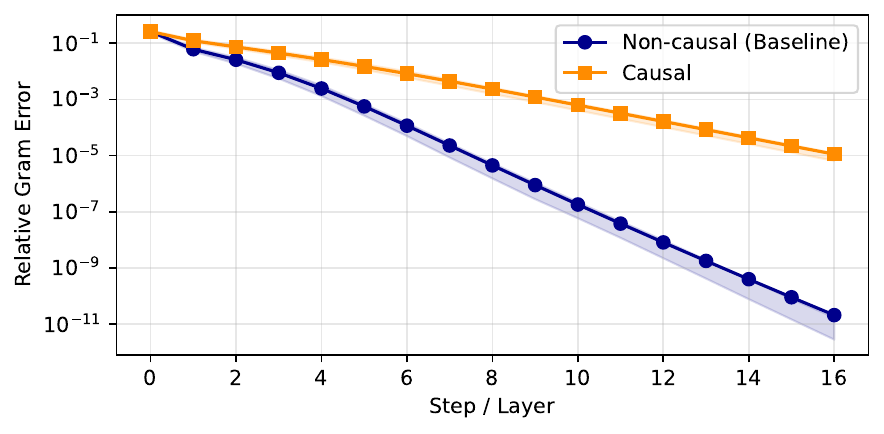}
            \caption{Causal masking (log scale)}
            \label{fig:gram_numerics_causal_log}
        \end{subfigure}

        \vspace{0.6em}
        \begin{subfigure}[t]{\linewidth}
            \centering
            \includegraphics[width=\linewidth]{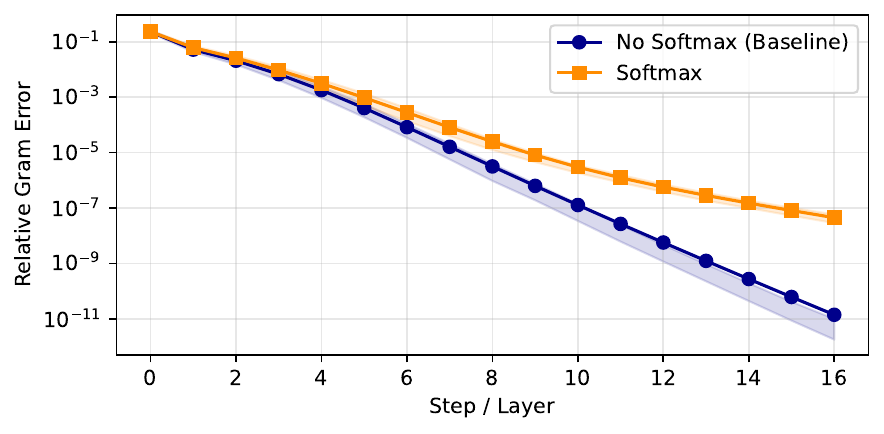}
            \caption{Softmax attention (log scale)}
            \label{fig:gram_numerics_softmax_log}
        \end{subfigure}
        
        \vspace{0.6em}
        \begin{subfigure}[t]{\linewidth}
            \centering
            \includegraphics[width=\linewidth]{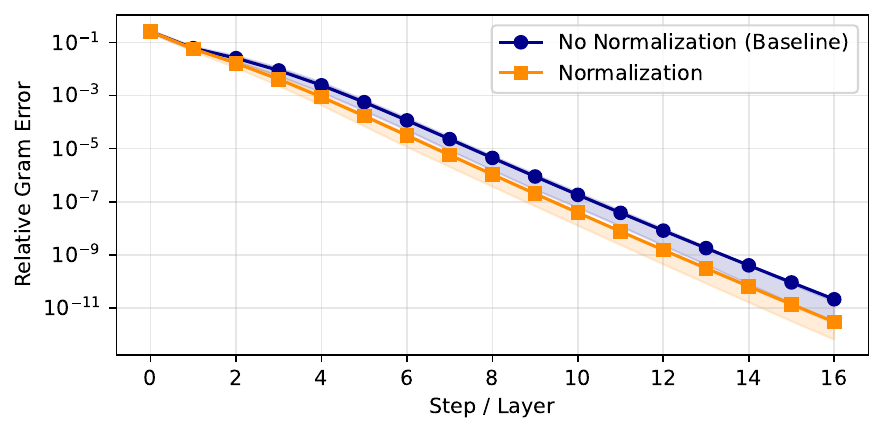}
            \caption{Normalization (log scale)}
            \label{fig:gram_numerics_normalization_log}
        \end{subfigure}
    \end{minipage}
    \caption{
    Numerical evaluation of Relative Gram Error under different initialization and attention settings.
    }
    \label{fig:gram_numerics}
\end{figure*}

\paragraph{Baseline.}
We first evaluate the transformer in the idealized setting considered in our theoretical analysis: non-causal linear self-attention, without softmax or normalization. 
Attention weights are initialized using \cref{eq:OV_and_QK}.
The relative Gram error decreases exponentially, as shown in \cref{fig:gram_numerics_baseline}.
For comparison, we plot the asymptotic decay rate predicted by \cref{eq:main_convergence}, anchored at $\varepsilon_0$,
\begin{equation}
    \varepsilon_\ell^{\mathrm{ref}}
    =
    \varepsilon_0
    \exp\!\left(-4\eta\lambda_M \ell\right).
\end{equation}
As expected, the empirical error decreases exponentially and follows the asymptotic decay.

\paragraph{Assumption 1: Causal attention.}
We add only an attention mask, removing all interactions from future examples while keeping the same weights.
As shown in \cref{fig:gram_numerics_causal}, causality slows the convergence rate.
Despite this, the relative Gram error decreases steadily even though the first example cannot attend to subsequent examples.

\paragraph{Assumption 2.a: Normalization.}
We test the effect of normalizing every $z_{i,\ell}$ to unit norm before each update.
As shown in \cref{fig:gram_numerics_normalization}, normalization has a negligible impact on convergence.

\paragraph{Assumption 2.b: Softmax attention.}
We apply softmax to the attention weights.
Softmax produces a weighted average, since its weights sum to one, whereas linear attention sums over the $M$ examples. We therefore multiply $W_V$ by $M=9$ to match the scale of the linear update.
Additionally, before the first update, we center $X$ and $Z_0$. It prevents softmax attention from continuing to change the compressed vectors after the Gram matrices already match. 
As shown in \cref{fig:gram_numerics_softmax}, softmax has a negligible impact on convergence under those additional assumptions.

\paragraph{Assumption 3: $Z$ initialization.}
We compare $Z_0$ Gaussian initialization, random projection $Z_0=W_{rand}^\top X$, and canonical initialization $Z_0=[I_M;0]$.
All three initializations satisfy the convergence assumptions ($rank(Z_0)=M$) and converge, as shown in \cref{fig:gram_numerics_initialization}.
Gaussian and projection initializations behave similarly.
Canonical initialization yields an error approximately ten times smaller.

\paragraph{Summary.}
\textbf{As shown in \cref{fig:gram_numerics}, in this setup with a $32\times$ dimensionality reduction, all tested settings reach a relative Gram error below $10^{-5}$ after all transformer layers.}

\section{Circuit Analysis}
\label{sec:circuit_analysis}

\subsection{More Ablation Experiments}
\label{sec:more_ablation_exp}

\subsubsection{Ablating the Top SDG Dimensions}

\begin{table}[t]
\centering
\small
\setlength{\tabcolsep}{3pt}
\begin{tabular}{clccccc}
\toprule
& Intervention & 1-shot & 2-shot & 3-shot & 4-shot & 5-shot \\
\midrule
\multicolumn{2}{l}{LVLM Accuracy} & 34.4\% & 77.2\% & 82.4\% & 84.2\% & 85.4\% \\
\midrule
\multirow{3}{*}{\rotatebox[origin=c]{90}{All}} & 1D to 128D - Control Ablation & \textbf{\textcolor{black}{-0.4}} & \textbf{\textcolor{black}{-1.2}} & \textbf{\textcolor{black}{-1.9}} & \textbf{\textcolor{black}{-2.5}} & \textbf{\textcolor{black}{-2.2}} \\
 & 1D to 15D - SDG Ablation & \textbf{\textcolor{black}{-7.1}} & \textbf{\textcolor{black}{-10.8}} & \textbf{\textcolor{black}{-12.9}} & \textbf{\textcolor{black}{-12.0}} & \textbf{\textcolor{black}{-10.6}} \\
 & 1D to 128D - SDG Ablation & \textbf{\textcolor{black}{-13.3}} & \textbf{\textcolor{black}{-11.3}} & \textbf{\textcolor{black}{-11.4}} & \textbf{\textcolor{black}{-11.7}} & \textbf{\textcolor{black}{-9.7}} \\
 & 16D to 128D - SDG Ablation & \textbf{\textcolor{black}{-5.7}} & \textbf{\textcolor{black}{+1.8}} & \textbf{\textcolor{black}{+1.6}} & \textbf{\textcolor{black}{+1.4}} & \textbf{\textcolor{black}{+0.5}} \\
 & 1D to 15D - SDG Boost & \textbf{\textcolor{black}{+7.5}} & \textbf{\textcolor{black}{+4.7}} & \textbf{\textcolor{black}{+3.9}} & \textbf{\textcolor{black}{+3.0}} & \textbf{\textcolor{black}{+2.9}} \\
 & 1D to 128D - SDG Boost & \textbf{\textcolor{black}{+12.8}} & \textbf{\textcolor{black}{+3.3}} & \textbf{\textcolor{black}{+1.5}} & \textbf{\textcolor{black}{+1.4}} & \textbf{\textcolor{black}{+1.3}} \\
 & 16D to 128D - SDG Boost & \textbf{\textcolor{black}{+6.9}} & \textbf{\textcolor{black}{-1.8}} & \textbf{\textcolor{black}{-3.0}} & \textbf{\textcolor{black}{-3.2}} & \textbf{\textcolor{black}{-2.9}} \\
\midrule
\multirow{3}{*}{\rotatebox[origin=c]{90}{Forerunner}} & 1D to 128D - Control Ablation & \textbf{\textcolor{black}{-1.9}} & \textbf{\textcolor{black}{+0.8}} & \textbf{\textcolor{black}{-0.0}} & \textbf{\textcolor{black}{-0.1}} & \textbf{\textcolor{black}{-0.1}} \\
 & 1D to 15D - SDG Ablation & \textbf{\textcolor{black}{-5.0}} & \textbf{\textcolor{black}{-1.0}} & \textbf{\textcolor{black}{-0.9}} & \textbf{\textcolor{black}{-1.5}} & \textbf{\textcolor{black}{-1.1}} \\
 & 1D to 128D - SDG Ablation & \textbf{\textcolor{black}{-9.9}} & \textbf{\textcolor{black}{-0.5}} & \textbf{\textcolor{black}{-0.5}} & \textbf{\textcolor{black}{-0.2}} & \textbf{\textcolor{black}{-0.6}} \\
 & 16D to 128D - SDG Ablation & \textbf{\textcolor{black}{-5.6}} & \textbf{\textcolor{black}{+1.1}} & \textbf{\textcolor{black}{+0.3}} & \textbf{\textcolor{black}{+0.6}} & \textbf{\textcolor{black}{+0.4}} \\
 & 1D to 15D - SDG Boost & \textbf{\textcolor{black}{+4.2}} & \textbf{\textcolor{black}{+2.2}} & \textbf{\textcolor{black}{+0.8}} & \textbf{\textcolor{black}{+0.8}} & \textbf{\textcolor{black}{+0.4}} \\
 & 1D to 128D - SDG Boost & \textbf{\textcolor{black}{+8.9}} & \textbf{\textcolor{black}{+2.1}} & \textbf{\textcolor{black}{+0.5}} & \textbf{\textcolor{black}{+0.3}} & \textbf{\textcolor{black}{+0.2}} \\
 & 16D to 128D - SDG Boost & \textbf{\textcolor{black}{+4.5}} & \textbf{\textcolor{black}{+0.0}} & \textbf{\textcolor{black}{-0.4}} & \textbf{\textcolor{black}{-1.5}} & \textbf{\textcolor{black}{-1.2}} \\
\bottomrule
\end{tabular}
\caption{Effect of top SDG interventions on model classification performance.
Baseline accuracy is reported in percent; intervention effects are changes in percentage points.}
\label{tab:sdg_interventions_supp}
\end{table}

At layer $\ell^*$, ablation removes the component of a token’s hidden state along selected Text SDG directions, while boosting doubles that component. Table~\ref{tab:sdg_interventions_supp} applies both interventions to the top 15 directions, all 128 directions, and the remaining directions after excluding the top 15. For each choice, we intervene either on all tokens or only on the forerunner token, measure the change in classification accuracy across one to five shots, and include ablation of a random 128-dimensional subspace as a control. 
Notably, ablating the SDG only from the forerunner tokens has less effect. This may be due to later layers copying the SDG from other text tokens back to the forerunner token.
For 2 to 5 shots, ablating all 128 dimensions has less effect than ablating only the top 15 dimensions. 
This confirms that the geometry used for classification is low-dimensional.
Ablating dimensions 16 to 128 slightly improves accuracy for 2 to 5 shots, while boosting them reduces it, suggesting a small opposing effect.

\subsubsection{Ablation impact on heads accuracy}
    
\begin{figure*}[t]
    \centering
    \begin{subfigure}{0.27\linewidth}
        \centering
        \includegraphics[width=\linewidth]{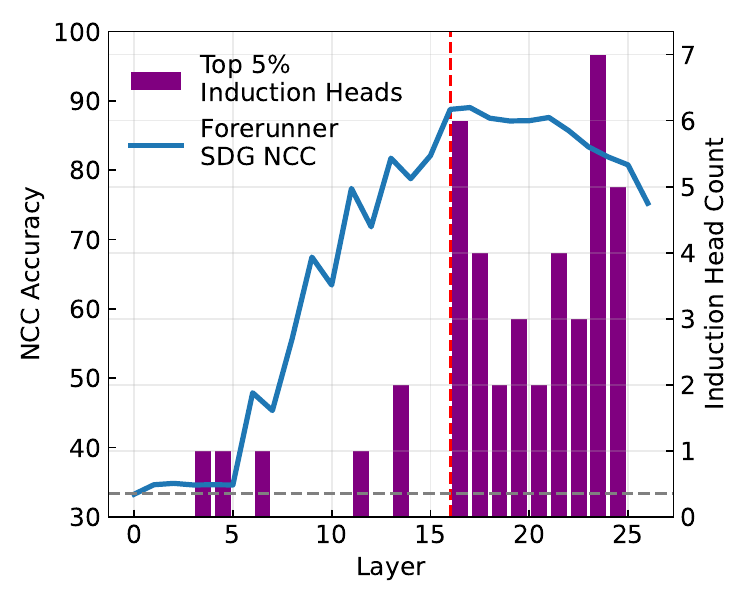}
        \label{fig:ih_count}
    \end{subfigure}
    \begin{subfigure}{0.34\linewidth}
        \centering
        \includegraphics[width=\linewidth]{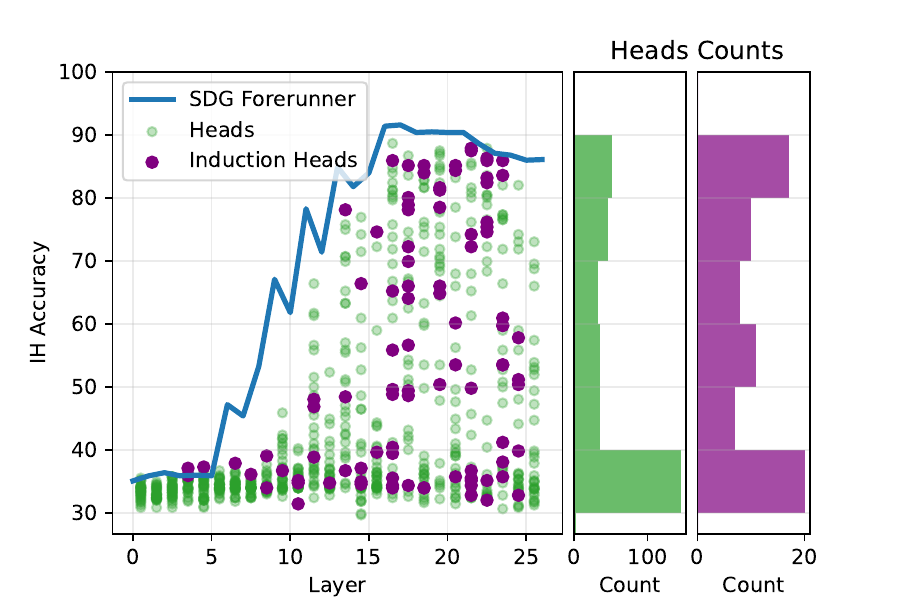}
        \label{fig:head_accuracy}
    \end{subfigure}
    \begin{subfigure}{0.34\linewidth}
        \centering
        \includegraphics[width=\linewidth]{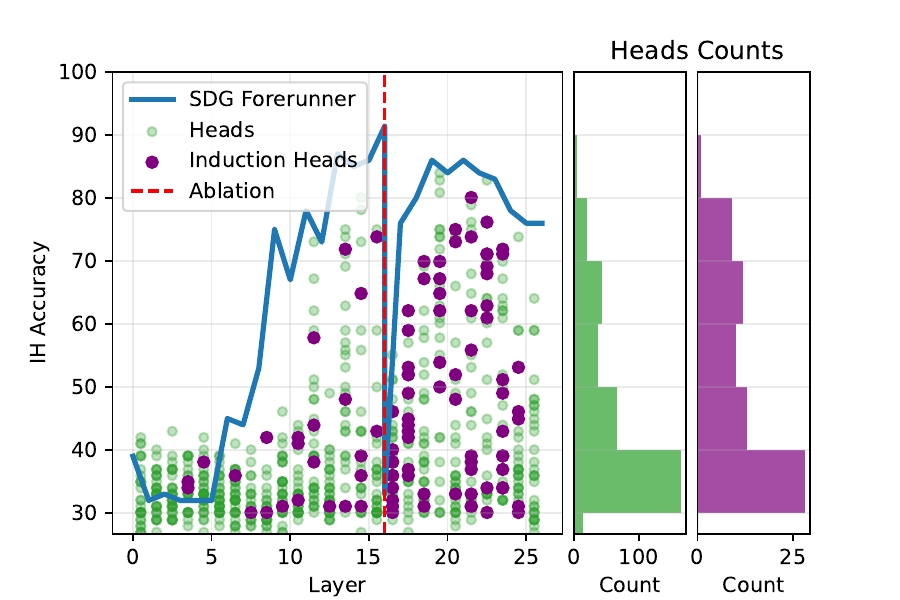}
        \label{fig:head_accuracy_ablated}
    \end{subfigure}
    \vspace{-10pt}
    \caption{
    \textbf{Left:} Number of induction heads across layers. 
    \textbf{Middle:} Head accuracy across layers, with head count per accuracy range.
    \textbf{Right:} Same as the middle after ablating the SDG at \lstar.
    }
    \label{fig:many_heads}
\end{figure*}

A possible reason for having a low dimensionality is allowing many heads to query the geometry without costing too much in parameters. Each head measures closeness of a query to support in-context through {bilinear similarity} in its learned query-key space $x_q^\top W_Q^\top W_K x_i$. 
If SDG dimensionality was 128, then an accurate {sample-to-sample} comparison would require all the heads' parameters, making the head able to only do that comparison without leaving capacity for other computations.
Therefore we investigate if many heads use the SDG. 
First we focus our attention on induction heads~\citep{olsson2022incontextlearninginductionheads}.
In short, those heads copy the label of support examples that are closest to the query.
We count a head as an induction head if it is in the top 5\% of heads attending to the support label tokens when predicting the query label following~\cite{cho2025revisiting}.
\Cref{fig:many_heads} (Left)
shows the number of induction heads per layer. 
We see that just after \lstar, there is a sudden increase in the number of induction heads. 
This shows induction starts only after the SDG is fully formed. 
We next ask whether the SDG is used by induction heads and, more broadly, across many attention heads.
\Cref{fig:many_heads} (Middle)
reports, for each head, the accuracy obtained by assigning the query to the class of supports receiving the largest amount of attention from the Forerunner Token.
Simply put, it tells us how accurately any given head attends to support examples of the same class.
While the SDG is the upper bound on the accuracy a head can have, we see many heads having a high accuracy close to it.
After its formation, 14.77\% of all heads and 23.29\% of induction heads have an accuracy above 80\%.
\Cref{fig:many_heads} (Right)
shows heads accuracy after ablating the SDG at layer \lstar.
Accuracy drops across heads with only 1.42\% of all heads and 1.37\% of induction heads having an accuracy above 80\%.
\textbf{This indicates that the SDG is used by both many heads and many induction heads.}
This supports the hypothesis that having lower-dimensional geometries helps many heads do better sample-to-sample comparisons.

\subsection{Analysis of Flow from Vision to Forerunner}
\label{sec:vision_forerunner_flow}

\begin{figure*}[t]
    \centering
    \begin{minipage}[t]{0.245\linewidth}
        \centering
        \adjincludegraphics[trim={{0.1\width} 0 {0.1\width} 0},clip,width=\linewidth]{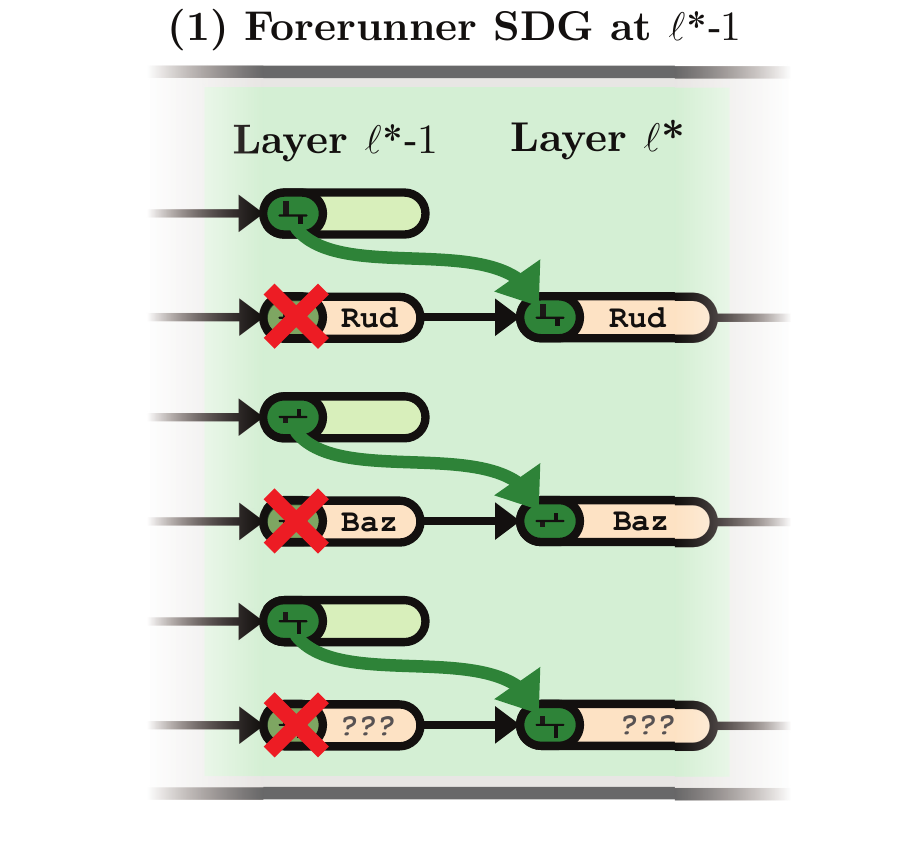}
        \par\smallskip
        Effect: $-3.2\%$
    \end{minipage}\hfill
    \begin{minipage}[t]{0.245\linewidth}
        \centering
        \adjincludegraphics[trim={{0.1\width} 0 {0.1\width} 0},clip,width=\linewidth]{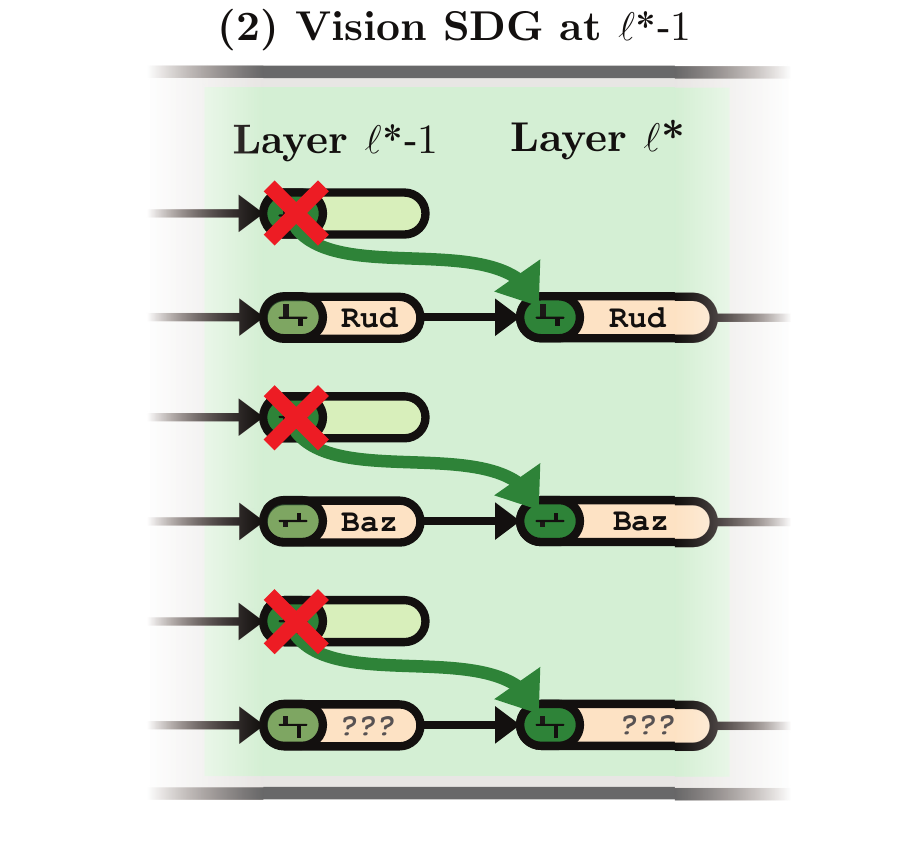}
        \par\smallskip
        Effect: $-6.4\%$
    \end{minipage}\hfill
    \begin{minipage}[t]{0.245\linewidth}
        \centering
        \adjincludegraphics[trim={{0.1\width} 0 {0.1\width} 0},clip,width=\linewidth]{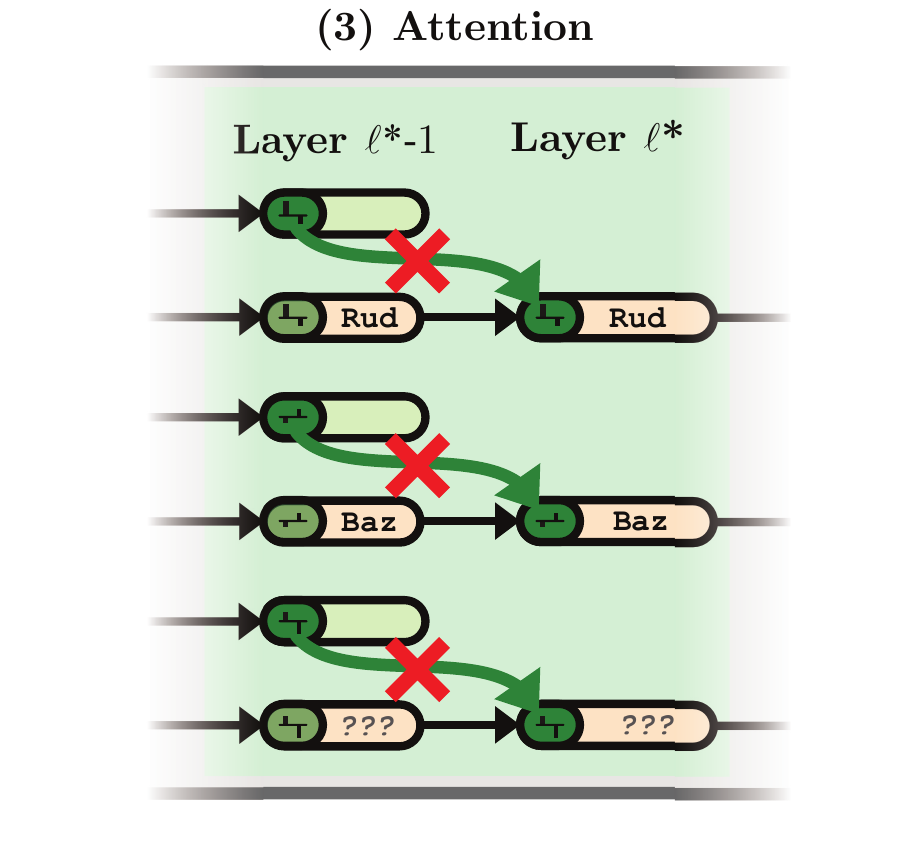}
        \par\smallskip
        Effect: $-5.2\%$
    \end{minipage}\hfill
    \begin{minipage}[t]{0.245\linewidth}
        \centering
        \adjincludegraphics[trim={{0.1\width} 0 {0.1\width} 0},clip,width=\linewidth]{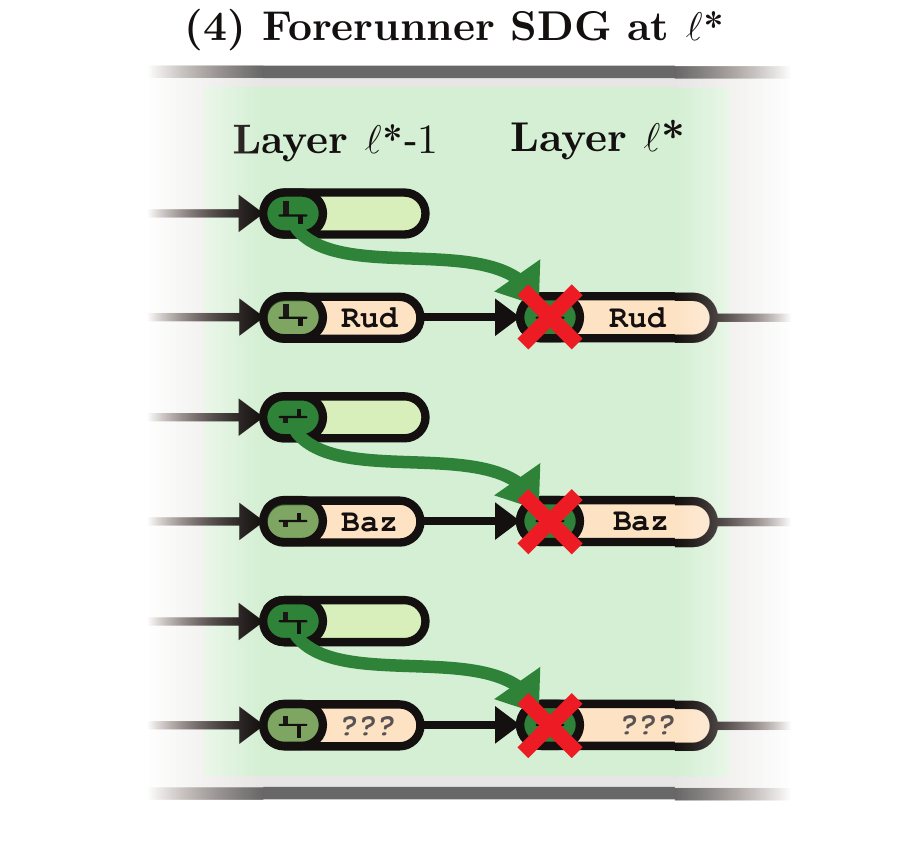}
        \par\smallskip
        Effect: $-11.4\%$
    \end{minipage}
    \caption{Schematization of the ablation performed in \cref{tab:vision_attention_ablation}.}
    \label{fig:schema_ablation_vision}
\end{figure*}

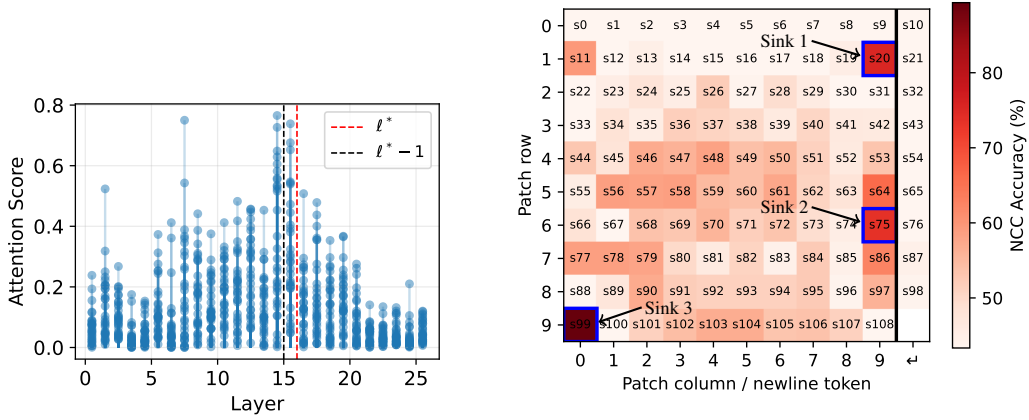
\begin{figure}[t]
    \centering
    \includegraphics[width=0.44\linewidth]{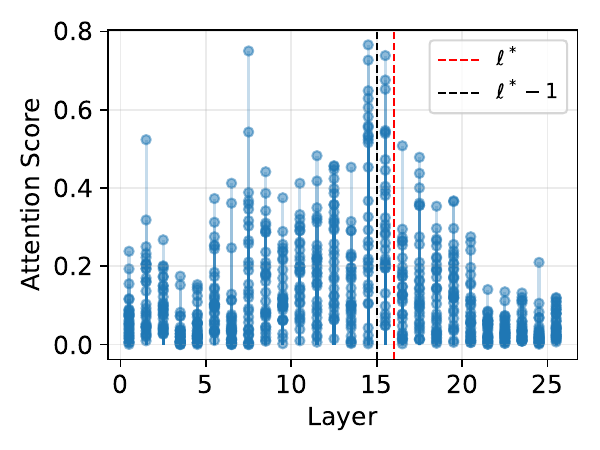}
    \hfill
    \begin{tikzpicture}
        \node[anchor=south west, inner sep=0] (img) at (0,0) {
            \includegraphics[width=0.52\linewidth]{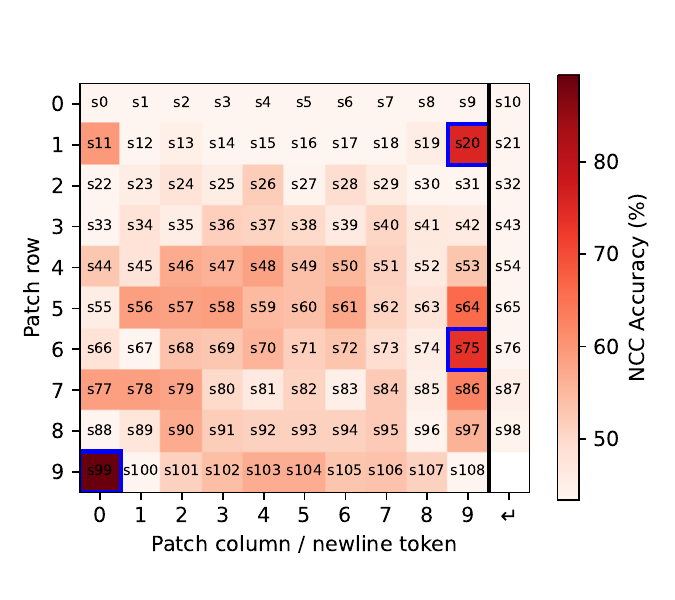}
        };
        \begin{scope}[x={(img.south east)}, y={(img.north west)}]
            \node[font=\scriptsize, inner sep=1pt] (sink1)
                at (0.52,0.795) {Sink 1};
            \draw[->, thick] (sink1.east) -- (0.658,0.765);
    
            \node[font=\scriptsize, inner sep=1pt] (sink2)
                at (0.52,0.455) {Sink 2};
            \draw[->, thick] (sink2.east) -- (0.658,0.425);
    
            \node[font=\scriptsize, inner sep=1pt] (sink3)
                at (0.31,0.250) {Sink 3};
            \draw[->, thick] (sink3.west) -- (0.182,0.220);
        \end{scope}
    \end{tikzpicture}
    \caption{
        \textbf{Left:} Average amount of attention of each head across layers. Attention from the forerunner token to the query image.
        \textbf{Right:} NCC accuracy at layer 15 per token position after projecting onto the Vision SDG.
        Columns 0-9 are image patches; the rightmost column contains newline tokens. Blue boxes mark the three patch-token sinks.
    }
    \label{fig:attn_prev_score}
\end{figure}

\begin{figure*}[t]
  \centering
  \setlength{\tabcolsep}{3pt}
  \renewcommand{\arraystretch}{1.15}
  \begin{tabular}{
      >{\centering\arraybackslash}m{0.03\textwidth}
      *{3}{>{\centering\arraybackslash}m{0.28\textwidth}}
    }
    & \textbf{Head 7} & \textbf{Head 8} & \textbf{Head 17} \\

    \rotatebox[origin=c]{90}{\textbf{Flowers}}
    & \adjincludegraphics[
        trim={{0.75\width} 0 0 {0.67\height}},clip,width=0.28\textwidth
      ]{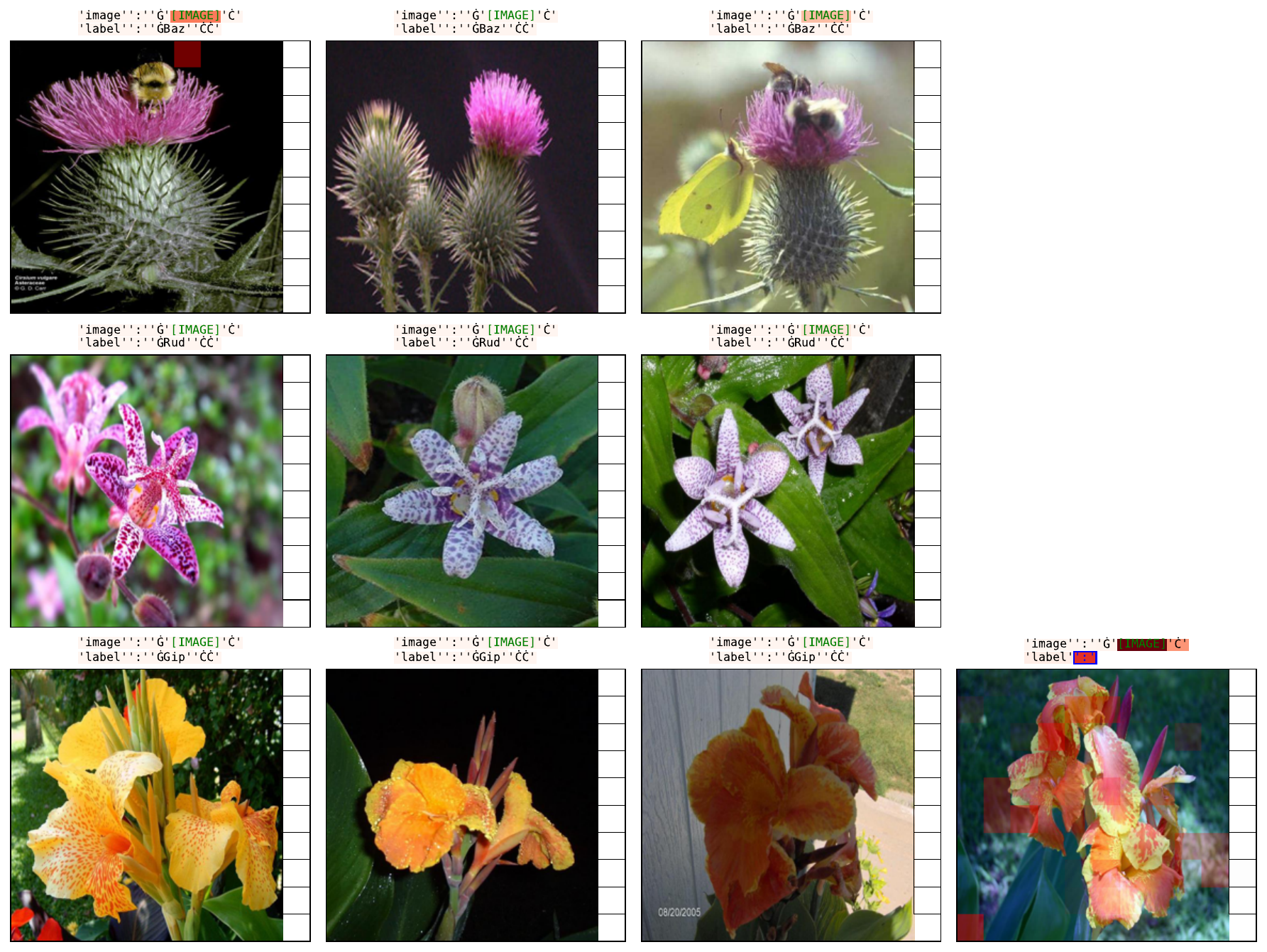}
    & \adjincludegraphics[
        trim={{0.75\width} 0 0 {0.67\height}},clip,width=0.28\textwidth
      ]{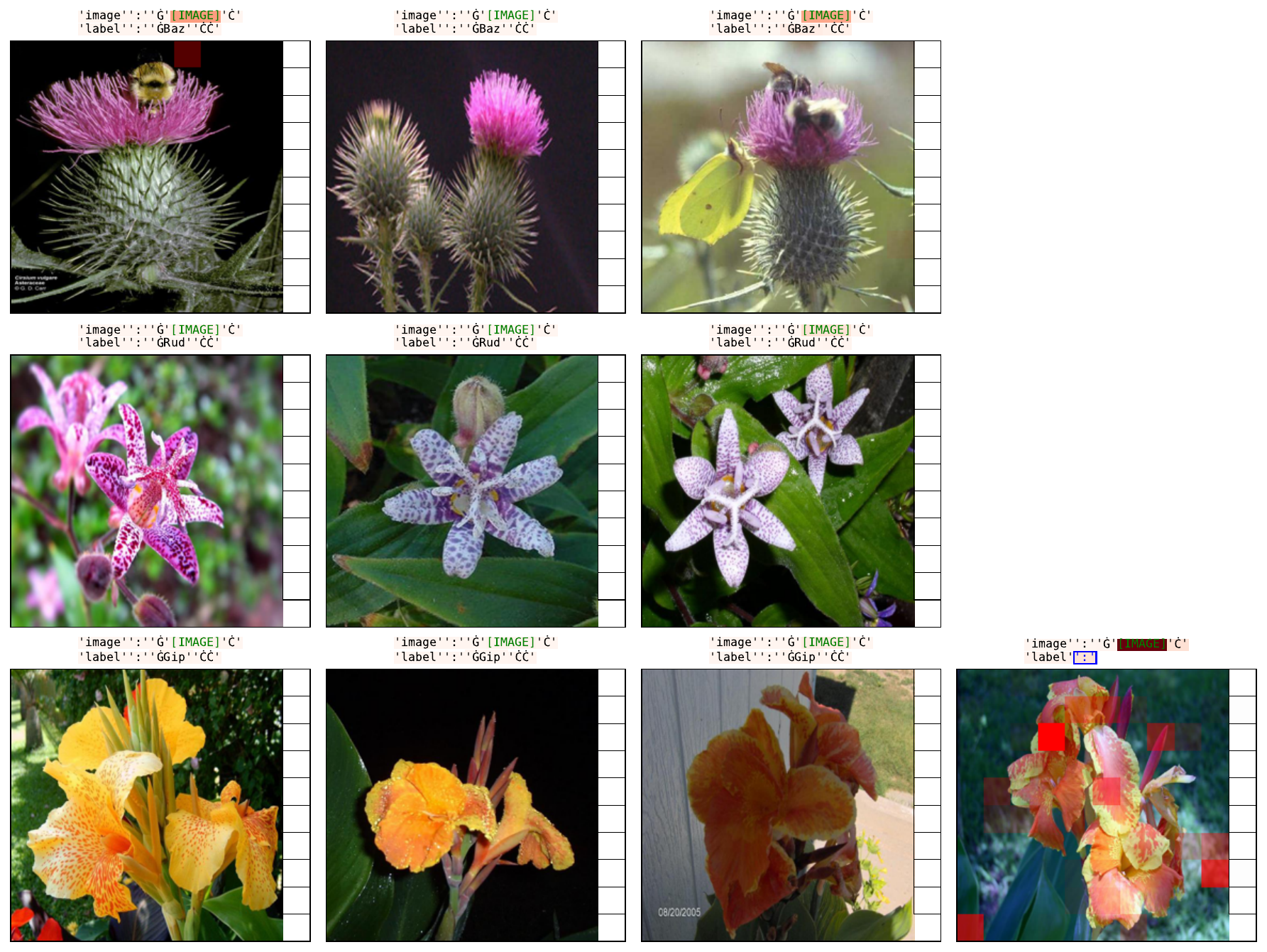}
    & \adjincludegraphics[
        trim={{0.75\width} 0 0 {0.67\height}},clip,width=0.28\textwidth
      ]{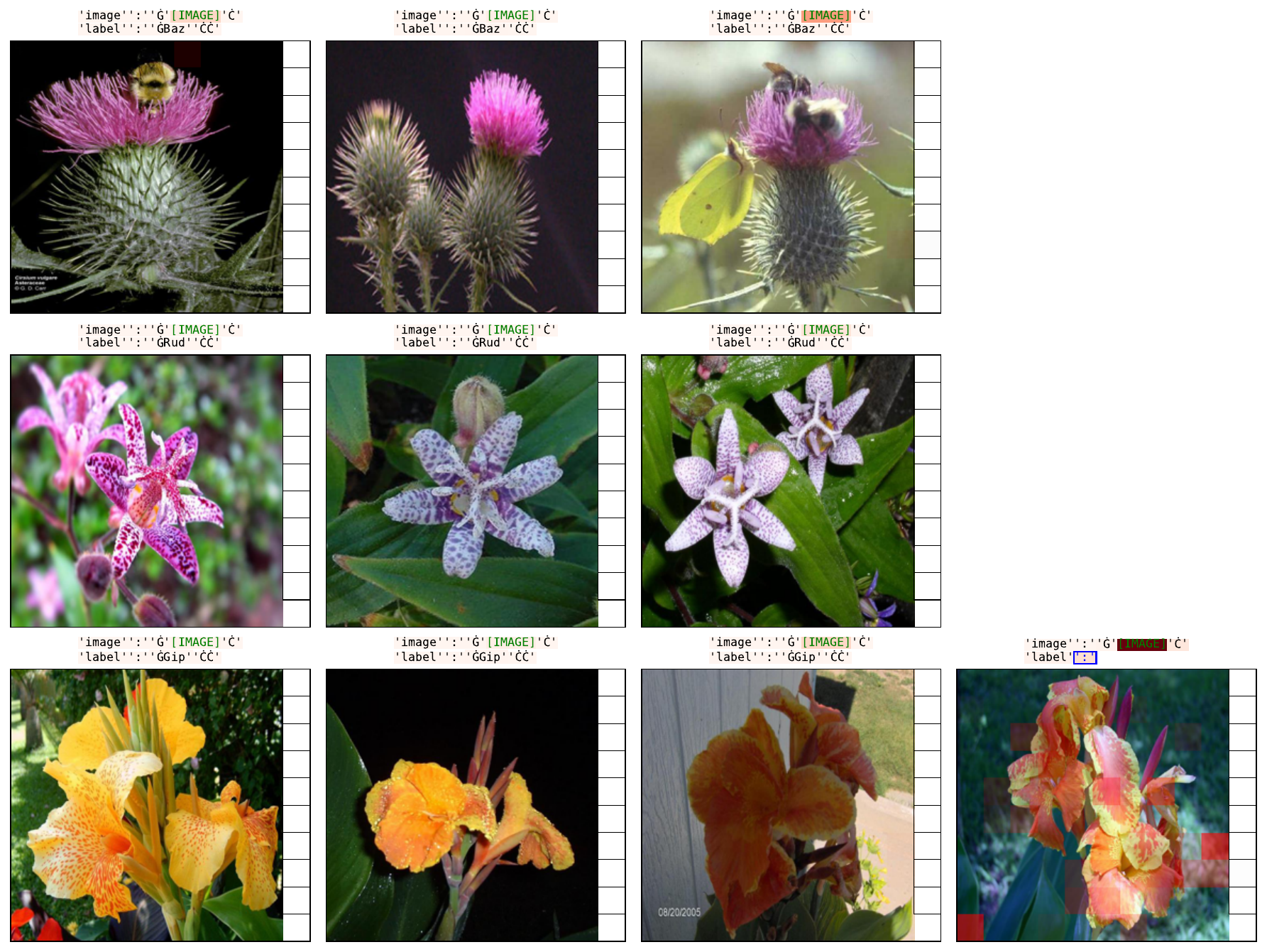} \\[4pt]

    \rotatebox[origin=c]{90}{\textbf{Textures}}
    & \adjincludegraphics[
        trim={{0.75\width} {0.67\height} 0 0},clip,width=0.28\textwidth
      ]{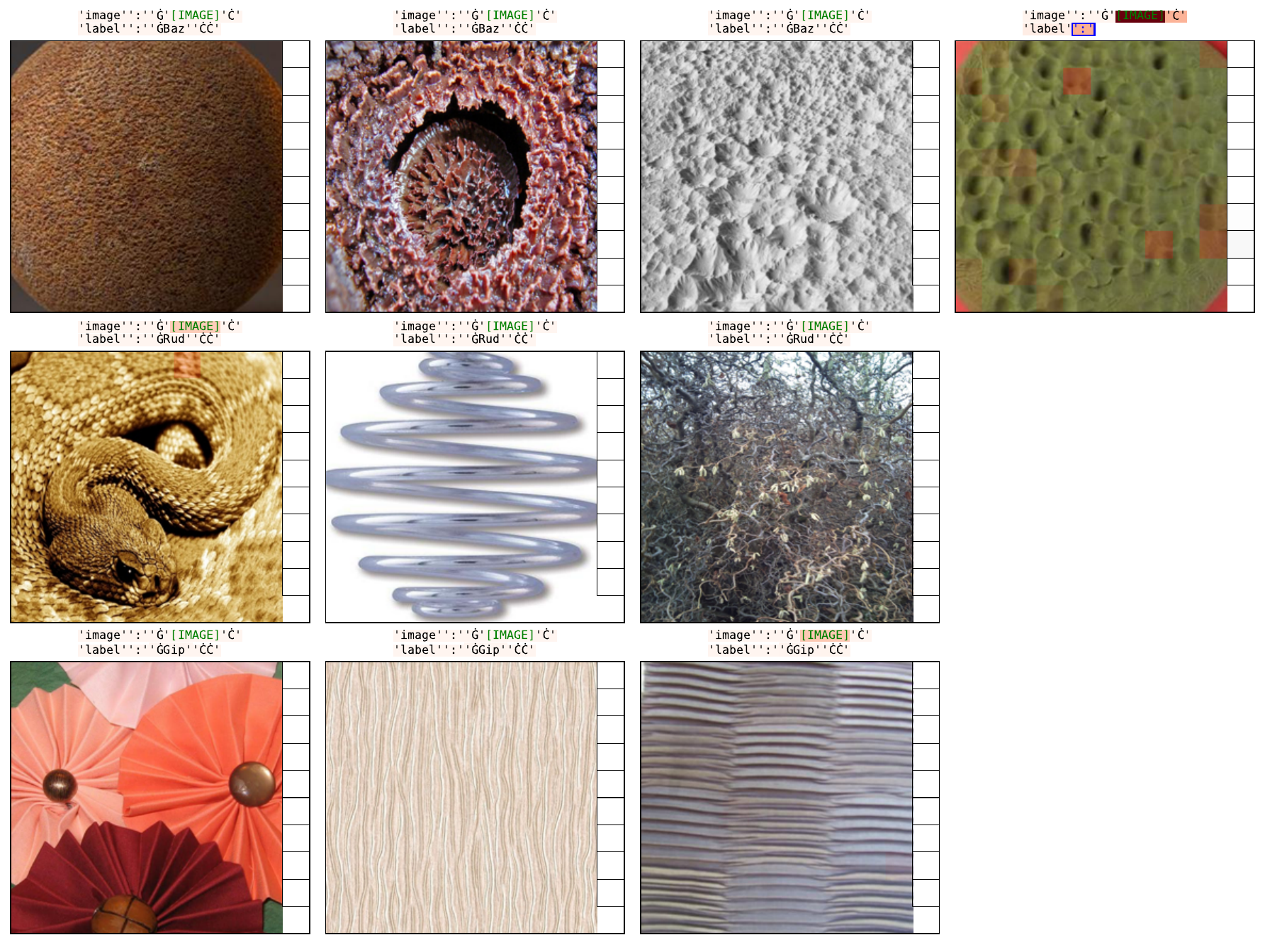}
    & \adjincludegraphics[
        trim={{0.75\width} {0.67\height} 0 0},clip,width=0.28\textwidth
      ]{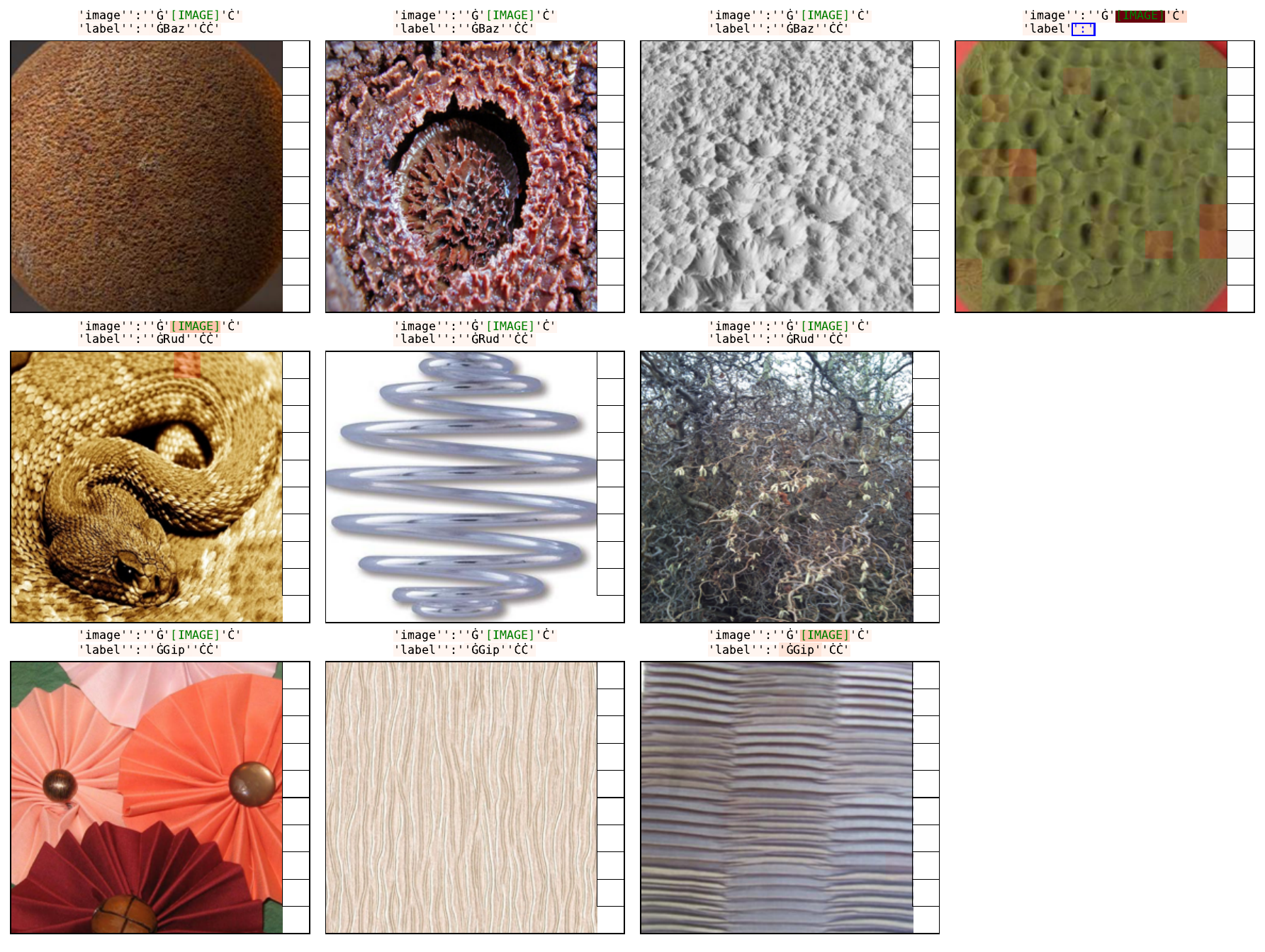}
    & \adjincludegraphics[
        trim={{0.75\width} {0.67\height} 0 0},clip,width=0.28\textwidth
      ]{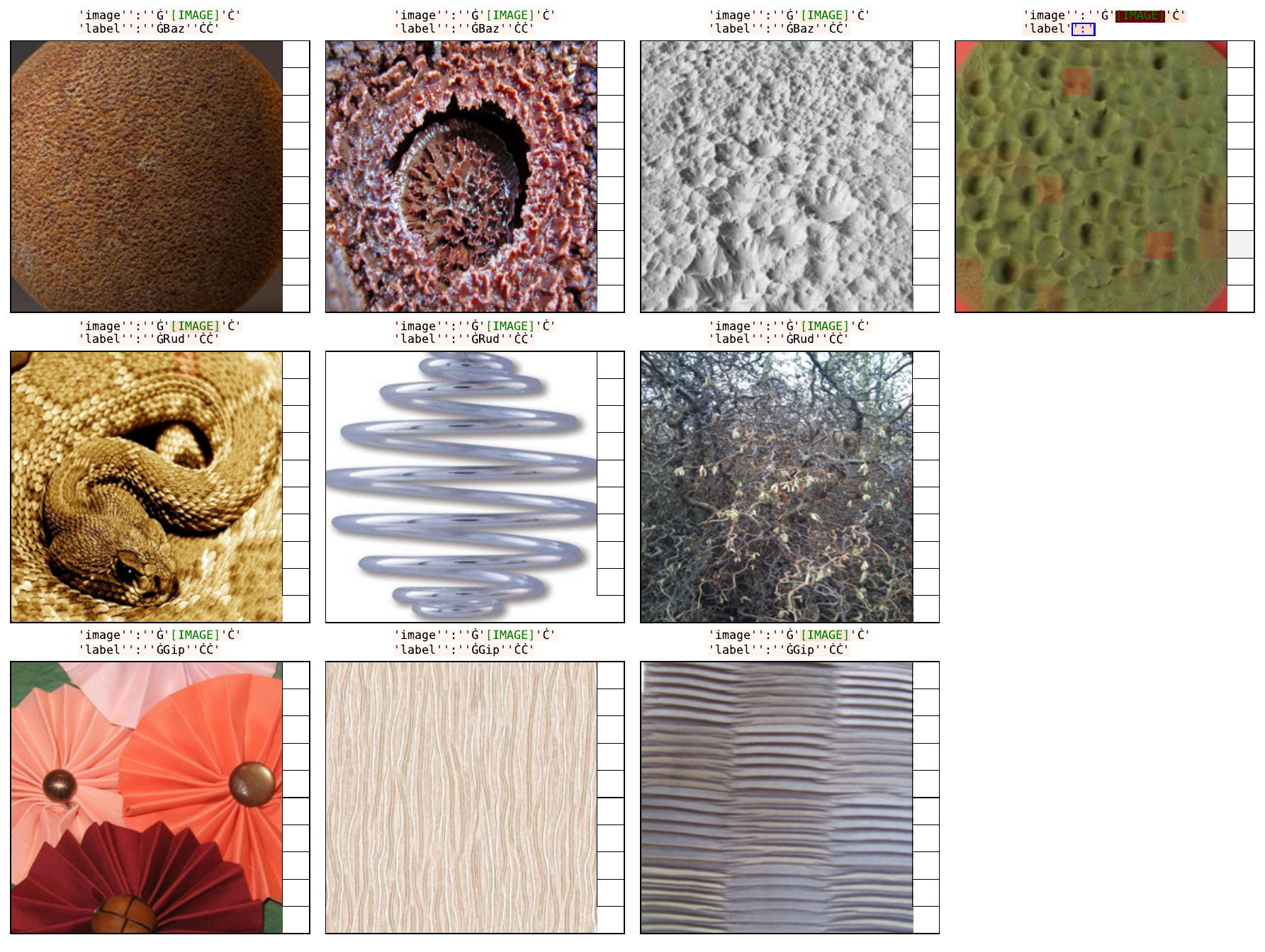} \\[4pt]

    \rotatebox[origin=c]{90}{\textbf{Cars}}
    & \adjincludegraphics[
        trim={{0.75\width} {0.67\height} 0 0},clip,width=0.28\textwidth
      ]{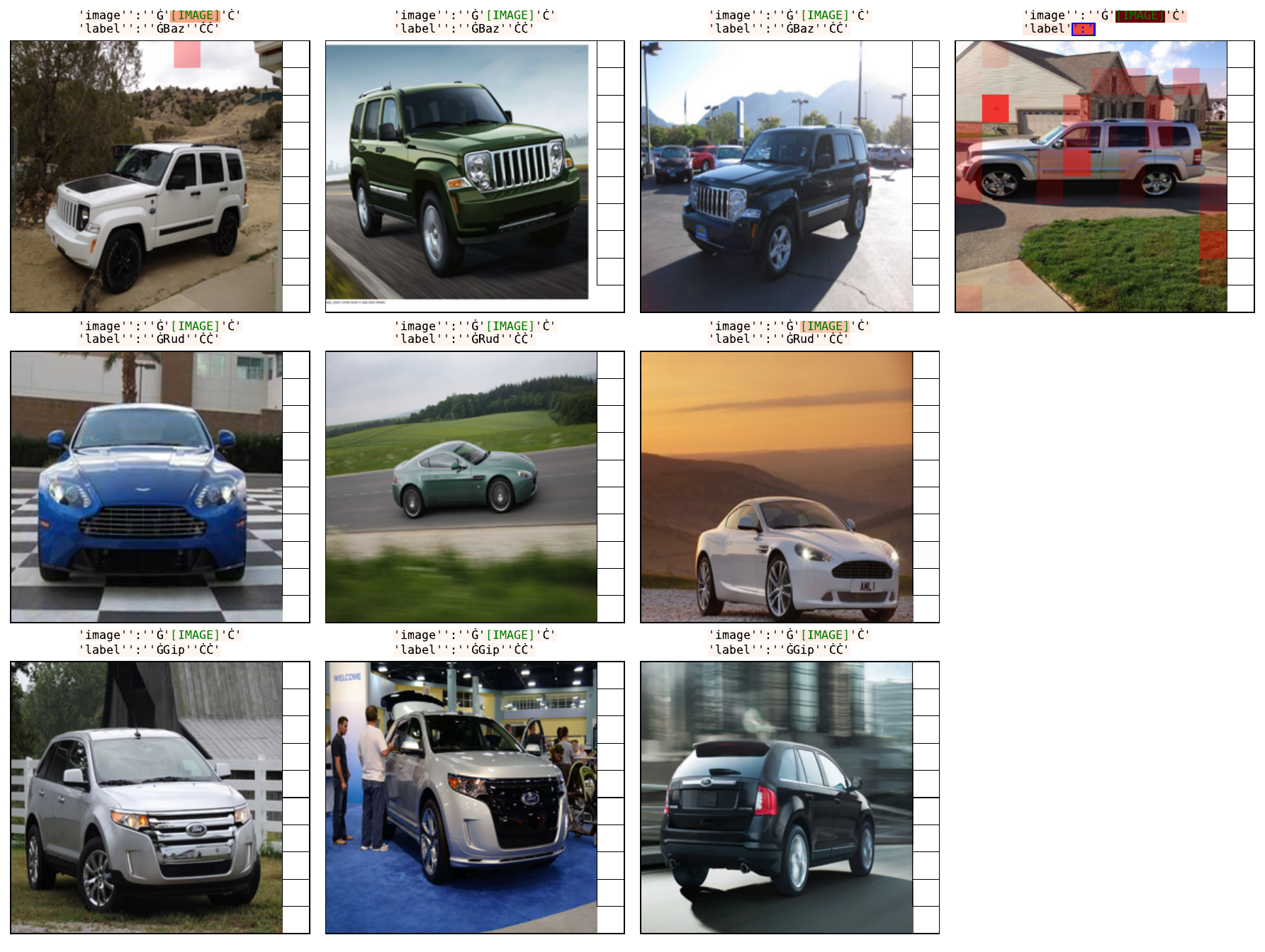}
    & \adjincludegraphics[
        trim={{0.75\width} {0.67\height} 0 0},clip,width=0.28\textwidth
      ]{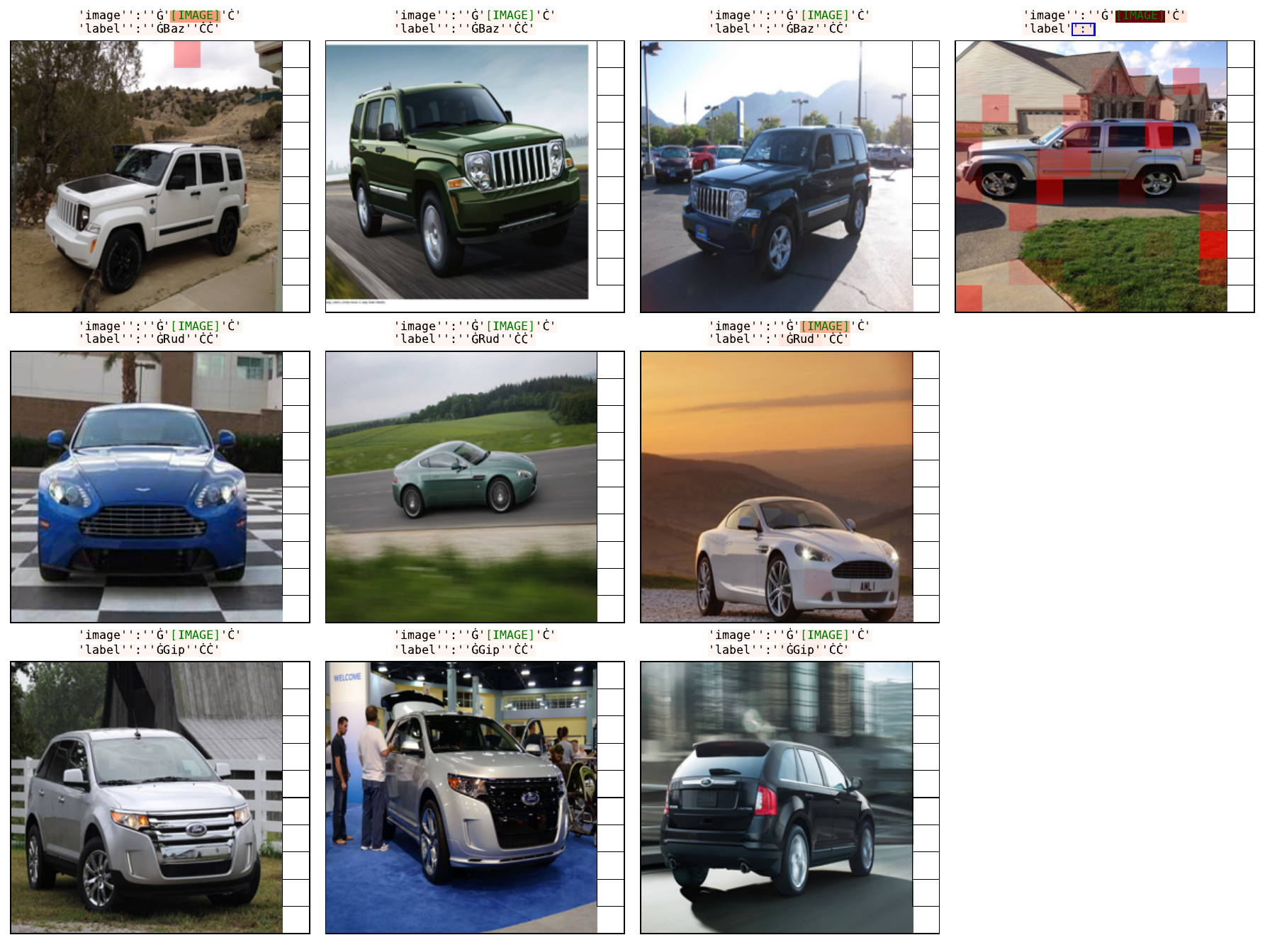}
    & \adjincludegraphics[
        trim={{0.75\width} {0.67\height} 0 0},clip,width=0.28\textwidth
      ]{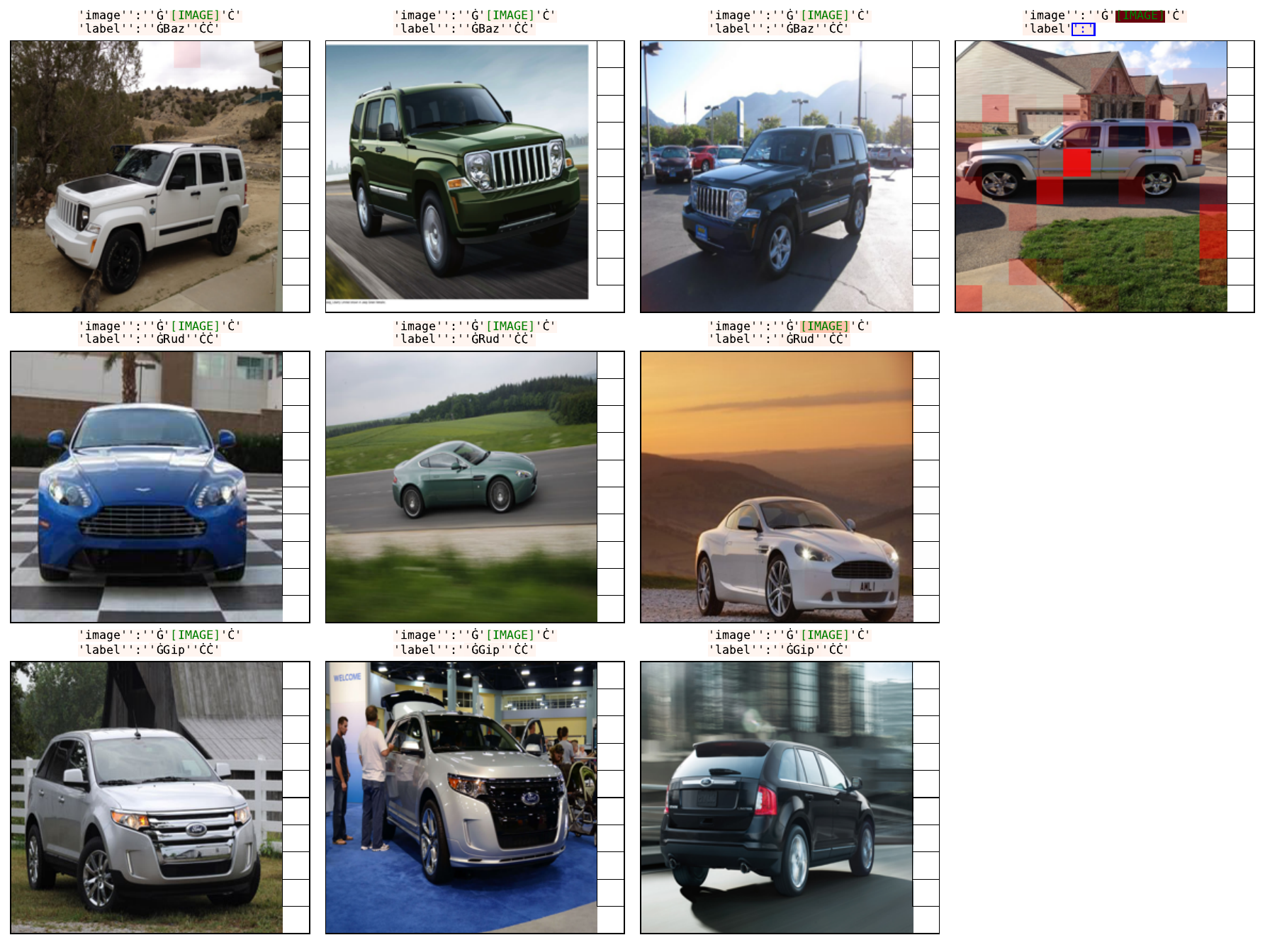}
  \end{tabular}
  \caption{Attention from the Forerunner token to image tokens
    at layer \lstar. Columns correspond to attention heads 7, 8,
    and 17; rows correspond to the Flowers, Textures and Cars
    datasets.}
  \label{fig:supp_foreigner_attention}
\end{figure*}

In this section we study in more detail the circuit that transfers information from Vision SDG in Vision tokens at \lstarprev to the Text SDG in the Forerunner token at \lstar.
We start by doing an experiment where we compute the NCC accuracy in the Vision SDG at \lstarprev for each token position in the image.
To do that, we simply replace average pooling with selecting the token at the desired position.
\Cref{fig:attn_prev_score} (Right) shows the results of that experiment. We observe that some tokens achieve better accuracy than others. In particular, 3 tokens have particularly high accuracy, which we name sink tokens 1, 2 and 3 \cite{kang2025see}.
Consistent with \cite{choi2026sinks}, this suggests that these tokens contain meaningful semantic information across tasks.
\Cref{fig:attn_prev_score} (Left) reports attention from the query forerunner token to the query image for each head. We see a spike in the 2 layers before \lstar where heads attend more to that image. The main paper experiment in \cref{tab:vision_attention_ablation} shows that most of the transfer happens between layers \lstarprev and \lstar. Thus we focus on the 3 heads that attend the most to the query image in this layer (heads 7, 8 and 17 of layer 16).
\Cref{fig:supp_foreigner_attention} shows the attention from the forerunner token to the query image for the 3 selected heads on 3 different tasks. As those heads do not attend to other in-context images we don't show attention to other in-context images.
The same token positions receive high attention from the 3 heads in the 3 displayed tasks. Attention patterns correlate with those in \cref{fig:attn_prev_score} (Right). In particular, sink tokens 2 and 3 always get attention from the 3 heads, although some attention goes to the tokens placed on the object of interest. \textbf{In conclusion, it seems that high-quality features concentrate in certain specific sink tokens at fixed positions on the image, and the Forerunner token attends to these sink tokens in order to transfer information from vision to text.}
The schematization of the ablations performed in \cref{tab:vision_attention_ablation} is shown in \cref{fig:schema_ablation_vision}


\subsection{SDG Alignment across Layers}
\label{sec:sdg_alignment}

\begin{figure*}[t]
    \centering
    \begin{subfigure}[t]{0.32\textwidth}
        \centering
        \includegraphics[width=\linewidth]{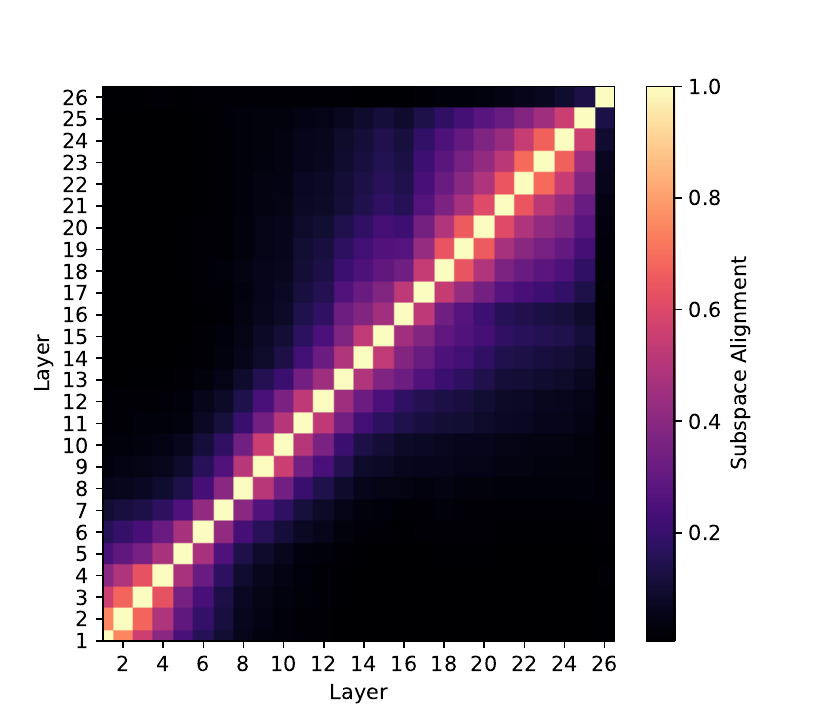}
        \caption{Vision SDG}
        \label{fig:sdg_alignment_vision}
    \end{subfigure}
    \hfill
    \begin{subfigure}[t]{0.32\textwidth}
        \centering
        \includegraphics[width=\linewidth]{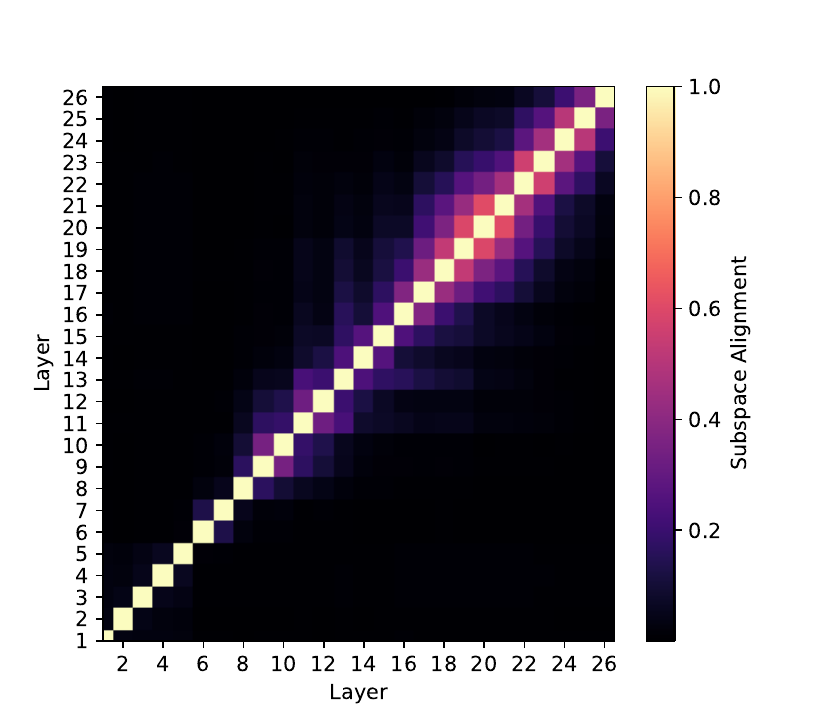}
        \caption{Text SDG}
        \label{fig:sdg_alignment_forerunner}
    \end{subfigure}
    \hfill
    \begin{subfigure}[t]{0.32\textwidth}
        \centering
        \includegraphics[width=\linewidth]{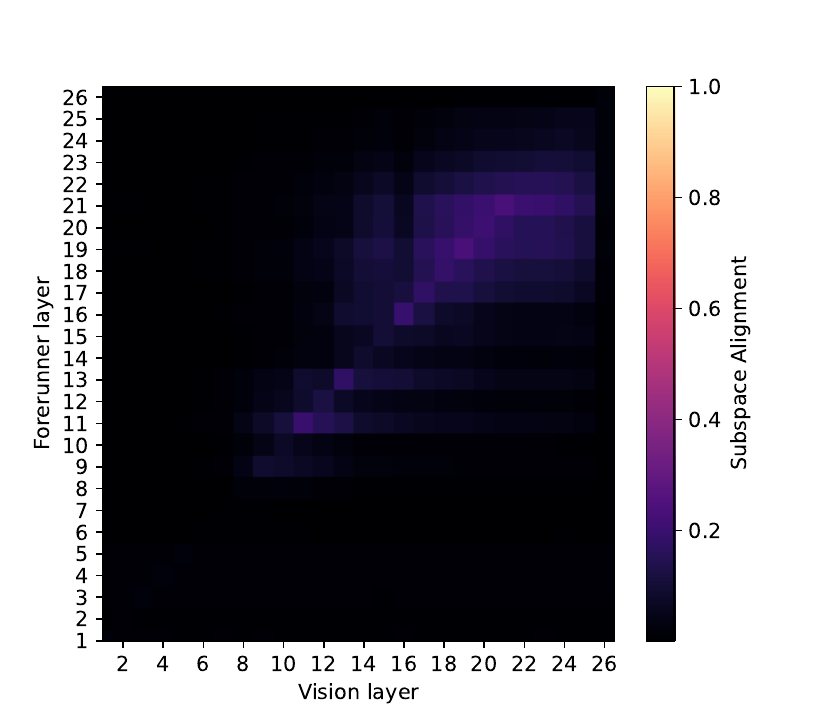}
        \caption{Text vs.\ Vision SDG}
        \label{fig:sdg_alignment_cross}
    \end{subfigure}
    \caption{SDG alignment across layers, measured using $1-\operatorname{BDE}$.\\
\textbf{Left:} $1-\operatorname{BDE}\bigl((\Phi_i^{\mathrm v})^\top\Phi_i^{\mathrm v},\,(\Phi_j^{\mathrm v})^\top\Phi_j^{\mathrm v}\bigr)$.
\textbf{Middle:} $1-\operatorname{BDE}\bigl(\Phi_i^\top\Phi_i,\,\Phi_j^\top\Phi_j\bigr)$.\\
\textbf{Right:} $1-\operatorname{BDE}\bigl(\Phi_i^\top\Phi_i,\,(\Phi_j^{\mathrm v})^\top\Phi_j^{\mathrm v}\bigr)$.
    Rows correspond to layer $i$ and columns to layer $j$.}
    \label{fig:sdg_alignment}
\end{figure*}

We compare the Vision and Text SDGs across layers using the $\operatorname{BDE}$ (see~\cref{eq:BDE}).
\Cref{fig:sdg_alignment_vision} shows that the Vision SDG remains relatively aligned between neighboring layers, but gradually drifts over larger layer distances.
This illustrates the benefit of learning a separate down-projection at each layer to track the changing geometry.
The Text SDG shows weaker alignment between neighboring layers before $\lstar$ and becomes more stable after that layer (\cref{fig:sdg_alignment_forerunner}).
Finally, alignment between the Vision and Text SDGs remains low (\cref{fig:sdg_alignment_cross}), suggesting that their dominant directions are approximately orthogonal. In other words, when the Vision SDG is transferred to text tokens, a rotation is applied.

\subsection{Analysis of Dimensionality Reduction Heads}
\label{sec:vision_compression_heads}

In this section we study in more detail the dimensionality reduction circuit in Vision tokens from layer 1 to \lstarprev.

\paragraph{Choice of $s_{OV}$.}
We derive the identity used in~\cref{eq:BDE}. Let $\Delta=S-S'$.
We normalize each similarity matrix as \(S\mapsto S/\|S\|_F\) before computing \(s_{OV}\), making the score invariant to independent rescaling.
For independent $x,y\sim\mathcal N(0,I_D)$,
\begin{align}
2\operatorname{BDE}(S,S')
&=\mathbb E_{x,y}\!\left[(x^\top\Delta y)^2\right] \\
&=\mathbb E_x\!\left[
x^\top\Delta\,\mathbb E_y[yy^\top]\,\Delta^\top x
\right] \\
&=\operatorname{Tr}\!\left(
\Delta\Delta^\top\mathbb E_x[xx^\top]
\right) \\
&=\operatorname{Tr}(\Delta\Delta^\top)
=\|\Delta\|_F^2.
\end{align}
Note that the Gaussian assumption can be relaxed: sampling \(x\) and \(y\) independently and uniformly from the unit sphere gives the same BDE when the expectation in \cref{eq:BDE} is multiplied by \(D^2\).
The score $s_{OV}$ measures how well $W_{OV}$ maps the previous Vision SDG at layer $\ell$ to the next Vision SDG at layer $\ell+1$.
Note that the score is bounded between 0 and 1 for positive semidefinite matrices, as is always the case here because we define bilinear similarity through \(\Phi^\top\Phi\). The bound does not hold for arbitrary matrices.
For our theoretical circuit, the $Z$ subspace is selected by
$
\Phi_\ell^{\mathrm v}=\Phi_{\ell+1}^{\mathrm v}
=\begin{pmatrix}0&I_{D_\downarrow}\end{pmatrix}.
$
Since $W_{OV}=\eta\Phi_\ell^{\mathrm v\top}\Phi_\ell^{\mathrm v}$, this gives
$
s_{OV}=0.
$
Note that this score is invariant to orthogonal rotations that $W_{OV}$ could perform between subspaces.
However, we find that this simple score allows us to select dimensionality reduction heads reliably.
\clearpage
As an alternative to the BDE, we could measure the largest bilinear distortion over unit-norm inputs:
\begin{equation}
\begin{aligned}
\mathrm{BDE}_{\mathrm{spec}}(S,S')
&:= \frac12\sup_{\|x\|_2=\|y\|_2=1}
\left(x^\top S y-x^\top S'y\right)^2\\
&= \frac12\sup_{\|x\|_2=\|y\|_2=1}
\left(x^\top(S-S')y\right)^2\\
&= \frac12\sup_{\|y\|_2=1}\|(S-S')y\|_2^2\\
&= \frac12\|S-S'\|_2^2.
\end{aligned}
\end{equation}
This choice requires no assumption on the distribution of \(x\) and \(y\). The resulting spectral norm is equivalent to the Frobenius norm, with \(\|\Delta\|_2\leq\|\Delta\|_F\leq\sqrt{\operatorname{rank}(\Delta)}\,\|\Delta\|_2\).

\begin{figure*}[t]
    \centering
    \includegraphics[width=0.9\linewidth]{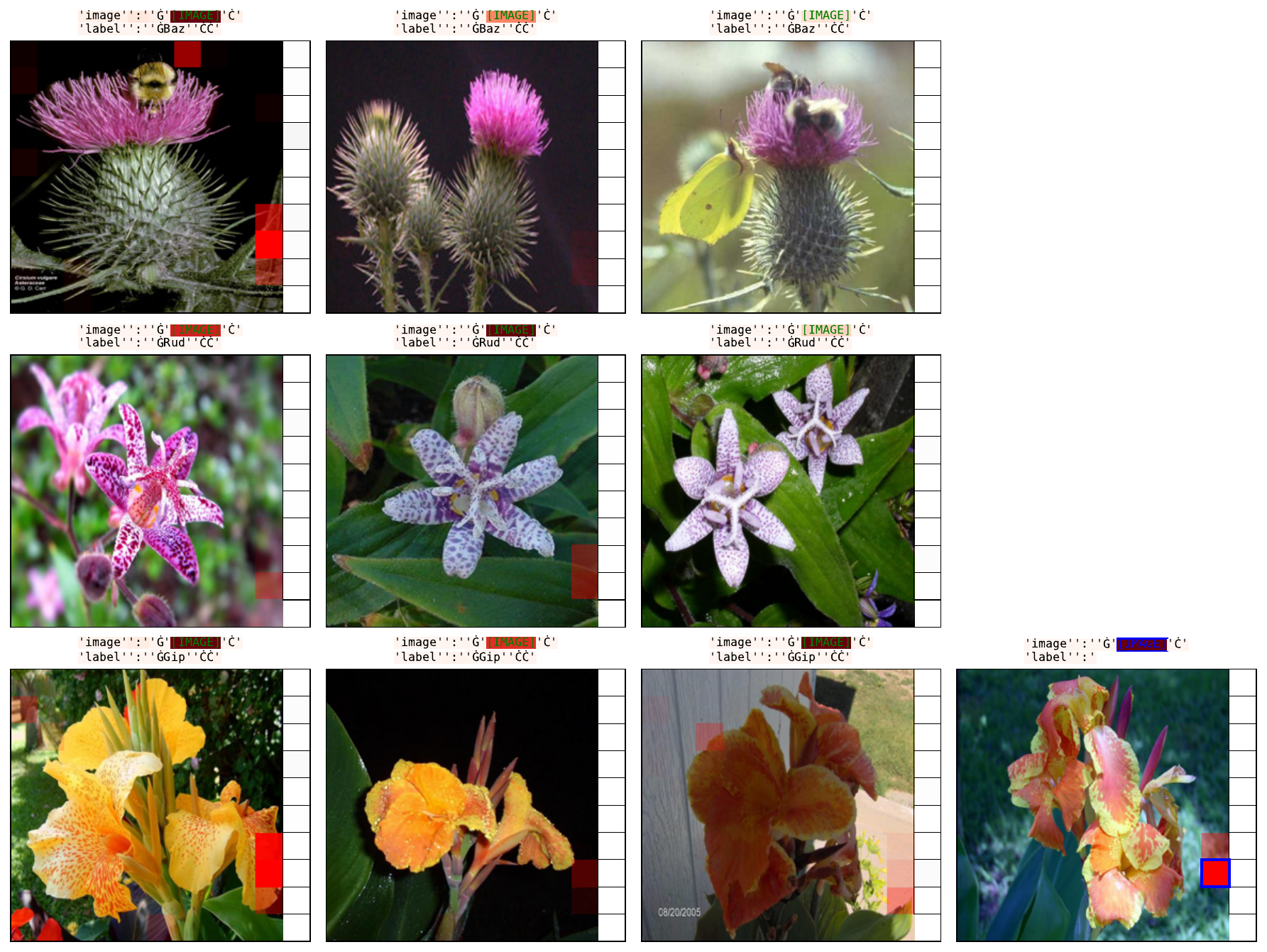}
    \caption{Visualization of attention from sink token 2 to other in-context images for head 27 at layer 12 (dimensionality reduction head), on a task from the Flowers dataset.}
    \label{fig:sink_example_1_1}
\end{figure*}

\begin{figure*}[t]
    \centering
    \includegraphics[width=0.9\linewidth]{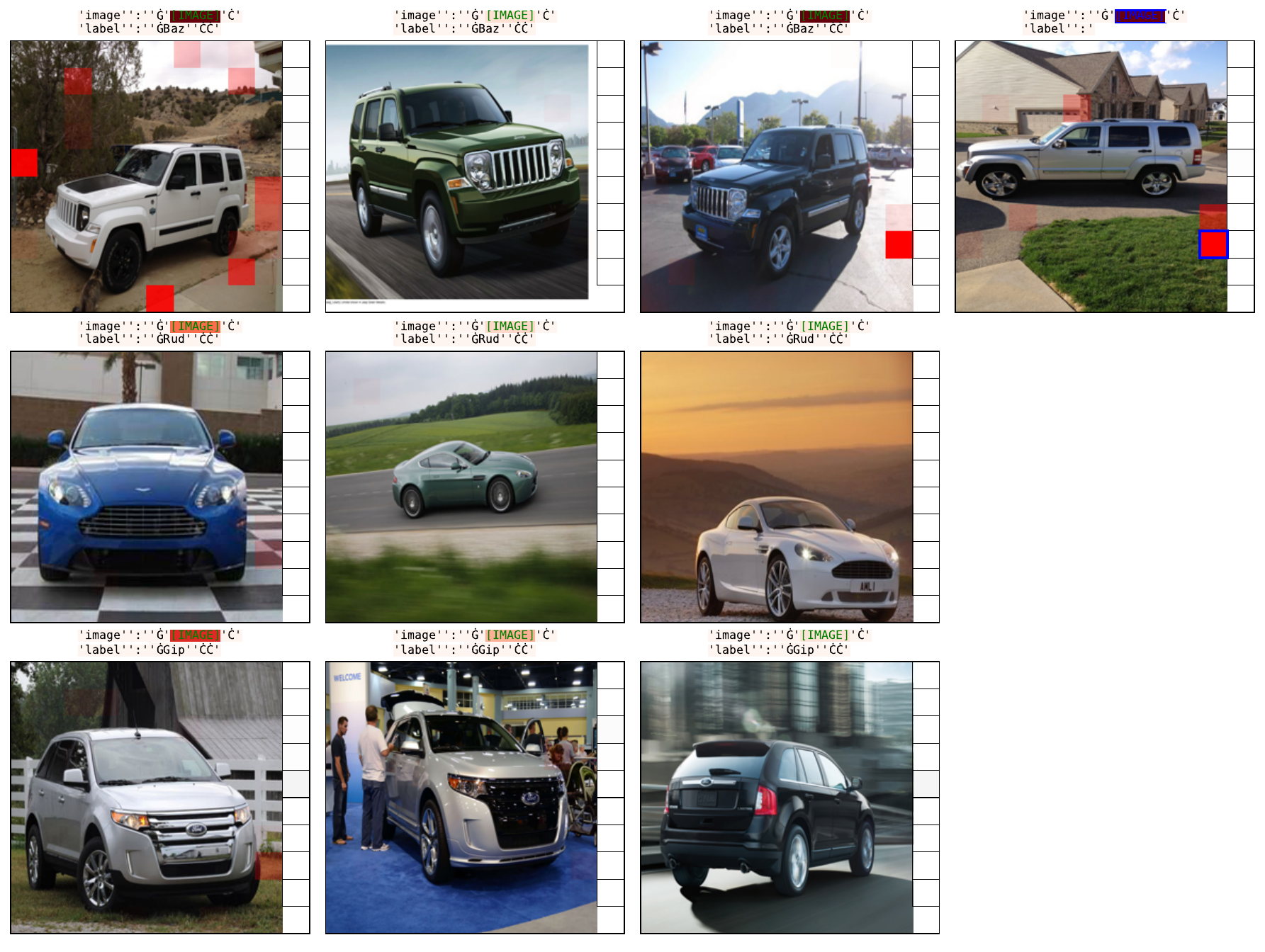}
    \caption{Visualization of attention from sink token 2 to other in-context images for head 27 at layer 12 (dimensionality reduction head), on a task from the Cars dataset.}
    \label{fig:sink_example_1_2}
\end{figure*}

\begin{figure*}[t]
    \centering
    \includegraphics[width=0.9\linewidth]{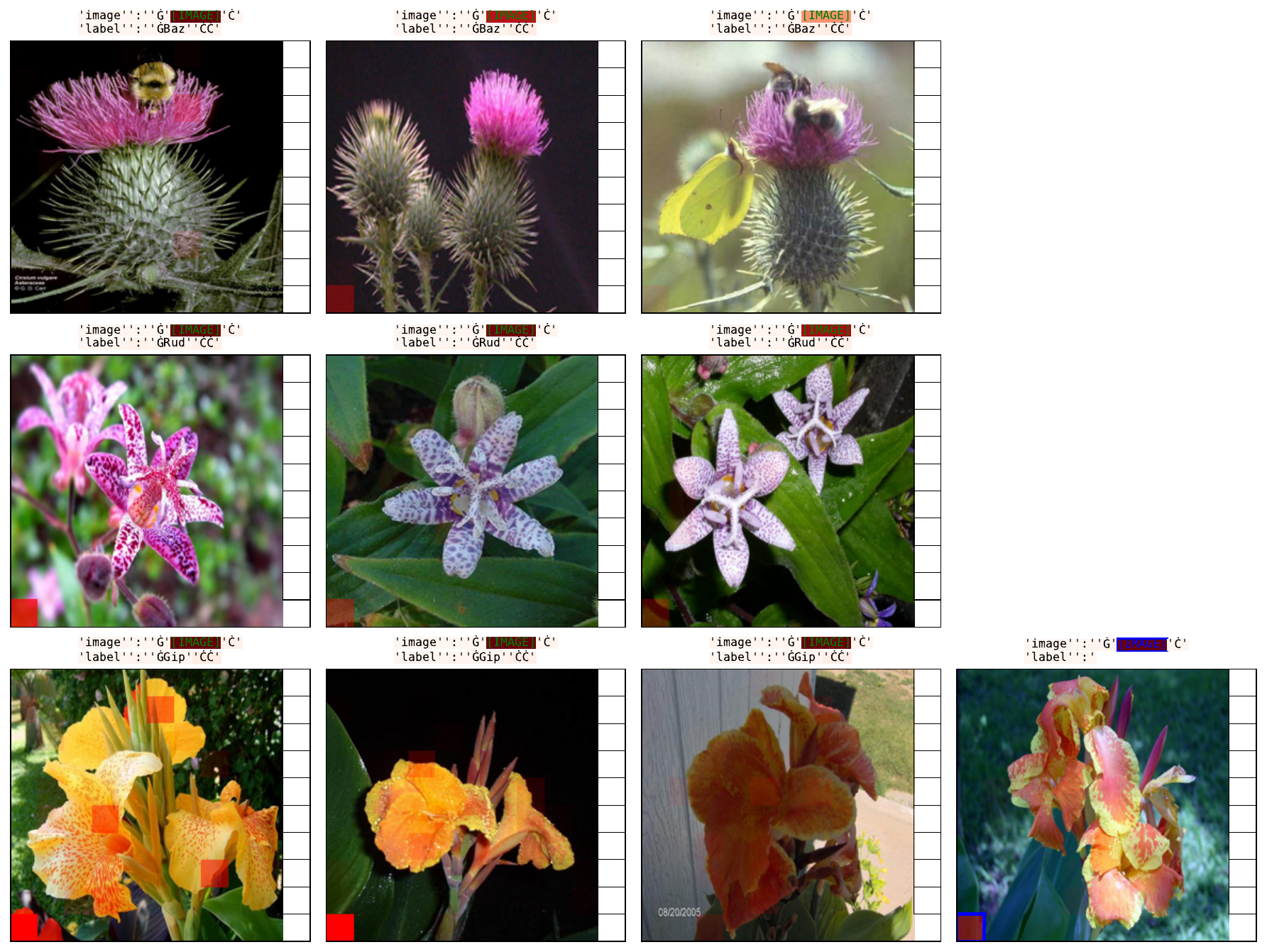}
    \caption{Visualization of attention from sink token 3 to other in-context images for head 26 at layer 10 (dimensionality reduction head), on a task from the Flowers dataset.}
    \label{fig:sink_example_2_1}
\end{figure*}

\begin{figure*}[t]
    \centering
    \includegraphics[width=0.9\linewidth]{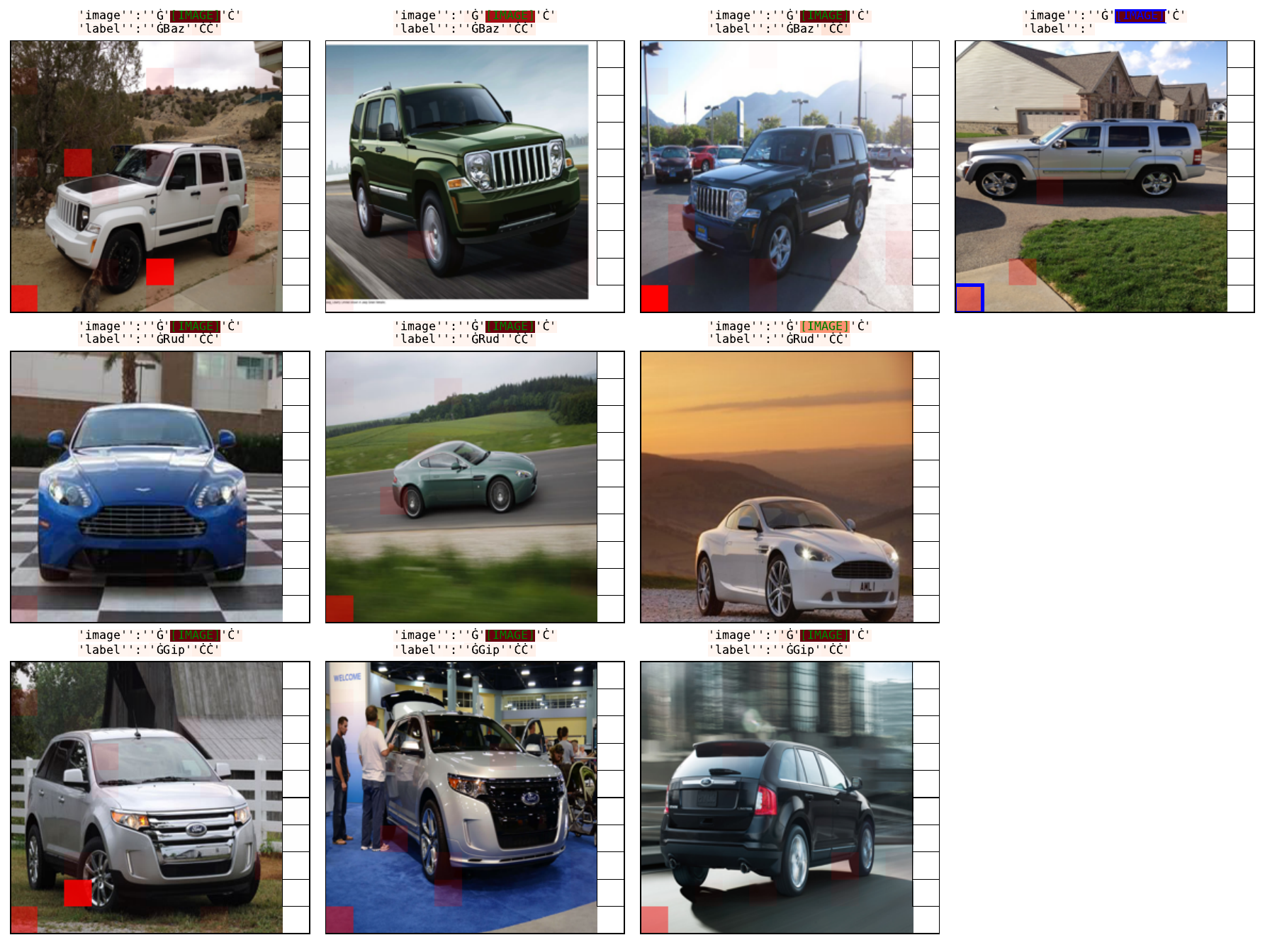}
    \caption{Visualization of attention from sink token 3 to other in-context images for head 26 at layer 10 (dimensionality reduction head), on a task from the Cars dataset.}
    \label{fig:sink_example_2_2}
\end{figure*}

%
%
\paragraph{Attention patterns.} Next we investigate the attention patterns of the selected dimensionality heads.
As there are many vision tokens we focus on the sink tokens identified earlier (see~\cref{fig:attn_prev_score} (Right)).
\Cref{fig:sink_example_1_1,fig:sink_example_1_2,fig:sink_example_2_1,fig:sink_example_2_2} show attention from sink tokens 2 and 3 to other in-context images on tasks from the Flowers and Cars datasets. The blue square marks the querying sink token, and the red shading shows where it attends.
We select the dimensionality reduction head with the lowest $s_{OV}$ score, head 27 in layer 12, and the head with the lowest $s_{OV}$ score in layer 10, head 26.
We see that those heads look more at the previous sink tokens of similar in-context examples, which is coherent with the theoretically derived circuit.

\clearpage
\section{Other Models}
\label{sec:more_models}

\begin{table*}[t]
\centering
\small
\setlength{\tabcolsep}{4pt}
\begin{tabular}{llrrrrrr}
\toprule
Model & Attention & Layers & Heads & Total heads & Token dim. & Head dim. & MLP dim. \\
\midrule
\rowcolor{blue!10}
Ministral 3 (3B) & Full & 26 & 32 & 832 & 3072 & 128 & 9216 \\
Ministral 3 (8B) & Full         & 34 & 32    & 1088    & 4096 & 128     & 14336 \\
Qwen 3.5 (4B)   & Full/linear  & 32 & 16/32 & 128/768 & 2560 & 256/128 & 9216  \\
Qwen 3.5 (9B)   & Full/linear  & 32 & 16/32 & 128/768 & 4096 & 256/128 & 12288 \\
Qwen 3 (4B)     & Full         & 36 & 32    & 1152    & 2560 & 128     & 9728  \\
\bottomrule
\end{tabular}
\caption{
LLM-decoder hyperparameters.
The \colorbox{blue!10}{highlighted model} is used for the main paper experiments and analyses.
Slash-separated values follow the order of attention types.
Each Qwen model has 8 full-attention layers and 24 linear-attention layers;
head counts refer to query heads for full attention and value heads for linear attention.
}
\label{tab:model_hparams}
\end{table*}

We experiment in the main paper with Ministral 3 (3B) ~\citep{liu2026ministral}. Their main architectural hyperparameters are reported in \cref{tab:model_hparams}.
Both decoders use grouped-query attention and a context length of 262,144 tokens. 
Both models use the same Pixtral vision encoder. 
\Cref{tab:model_hparams} also contains hyperparameters of the other models we experiment with.

\begin{figure*}[htbp]
    \centering
    \begin{subfigure}[t]{0.48\textwidth}
        \centering
        \includegraphics[width=\linewidth]{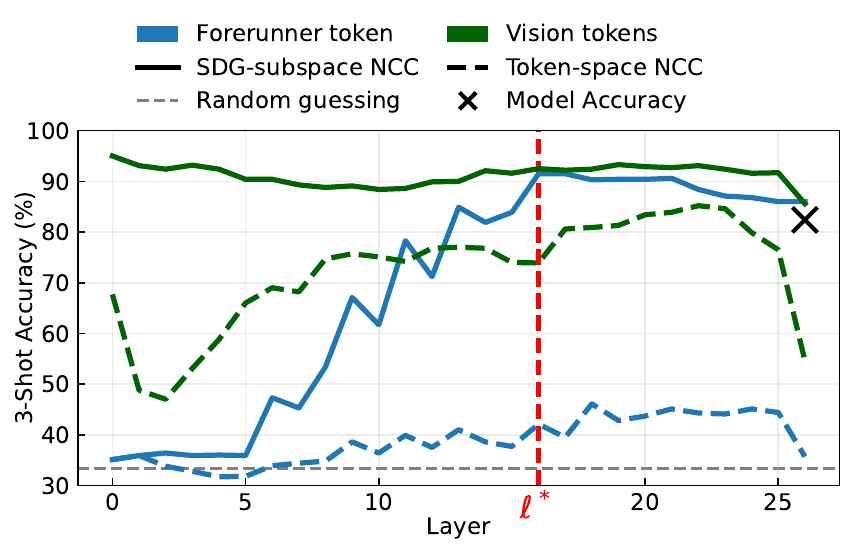}
        \caption{Ministral 3 (3B)}
    \end{subfigure}
    \hfill
    \begin{subfigure}[t]{0.48\textwidth}
        \centering
        \includegraphics[width=\linewidth]{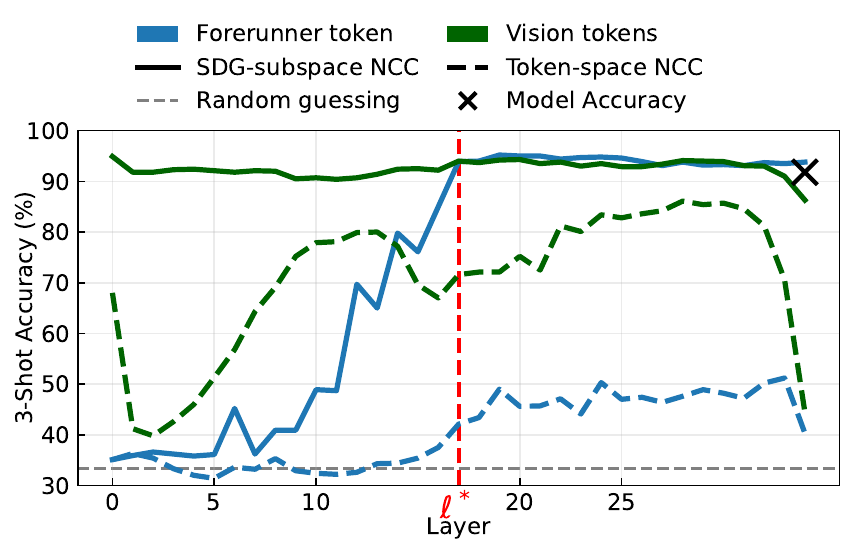}
        \caption{Ministral 3 (8B)}
    \end{subfigure}

    \medskip
    \begin{subfigure}[t]{0.48\textwidth}
        \centering
        \includegraphics[width=\linewidth]{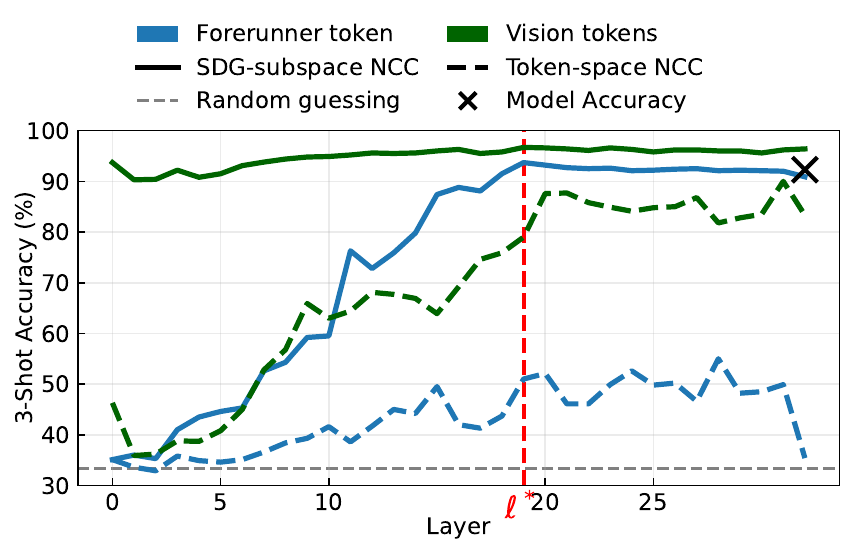}
        \caption{Qwen 3.5 (9B)}
    \end{subfigure}
    \hfill
    \begin{subfigure}[t]{0.48\textwidth}
        \centering
        \includegraphics[width=\linewidth]{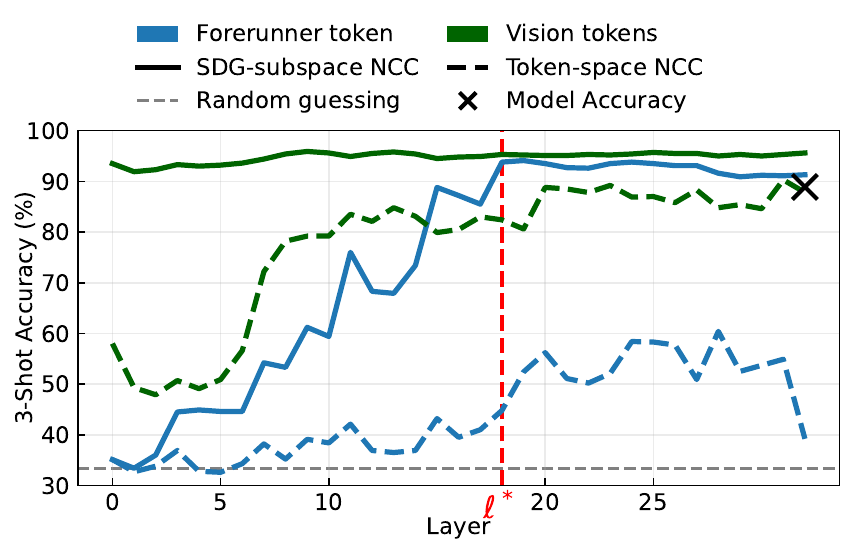}
        \caption{Qwen 3.5 (4B)}
    \end{subfigure}

    \medskip    
    \begin{subfigure}[t]{0.48\textwidth}
        \centering
        \includegraphics[width=\linewidth]{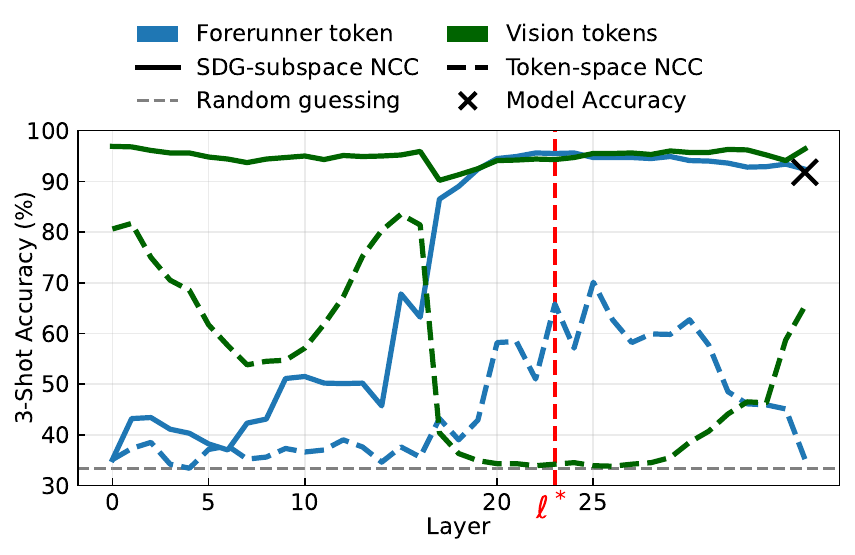}
        \caption{Qwen 3 (4B)}
    \end{subfigure}
    \hfill
    
    \caption{NCC accuracy before and after projection onto the Shared Discriminative Geometry.}
    \label{fig:more_models_ncc_trained}
\end{figure*}

\begin{figure*}[htbp]
    \centering
    \begin{subfigure}[t]{0.48\textwidth}
        \centering
        \includegraphics[width=\linewidth]{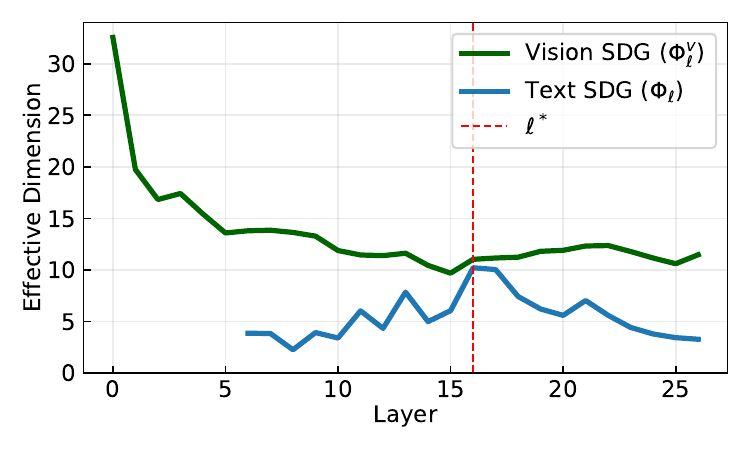}
        \caption{Ministral 3 (3B)}
    \end{subfigure}
    \hfill
    \begin{subfigure}[t]{0.48\textwidth}
        \centering
        \includegraphics[width=\linewidth]{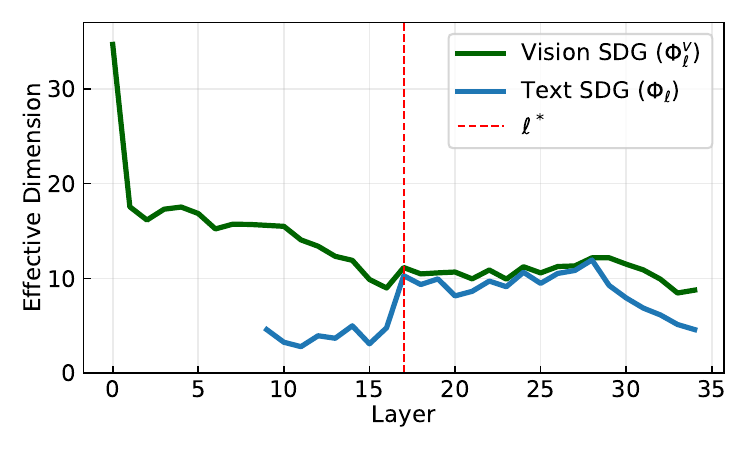}
        \caption{Ministral 3 (8B)}
    \end{subfigure}

    \medskip
    \begin{subfigure}[t]{0.48\textwidth}
        \centering
        \includegraphics[width=\linewidth]{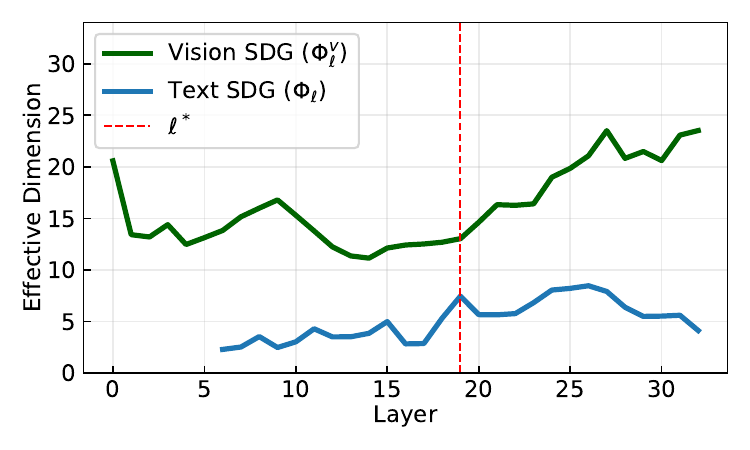}
        \caption{Qwen 3.5 (9B)}
    \end{subfigure}
    \hfill
    \begin{subfigure}[t]{0.48\textwidth}
        \centering
        \includegraphics[width=\linewidth]{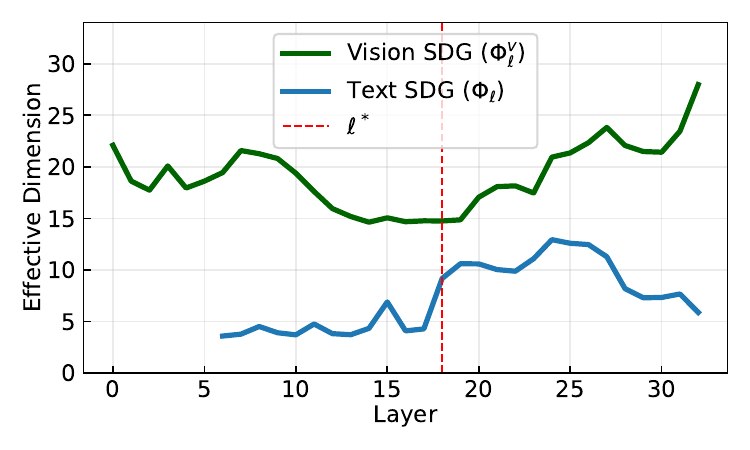}
        \caption{Qwen 3.5 (4B)}
    \end{subfigure}

    \medskip
    \begin{subfigure}[t]{0.48\textwidth}
        \centering
        \includegraphics[width=\linewidth]{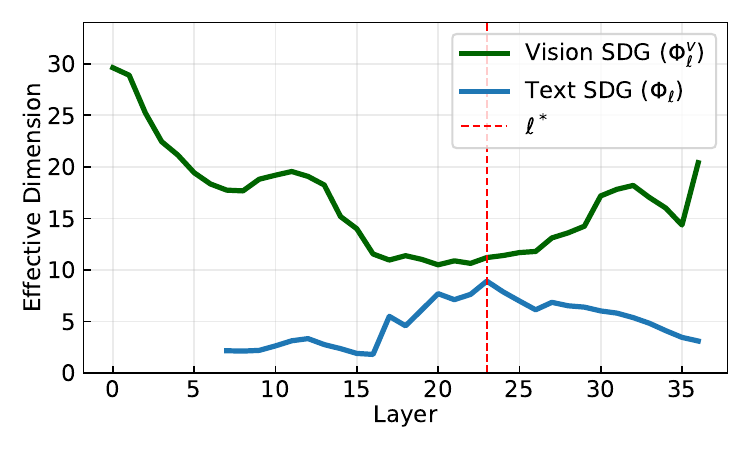}
        \caption{Qwen 3 (4B)}
    \end{subfigure}

    \caption{Effective dimension across layers of $\Phi_\ell$ (Text SDG) and $\Phi^{\mathrm v}_\ell$ (Vision SDG).}
    \label{fig:more_models_effective_dim}
\end{figure*}

\begin{figure*}[htbp]
    \centering
    \begin{subfigure}[t]{0.44\textwidth}
        \centering
        \includegraphics[width=\linewidth]{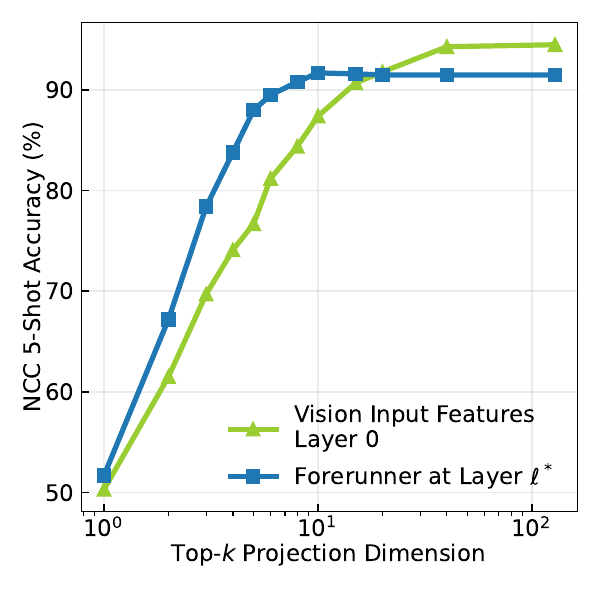}
        \caption{Ministral 3 (3B)}
    \end{subfigure}
    \hfill
    \begin{subfigure}[t]{0.44\textwidth}
        \centering
        \includegraphics[width=\linewidth]{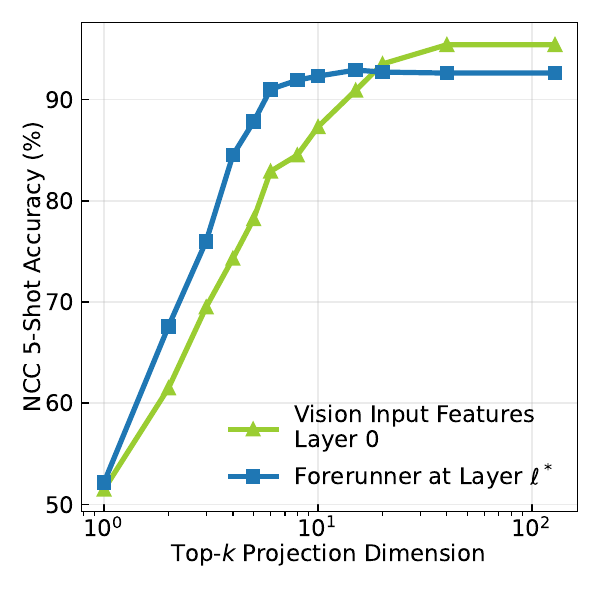}
        \caption{Ministral 3 (8B)}
    \end{subfigure}

    \medskip
    \begin{subfigure}[t]{0.44\textwidth}
        \centering
        \includegraphics[width=\linewidth]{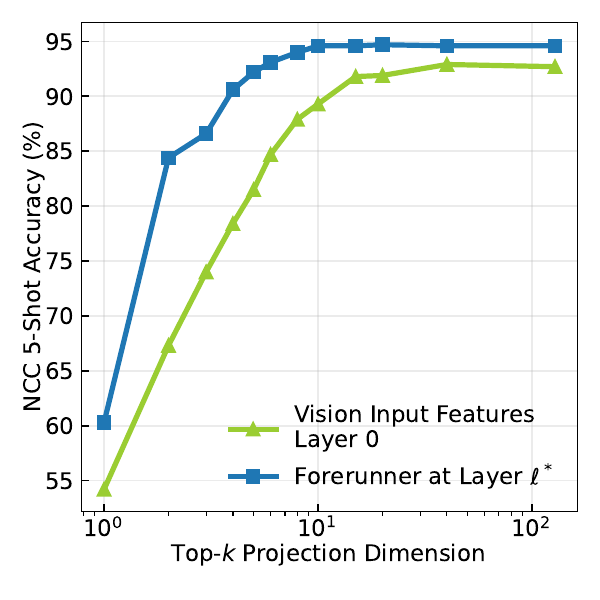}
        \caption{Qwen 3.5 (9B)}
    \end{subfigure}
    \hfill
    \begin{subfigure}[t]{0.44\textwidth}
        \centering
        \includegraphics[width=\linewidth]{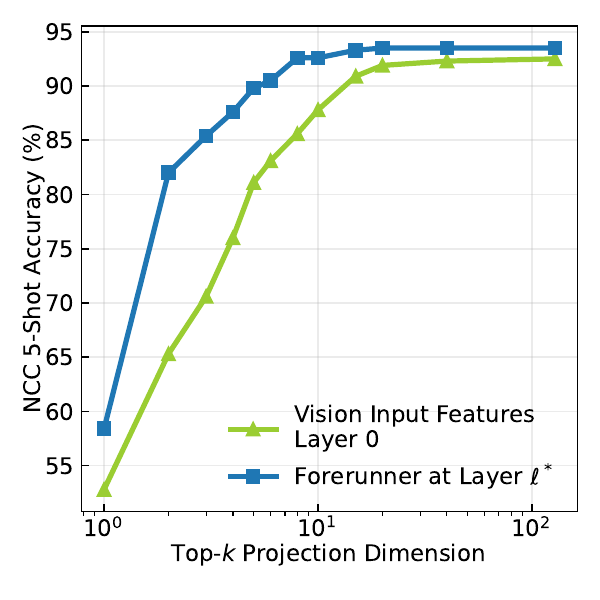}
        \caption{Qwen 3.5 (4B)}
    \end{subfigure}

    \medskip
    \begin{subfigure}[t]{0.44\textwidth}
        \centering
        \includegraphics[width=\linewidth]{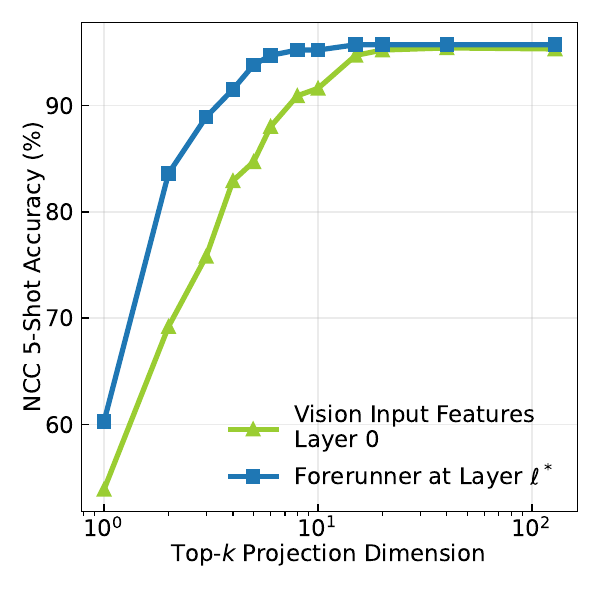}
        \caption{Qwen 3 (4B)}
    \end{subfigure}

    \caption{NCC accuracy in the top-$k$ SDG dimensions of the input vision tokens (Layer $\ell=0$) and the forerunner token at $\lstar$.}
    \label{fig:more_models_ncc_dim}
\end{figure*}

We report results for Ministral 3 (3B and 8B), Qwen 3 (4B) and Qwen 3.5 (4B and 9B): NCC accuracy before and after projection onto the SDG (\cref{fig:more_models_ncc_trained}), effective dimension of the Text and Vision SDGs across layers (\cref{fig:more_models_effective_dim}), and NCC accuracy as a function of projection dimension (\cref{fig:more_models_ncc_dim}).
The peak-accuracy layer $\lstar$ and the effective dimension of the Text SDG at that layer are summarized in \cref{tab:more_models_effective_dim}.

\begin{table}[t]
    \centering
    \begin{tabular}{lcc}
        \toprule
        Model & $\lstar$ & Effective dimension of $\Phi_{\lstar}$ \\
        \midrule
        Ministral 3 (3B) & 16 & 10.22 \\
        Ministral 3 (8B) & 17 & 10.28 \\
        Qwen 3.5 (9B)   & 19 & 7.46 \\
        Qwen 3.5 (4B)   & 18 & 9.18 \\
        Qwen 3 (4B)     & 23 & 8.92 \\
        \bottomrule
    \end{tabular}
    \caption{Peak NCC layer and Text SDG effective dimension across models.}
    \label{tab:more_models_effective_dim}
\end{table}

%
%

\subsection{NCC Accuracy across Layers}

\clearpage
\subsection{NCC Accuracy in the Shared Discriminative Geometry}

\clearpage
\subsection{Effective Dimension}

\clearpage
\subsection{NCC Accuracy by Projection Dimension}

\section{Other Tasks Details}
\label{sec:multimodal_task}

\paragraph{SST-2.}
SST-2 \citep{SST2} is a binary sentiment classification dataset of movie-review sentences labeled positive or negative. We follow the same evaluation protocol, replacing each image with the sentence's sequence of text tokens. In \cref{eq:prompt}, we replace $\NLstr{image:}$ with $\NLstr{text:}$ and replace $f_v(img)$ with the corresponding sentence sequence of text-token embeddings.

\paragraph{Open MI.}
Fast Open-Ended MiniImageNet (Open MI in the main paper)
, from VL-ICL Bench~\citep{zong2025VLICL}, asks the model to
classify ImageNet images using concept names introduced by the
demonstrations. We evaluate all 200 provided tasks, using the support
images supplied for each task.

\paragraph{Matching MI.}
Fast Matching MiniImageNet (Matching MI in the main paper)
, also from VL-ICL Bench, asks whether two images satisfy
a relationship illustrated by the demonstrations. We evaluate all
200 provided tasks and sample demonstrations from the supplied
support set, including both positive and negative pairs.

\paragraph{VLGuard.}
VLGuard~\citep{zong2024VLGuard} contains images paired with user
instructions. We ask the model to classify each pair as
\texttt{harmful} or \texttt{unharmful} and evaluate the first
200 pairs in our prepared dataset.

\paragraph{VizWiz.}
VizWiz~\citep{gurari2018VizWiz} contains photographs and questions
submitted by blind users. We use its answerability labels:
the model predicts whether the question can be answered. We evaluate on 200 generated tasks.

\paragraph{SugarCrepe.}
SugarCrepe~\citep{hsieh2023SugarCrepe} pairs COCO images with correct
captions and modified captions that do not match the image.
We turn each image--caption pair into a yes/no question. We evaluate on 200 generated tasks.

\paragraph{MHaluBench.}
MHaluBench~\citep{chen2024MHaluBench} provides images and textual claims
annotated for hallucination. We ask the model to classify each
claim as \texttt{hallucination} or \texttt{non-hallucination}. We evaluate on 200 generated tasks.

\paragraph{NaturalBench.}
We use the image-caption matching version of NaturalBench~\citep{li2024NaturalBench}. Each caption is presented as a yes/no question
asking whether it describes the image. We evaluate on 200 generated tasks.

Questions and answers retain the dataset-specific formatting in
Table~\ref{tab:question_formats}. Each image placeholder occupies a
separate line, followed by the question and a new line containing
\texttt{Answer: <answer>}. Demonstrations are separated by a newline;
the query ends at \texttt{Answer:}.
Evaluation matches the greedy next token.

\begin{table}[t]
\centering
\small
\begin{tabular}{p{0.16\linewidth} p{0.45\linewidth} p{0.27\linewidth}}
\hline
Dataset & Question format & Answers \\
\hline
Open MI & \texttt{This is a} & Provided concept names \\
VLGuard & Original \texttt{instruction} field & harmful / unharmful \\
VizWiz & Original \texttt{question} field & answerable / unanswerable \\
Matching MI & \texttt{Do the two images satisfy the induced relationship?}
& Yes / No \\
SugarCrepe & \texttt{Does this image show '<caption>'?} & Yes / No \\
MHaluBench & Original \texttt{claim} field & hallucination / non-hallucination \\
NaturalBench & \texttt{Does this image show '<caption>'?} & Yes / No \\
\hline
\end{tabular}
\caption{Question and answer formatting. Matching MI uses two images
per example; all other benchmarks use one. Caption-based questions
are read directly from the formatted dataset files.}
\label{tab:question_formats}
\end{table}

\end{document}